\documentclass[12pt,reqno]{amsproc}
\usepackage[utf8]{inputenc} % allow utf-8 input
\usepackage[T1]{fontenc}    % use 8-bit T1 fonts
\usepackage{hyperref}       % hyperlinks
\usepackage{url}            % simple URL typesetting
\usepackage{booktabs}       % professional-quality tables
\usepackage{amsfonts}       % blackboard math symbols
\usepackage{nicefrac}       % compact symbols for 1/2, etc.
\usepackage{microtype}      % microtypography
\usepackage{algorithm}
\usepackage{algpseudocode}
\usepackage{amssymb}
\usepackage{amsmath}
\usepackage{amsthm}
\usepackage{amsfonts}
\usepackage{enumerate}
\usepackage{tikz-cd} 
\usepackage{comment}
\usepackage{hyperref}
\usepackage{float}
\usepackage{verbatim}
\usepackage[inline]{enumitem}
\usepackage{fancyhdr, mathpazo}
\usepackage{todonotes}
\usepackage{svg}
\usepackage{multirow, url}
\usepackage{textcomp}
\usepackage{bm}
\usepackage{courier}
\usepackage{wasysym}
\usepackage{xcolor}
\usepackage{xspace}
\usepackage{todonotes}
\usepackage[foot]{amsaddr} % Address
\usepackage{svg}
\usepackage{graphicx}

\usepackage{fullpage}
\hypersetup{colorlinks}
\hypersetup{colorlinks=true, linkcolor=red, citecolor=blue, filecolor=blue, urlcolor=blue}

\newtheorem{thm}{Theorem}
\newtheorem{cor}[thm]{Corollary}
\newtheorem{lem}[thm]{Lemma}
\newtheorem{prop}[thm]{Proposition}
\theoremstyle{definition}
\newtheorem{defn}[thm]{Definition}
\numberwithin{equation}{section} 
\numberwithin{table}{section}

\newcommand{\nc}{\newcommand}
\newcommand{\rc}{\renewcommand}
\rc{\t}{\text}
\nc{\tb}{\textbf}
\nc{\mb}{\mathbb}
\nc{\tc}{\textcolor}
\nc{\mf}{\mathfrak}
\nc{\mc}{\mathcal}
\nc\on{\operatorname}
\rc{\paragraph}[1]{\vspace{3mm}\textbf{#1}}

\nc{\CC}{\mathbb{C}}
\nc{\DD}{\mathbb{D}}
\nc{\EE}{\mathbb{E}}
\nc{\NN}{\mathbb{N}}
\nc{\RR}{\mathbb{R}}
\rc{\SS}{\mathbb{S}}
\nc{\ZZ}{\mathbb{Z}}

\rc{\d}{\on{d}}
\nc{\Prob}{\on{Pr}}
\nc{\unif}{\mathfrak{u}}
\nc{\unifmmd}{\mathfrak{m}}
\nc{\cutoff}{\xi}
\nc{\tse}{\mathcal{L}}
\nc{\pran}{{\underline{\measuredangle}}}
\rc{\angle}{\measuredangle}
\nc{\specgap}{\lambda^{\on{gap}}}
\nc{\Tail}{\on{Tail}}
\nc{\TQ}{\on{TQ}}
\nc{\Wass}{\on{W}}
\nc{\MMD}{\Delta}
\nc{\Ball}{\mc{B}}
\nc{\Hades}{\textsf{HADES}\xspace}
\nc{\codeurl}{(Fill)}
\newcommand\smallbmat[1]{{{%
  \tiny\arraycolsep=0.1\arraycolsep\ensuremath{\begin{bmatrix}#1\end{bmatrix}}}}}

\begin{document}

\title{HADES: Fast Singularity Detection \\ with Local Measure Comparison}

\author[U. Lim]{Uzu Lim}
\email{lims@maths.ox.ac.uk}
\author[H. Oberhauser]{Harald Oberhauser}
\email{oberhauser@maths.ox.ac.uk}
\author[V. Nanda]{Vidit Nanda}
\address{Mathematical Institute, 
University of Oxford, Radcliffe Observatory, Andrew Wiles Building, Woodstock Rd, Oxford OX2 6GG}
\email{nanda@maths.ox.ac.uk}

\begin{abstract}
    We introduce \Hades, an unsupervised algorithm to detect singularities in data. This algorithm employs a kernel goodness-of-fit test, and as a consequence it is much faster and far more scaleable than the existing topology-based alternatives. Using tools from differential geometry and optimal transport theory, we prove that \Hades correctly detects singularities with high probability when the data sample lives on a transverse intersection of equidimensional manifolds. In computational experiments, \Hades recovers singularities in synthetically generated data, branching points in road network data, intersection rings in molecular conformation space, and anomalies in image data. 
\end{abstract}

\maketitle

\section{Introduction}

The Manifold Hypothesis asserts that high-dimensional datasets encountered in practice tend to concentrate near smooth manifolds of low intrinsic dimension. It is often used to justify the effectiveness of machine learning algorithms in high-dimensional settings, since the curse of dimensionality can be circumvented if the data concentrates on a low-dimensional manifold. It is, however, evident that several low-dimensional (and hence, visualisable) datasets do not satisfy the Manifold Hypothesis. 
Instead, such data can have \emph{singularities} --- points at which the local geometry does not resemble $n$-dimensional Euclidean space for any $n$. Prime examples of singular loci of datasets include branching points in neurons and cosmic filaments. Furthermore, standard image datasets (such as MNIST and CIFAR-10) are known to have non-constant intrinsic dimension \cite{bcabrown_union}, whereas a connected manifold must possess the same intrinsic dimension throughout.

Whenever such non-manifold behaviour within datasets is of interest, it becomes natural to wonder whether it can be accurately and automatically identified. Particularly in large, high-dimensional datasets where visual inspection is impossible, we seek tools to identify and locate singularities within datasets. Our focus here is on unsupervised singularity detection, where one has recourse neither to a plethora of training data, nor the opportunity to regenerate samples along an unknown probability measure. 

\paragraph{This Paper.} Here we propose \Hades, a \textit{Hypothesis-testing Algorithm for the Detection and Exploration of Singularities}. The basic philosophy is rooted in two elementary observations. First, by definition, an $n$-manifold locally resembles a standard Euclidean $n$-dimensional disk; and  second, this resemblance can be precisely quantified by measuring the distance between (the local restriction of) an empirical measure and the uniform measure on the $n$-disk. \Hades employs a goodness-of-fit test to measure this distance; we are therefore able to obtain a p-value for rejecting the null hypothesis that a given data point lies in the nonsingular locus of the underlying space.

Before proceeding to the details, we highlight three important features of the proposed algorithm below ---
\begin{enumerate}
\item {\bf Efficiency: } \Hades uses an explicit formula for kernel MMD (maximum mean discrepancy) to perform its goodness-of-fit test; this has a linear time complexity in the dimension of data, which forms a substantial improvement on the exponential complexity of the existing topological methods. 
\item {\bf Correctness: } We show in Theorem \ref{thm:main} that \Hades correctly identifies the singular set arising from the union of two transversely intersecting equidimensional submanifolds of Euclidean space. The proof uses tools from differential geometry, optimal transport theory, and concentration inequalities. 
\item {\bf Validation: } In Section \ref{sect:exp}, we run \Hades on several synthetic and real datasets. On synthetic data, we observe that singularities are correctly detected (Figure \ref{fig:synthetic lowdim}, \ref{fig:synthetic highdim}). And in the real datasets where we have no access to ground truth, the singularities identified by \Hades exhibit interesting and observable anomalous behaviour when compared to their nonsingular counterparts.
\end{enumerate}

\paragraph{Related Work.} Identifying non-manifold points and studying their structure often goes under the name of \textit{stratified learning}, which attempts to model data using stratified spaces, instead of manifolds. An early example of studying non-manifold behaviour in data is seen in \cite{haro}, where a Poisson mixture model was used to measure locally evaluated intrinsic dimension that may vary across data. 
Follow-up works considered data sampled from a union of multiple manifolds. In multi-manifold clustering, one starts with a data sampled from a union of intersecting manifolds and clusters data by separating them into the individual manifolds \cite{strat_multispace, strat_multimanifold, strat_multimanifold2, strat_multimanifold3, strat_multimanifold4, arias_spectral}. Evidence for real world data containing multiple manifolds of mixed dimension have been recently studied \cite{bcabrown_union, bcabrown_union2, cyclo_octane}. We remark that unions of manifolds only constitute a small subset of all stratified spaces. While our algorithm doesn't recover the structural information of manifolds, it detects more diverse types of singularities not present in a union of manifolds.

Stratification learning received considerable attention from the topological data analysis community. The flagship tool here is \textit{persistent homology}, which extracts topological information at multiple scales of data. In \cite{bendich2, bendich3, bendich1}, persistent intersection homology was used to discover stratified structure of data. In \cite{waas, bokor1, bokor2}, algorithms for recovering low-dimensional stratification structure and homotopy type of a stratified space has been studied. Discovering a stratification structure of a given simplicial complex \cite{vidit_coh} and a complex projective variety \cite{vidit_conormal} has also been studied. In \cite{gadata, tardis}, persistent homology was used to detect singularities in data, and their algorithms have the same objective as our algorithm. Compared to their algorithms, our algorithm has a significantly improved time complexity and theoretical foundation.

Dimension estimation and reduction are key steps in our algorithm, for which we simply apply PCA locally. Nevertheless there are many more advanced dimension estimation methods available, such as \cite{levina_bickel, dimest_danco, dimest_fractal, dimest_mada, dimest_skew, dimest_nn, tardis}. Dimension reduction methods in the literature include \cite{dimred_lapeig, dimred_ltsa, isomap, umap, tsne}. For a survey of dimension estimation and dimension reduction algorithms, see \cite{dimest_survey, dimred_survey}.

\begin{figure}
\centering
    \includegraphics[trim = {3cm 0cm 3cm 0cm}, width=0.8\textwidth]{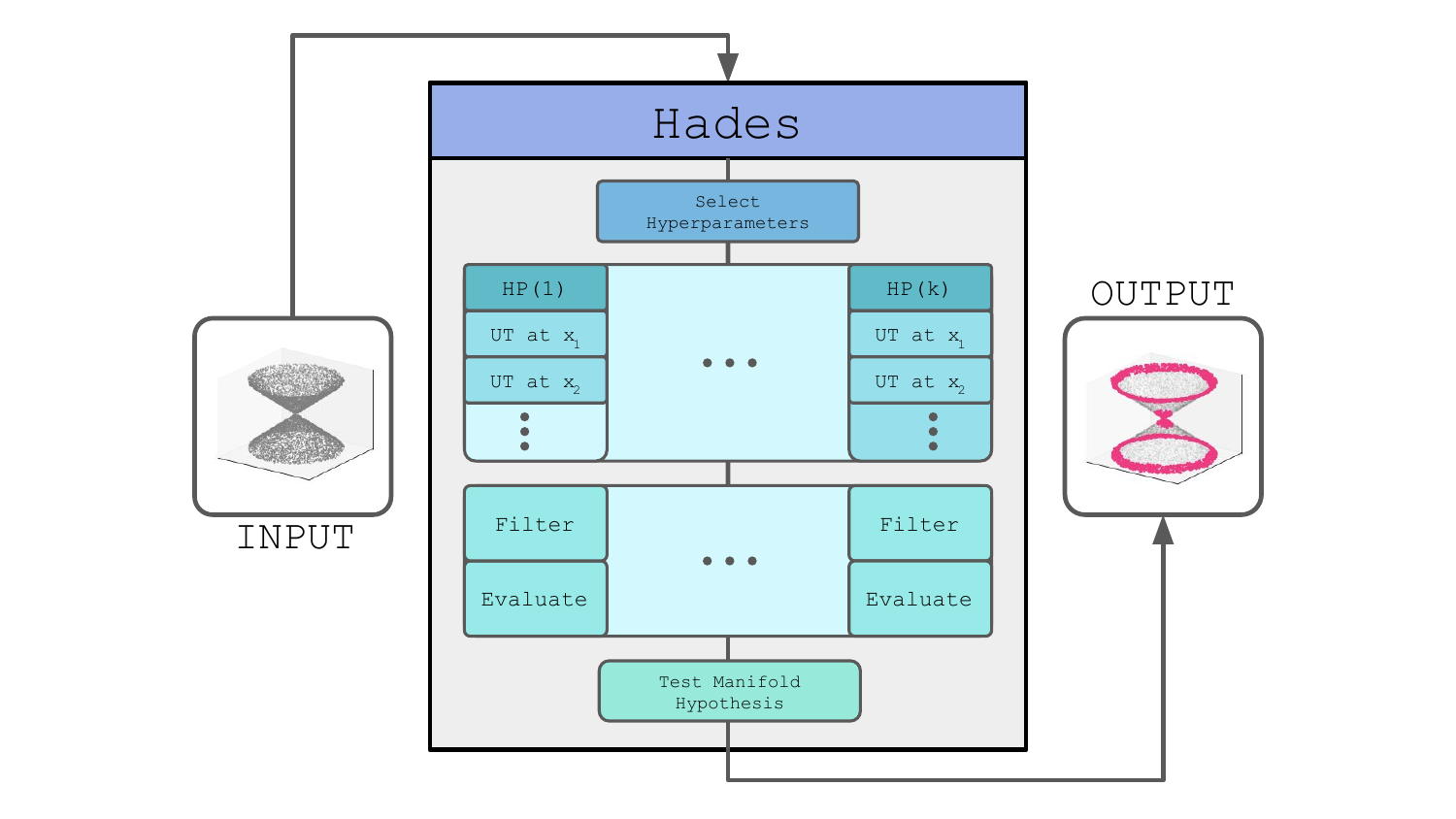}
    \caption[Algo Schematic]{Schematic for a fully automated run of \Hades. UT stands for Uniformity Test.}
    \label{fig:schematic}
\end{figure}

\pagebreak

\section{Algorithm}

The main idea of \Hades is to perform the Uniformity Test at the neighborhood of each data point. The Uniformity Test measures resemblance of each neighborhood to a flat disk, and thus determines whether a data point is smooth or singular. Hyperparameters are required for the Uniformity Test, which can be chosen either manually or automatically. The output is then filtered and evaluated, and the best set of hyperparameters is chosen based on the output evaluation. The fully automated version of \Hades is illustrated on Figure \ref{fig:schematic}.

\subsection{Uniformity Test}

The Uniformity Test works in two steps:
\begin{enumerate}
\item Dimensionality reduction
\item Goodness-of-fit test against the uniform distribution over a disk
\end{enumerate}

We use PCA projection to reduce dimension of each \textit{local} neighborhood, and this requires a threshold hyperparameter $\eta$. Suppose the points $\mathbf z \subset \RR^D$ are plugged into the Uniformity Test. Firstly its estimated dimension $\hat d$ is the number of principal components required to explain $\eta$ of the total variance. Then the local neighborhood of data is projected to the $\hat d$ principal components, producing $\tilde{\mathbf z}$.

The goodness-of-fit test uses a kernel method. Namely, we first compute the MMD (maximum mean discrepancy) and then compute the p-value associated to a null hypothesis. Given $\tilde{\mathbf z}$ obtained from projection, define the empirical measure $\hat \mu_z= n_z^{-1}\sum_{x \in \tilde{\mathbf z}} \delta_x$, where $\delta_x$ is the Dirac delta measure centered at $x$. Let $\unif_{\hat d}$ be the uniform measure over the unit $\hat d$-dimensional disk. Let $\Delta$ be the kernel MMD (maximum mean discrepancy) associated to a kernel $\kappa$. Let $S = \MMD(\mathbf X_n, \unif_d)$ follow the null distribution, where $\mathbf X_n$ is an i.i.d. sample of size $n$ drawn from $\unif_d$. We define the \textit{singularity score} $\sigma(z)$ and the \textit{singularity p-value} $\tilde{\sigma}(z)$ as follows:
\begin{align*}
    & \sigma (z) = \MMD (\hat \mu_z, \unif_{\hat d}) \\
    & \tilde{\sigma} (z) = \mathbb{P}\bigg[ S \ge \sigma(z) \bigg]
\end{align*}

Let’s understand the intuition behind the Uniformity Test. Suppose a data distribution $\mu$ is a uniform distribution on $M \subseteq \RR^D$. Given $x$, consider $\mu_{x,r}$ = the restriction of $\mu$ to the ball of radius $r$ centered at $x$. There are two possible outcomes of the Uniformity Test:

\begin{enumerate}
    \item \textbf{If $M$ is smooth at $x$.} Then $\mu_{x,r}$ is supported on a slightly curved disk when $r$ is sufficiently small. The dimensionality reduction flattens the slightly-curved disk into $\mu^\perp_{x, r}$, which is supported on a flat disk and has a small non-uniformity. Then $\mu^\perp_{x, r} \approx \unif_{\hat d}$ and thus the goodness-of-fit test fails to reject the null hypothesis. The point $x$ is declared as highly \textit{unlikely} to be singular.
    \item \textbf{If $M$ is singular at $x$.} The support of $\mu_{x, r}$ does not resemble a flat disk no matter how small $r$ is. The dimensionality reduction identifies a low-dimensional subspace containing the $\tilde \mu_{x, r}$, but $\tilde \mu_{x,r}$ only takes up a small portion of the Euclidean ball it spans. As such, the goodness-of-fit test will reject the null hypothesis. The point $x$ is declared as highly \textit{likely} to be singular.
\end{enumerate}

\paragraph{Technical details.} We use the following formula to compute the MMD of a power series kernel. Its proof is given in Appendix \ref{app:mmd_formula}.
\begin{thm}\label{thm:mmd_eval}
    Let $\hat \mu_n = \frac1n(\delta_{x_1} + \cdots + \delta_{x_n})$ be a discrete (non-random) measure and let $\unif_d$ be the uniform distribution over the unit $d$-dimensional disk in $\RR^d$. Let $\kappa$ be a kernel given by $\kappa(x,y) = \sum_{k=0}^\infty a_k \langle x,y\rangle^k$, and let $\Delta$ be the MMD associated to $\kappa$. Then we have:
    \begin{align*}
        \MMD^2(\hat\mu_n, \unif_d) = \frac1{n^2}\sum_{i=1}^n \sum_{j=1}^n \kappa(x_i, x_j) + \sum_{k=0}^\infty a_{2k} \beta_{d, k} \left( \frac d{d+2k} - \frac2n \sum_{i=1}^n \|x_i\|^{2k} \right)
    \end{align*}
    where the numbers $\beta_{d,k}$ are defined using the Gamma function $\Gamma$ as:
    \begin{align*}
        \beta_{d, k} = \frac1{\sqrt \pi} \frac{\Gamma(\frac d2 + 1) \Gamma(k + \frac12)}{\Gamma(k + \frac d2 + 1)}
    \end{align*}
\end{thm}

To evaluate the p-value arising from the MMD, we use its asymptotic distribution for large sample size $n$. The MMD is a V-statistic, for which asymptotic convergence under scaling by sample size holds true (Section 5, \cite{serfling}):
\begin{thm}\label{thm:asymptotic}
    Let $\mu$ be a Borel measure on $\mathcal X \subseteq \RR^d$ and let $\hat \mu_n$ be the empirical measure of size $n$ drawn from $\mu$. Let $\kappa: \mathcal X \times \mathcal X \rightarrow \RR$ be a function satisfying $\kappa(x,y) = \kappa(y, x)$, and let $\Delta$ be the MMD associated to $\kappa$. Then there is a converge in distribution as $n \rightarrow \infty$:
    \[ n \cdot \MMD^2 (\hat \mu_n, \mu ) \longrightarrow c_{\kappa} + \sum_{i=1}^\infty \lambda_i(Z_i^2 - 1) \]
    Here $Z_k$ are independent standard normals, $c_K := \EE [\kappa(X, X)] - \EE [\kappa(X, Y)]$, and $\lambda_k$ are eigenvalues of the integral operator:
    \[ L[\phi] = \int \tilde \kappa(x, -) \phi(x) \d \mu(x) \]
    where $\tilde \kappa(x,y) = \kappa(x,y) - \EE [\kappa(X, y)] - \EE [\kappa(x, Y)] + \EE [\kappa(X, Y)]$. 
\end{thm}
We obtain the asymptotic distributions by Monte Carlo, i.e. by directly sampling the null statistics $n \cdot \MMD^2(\hat\mu_n, \mu)$ and using this to construct an empirical cumulative distribution function. To compute p-values for events that lie far outside the Monte Carlo simulation, we use exponential decay to estimate the p-values (see \cite{tail_rootzen}).

\paragraph{Remark.} There are many alternative choices for dimension estimation, dimensionality reduction, and the statistical distance used to perform the goodness-of-fit test. These parts can be swapped out in a modular way, and the algorithm can be modified to match the user's needs. For example, a more sophisticated application might use a fractal dimension estimator, use UMAP \cite{umap} to perform local dimensionality reduction, and use statistical distances such as the Wasserstein distance. In practice, we found that the kernel MMD is more sensitive to detecting non-uniformity compared to the Wasserstein distance or its regularised Sinkhorn approximation \cite{sinkhorn}.  Future improvements of the algorithm could be made by fine-tuning each of this step while keeping using the same conceptual framework. We also remark that Uniformity Test can be trivially parallelized since all computations are local. Thus running \Hades parallel on multiple cores directly speeds it up.

\subsection{Filtering and Evaluation}

We now explain how to filter the singularity p-values into a binary label and evaluate the quality of the labeling. The labeling quality is ascribed to the hyperparameter set used to run the Uniformity Tests, and this gives us a way to choose the best set of hyperparameters. 

The singularity p-values are filtered by applying a knee detection algorithm to the empirical probability density function of $\log (1/\tilde{\sigma}_i)$. Here, we apply the logarithm to separate very small p-values. We use Gaussian kernel density estimation to produce an empirical probability density function. We use the Kneed \cite{kneed} algorithm to detect the knee of the probability density, and declare all points appearing after the knee to be singular points. The knee detection effectively identifies smooth points since their singularity p-values are relatively large, so that their distribution of $\log(1/\tilde{\sigma}_i)$ forms a concentrated mass near $0$. 

The quality of the binary label produced by the filtering step is evaluated using a metric we named \textit{dispersion score}. The dispersion score is defined purely using data points and any binary label on them. The dispersion score is defined using \textit{purity score} and \textit{separation score}:

\begin{defn}
    Let $\mathbf x = (x_1, \ldots x_n), \> \mathbf y = (y_1, \ldots y_n)$ be points and their binary labels, $x_i \in \RR^D, \> y_i \in \{0, 1\}$. For each $i = 1, \ldots n$, let $\mathcal N(i) \subseteq \{1, \ldots n\}$ be a set satisfying $i \in \mathcal N(i)$. Define a partition $\mathcal I_0 \sqcup \mathcal I_1 = \{1, \ldots n\}$, where $\mathcal I_a = \{i \>|\> y_i = a \}$. 

    The \textit{purity score} $p_i$ is the proportion of indices $j \in \mathcal N(i)$ with $y_j = 1$:
    \begin{align*}
        p_i(\mathbf y, \mathcal N) =& \frac{\# (\mathcal{N}(i) \cap \mathcal{I}_1)}{\# \mathcal{N}(i)}
    \end{align*}
    The \textit{separation score} is defined as an AUC (area-under-curve) score:
    \begin{align*}
        s_i(\mathbf x , \mathbf y, \mathcal N) =& \on{AUC}\bigg\{ (t_{ij}, \> y_j)  \> \bigg| \> j \in \mathcal{N}(i) \bigg\}
    \end{align*}
    where $t_{ij}$ are real numbers defined as follows:
    \begin{align*}
        t_{ij} =& \bigg \langle x_j - x_i, \> \frac{\tilde x_i}{\| \tilde x_i \|} \bigg \rangle, \quad \text{where} \quad \tilde x_i = \sum_{j \in \mathcal N(i) \cap \mathcal I_1} (x_j - x_i)
    \end{align*}
    The \textit{dispersion score} is defined as:
        \[ \mathfrak{D}(\mathbf x, \mathbf y, \mathcal N) = \alpha \cdot \mathcal D_1(P) +  \sum_{i \in \mathcal I_1} \mathcal{D}_2(q_i), \quad \text{where} \quad q_i = 1 - \frac12(s_i + p_i)  \]
    where $P = \#(\mathcal I_1) / n$ is the global purity score, $\alpha$ is a regularisation constant, and $\mathcal{D}_1, \mathcal{D}_2$ are \textit{damping functions}, which are bijections $\mathcal{D}_i: [0,1] \rightarrow [0,1]$ satisfying $\mathcal{D}_i(x) \le x$. \footnote{In the code, the default choice of the damping functions is given by $\mathcal{D}_1 = F_{0, 2}$ and $\mathcal{D}_2 = F_{0.5, 5}$, where $F_{a, b}(t)= (\frac{t-a}{1-a})^b$. } 
\end{defn}

Separation score quantifies how well the binary labels are cleanly separated along locally defined axes of direction, $\tilde x_i$. Indeed $\tilde x_i$ is the sum of displacements $x_j - x_i$ for which $y_j = 1$, and $t_{ij}$ is the projected length of the displacement $x_j - x_i$ onto $\tilde x_i$. Thus, $s_i$ measures how well the numbers $t_{ij}$ can classify the binary labels $y_j$ when $j \in \mathcal N(i)$.

Dispersion score detects points $x_i$ for which both $s_i$ and $p_i$ are \textit{simultaneously} small, whilst also penalising the degenerate case $P \approx 1$, when almost all points satisfy $y_i = 1$. The points $x_i$ satisfying $i \in \mathcal I_1$ and $s_i + p_i \approx 0$ are far away from other indices $j \in \mathcal I_1$, and have poorly defined local boundary for separating the label 1 from the label 0. By using the damping functions $\mathcal D_1, \mathcal D_2$, we ensure that only the points $x_i$ for which $q_i$ is sufficiently large make a meaningful contribution to $\mathfrak D$, and also only the degenerate case for which $P \approx 1$ makes a meaningful contribution to $\mathfrak D$.

\paragraph{Remark.} \Hades is an \textit{unsupervised learning} algorithm, for which there is no training dataset whose loss value can be minimised over many sets of hyperparameters. Instead, like clustering algorithms, the best set of hyperparameters is chosen by optimising a qualitatively defined criterion - the dispersion score. The dispersion score differs from the classical clustering quality measures that rewards concentration around centroids of clusters. The difference is that it aggregates \textit{local} clustering information gathered from the data points, and thus the dispersion score can still be made small for complex shapes formed by the binary labels. This is adequate since the set of singularities of a stratified space have no reason to be concentrated around their centroid. (See Figure \ref{fig:synthetic lowdim}, the singular points marked in blue are not point-like clusters sought by the classical clustering quality measures.)

\subsection{Hyperparameter selection}

\Hades uses the following three hyperparameters:

\begin{enumerate}
\item \textbf{Local radius $r$.} Used to isolate neighborhoods. 
\item \textbf{PCA threshold $\eta$.} Used for dimension estimation. 
\item \textbf{Kernel parameter $\alpha$.} Used in MMD of the Uniformity Test.
\end{enumerate}

The hyperparameters have the following effects. The radius $r$ and threshold $\eta$ both need to be at the right range to ignore noise and curvature. (For a thorough mathematical analysis, see \cite{uzu_pca}) The effect of the kernel parameter $\alpha \in (0, 1)$ is less obvious. Choosing a different kernel parameter causes a different local singular geometry to be penalised. However, we found that the correctness of the output has a low sensitivity to the kernel parameter.

We explain how the sets of hyperparameters to run are automatically chosen by \Hades. As explained before, the basic idea is to optimise the dispersion score over multiple sets of hyperparameters. These can be either supplied manually by the user or be chosen automatically by \Hades. In the automatic hyperparameter selection, we use a grid $r \in [r_{\min}, r_{\max}]$, $\eta \in [\eta_{\min}, \eta_{\max}]$, $\alpha \in [\alpha_{\min}, \alpha_{\max}]$, where we use default values of $\eta \in [0.7, 0.9]$ and $\alpha \in [0.3, 0.7]$ for the PCA threshold and the kernel parameter. 

Meanwhile, the range of radius hyperparameter $r$ is chosen using a local scale detection algorithm. The idea here is to slowly enlarge a local neighborhood until the intrinsic dimension of the neighborhood stabilises. This process is done for multiple data points, and curves of intrinsic dimension estimates are averaged over them. The Kneed algorithm \cite{kneed} is used to detect the threshold at which intrinsic dimension stabilises, by going backwards from the dimension estimate of the largest neighborhood and shrinking them, and detecting a knee of the curve. The standard intrinsic dimension estimator by Levina-Bickel \cite{levina_bickel} was used to calculate the intrinsic dimensions of the expanding neighborhoods. After obtaining the knee $\tilde r$, we use the range $r \in [1.5 \tilde r, 5.0 \tilde r]$.

When the optimal set of hyperparameters is found at the boundary of the grid search, \Hades expands the search range towards that direction of the hyperparameter grid. For example, consider the grid search on $(r, \eta, \alpha) \in [0.1, 0.2] \times [0.7, 0.9] \times [0.3, 0.7]$, and suppose the dispersion score was minimised for $(r, \eta, \alpha) = (0.2, 0.8, 0.5)$. Since the optimal choice of $r$ is found at the maximum of the range $[0.1, 0.2]$, \Hades will do another grid search on the range $[0.2, 0.3] \times [0.7, 0.9] \times [0.3, 0.7]$ afterwards\footnote{In the implementation, we actually use a slightly more sophisticated method for expanding radius range. Observe that the volume of a $d$-dimensional ball with radius $r$ is $\omega_d r^d$. We expand the radius parameter range such that this estimated local volume expands linearly.}. This process is repeated until a pre-specified end of search bounds are reached.

\subsection{Testing the Manifold Hypothesis}\label{sect:mh_explanation}

We explain an algorithm used to test whether the geometric space underlying a dataset is a manifold. The main idea is the following: Given an iid sample $X_1, \ldots X_n$ drawn from a  geometric space $M$ and singularity p-values $\sigma_1, \ldots \sigma_n$ calculated from them, the following should hold:

\begin{itemize}

\item If $M$ is a manifold, then $\sigma_1, \ldots \sigma_n$ should distribute uniformly over $[0,1]$.

\item If $M$ has a singularity, then $\sigma_1, \ldots \sigma_n$ should be concentrated near $0$.

\end{itemize}

We give a heuristic argument for the above criterion. Firstly given any random variable $Z$ with the probability density $\varphi$, the p-value variable $\tilde Z := \int_{Z}^\infty \varphi(t) \d t$ is follows the uniform distribution over $[0, 1]$. This is because if we let $\tilde a = \int_a^\infty \varphi$, we have:

\[ \Prob[\tilde Z \le \tilde a] = \Prob\left[ \int_Z^\infty \varphi \le \int_a^\infty \varphi \right] = \Prob[Z \le a] = \tilde a \]

Now for a fixed $i$, consider the singularity p-value $\sigma_i$ calculated at the neighborhood of $X_i$ by using the random sample $X_1, \ldots X_n$ drawn iid from a $d$-dimensional \textit{manifold} $M$. Assuming that the local radius parameter is sufficiently small and $n$ is sufficiently large, (1) the estimated dimension at $X_i$ is $d$ and (2) the empirical measure formed by the local neighborhood at $X_i$ closely approximates the uniform distribution over a tangential disk at $X_i$. 

Conditioning on $k$ points among $X_1, \ldots X_n$ landing in the local neighborhood of $X_i$, we see that this marginal distribution of $\sigma_i$ approximates $\tilde Z_k$. Here, $Z_k = \Delta(\hat \nu_k, \unif_d)$ where $\hat \nu_k$ is the empirical distribution constructed from an iid sample of size $k$ drawn from the uniform distribution $\unif_d$, and $\tilde Z_k$ is the p-value of $Z_k$ constructed in the way described above. Therefore, we expect $\sigma_i$ to be approximately uniformly distributed over $[0,1]$ when $r \rightarrow 0, n \rightarrow \infty$. Lastly, assuming sufficiently small $r$, most pairs of local neighborhoods at $X_1, \ldots X_n$ do not overlap, and we may expect the singularity scores $\sigma_1, \ldots \sigma_n$ to behave almost independently, so that their distribution over $[0,1]$ is almost uniform. On the contrary, if $M$ possessed a singularity, then near each singularity the singularity score (which is kernel MMD) becomes large and the singularity p-value will become small. Thus we expect a high concentration of singularity p-values near $0$ if $M$ has a singularity.

To differentiate between a uniform distribution of p-values over $[0,1]$ and a distribution possessing a sharp spike of p-values near 0, we use the following three methods:

\begin{enumerate}
	\item SUPC (Small Uniformity p-value Concentration). Choose threshold values $\{q_1, \ldots q_k\} \subset [0,1]$ and for each $q \in \{q_1, \ldots q_k\}$, calculate
\[ \on{SUPC} := \max(q_1^\dagger, \ldots q_k^\dagger) \text{, where } q^\dagger = \frac{\# \{ \sigma_i \le q \} }{nq} \]
	\item UPUP (Uniformity p-value Uniformity p-value). Construct an empirical distribution $\hat \nu$ from $\sigma_1, \ldots \sigma_n$ and perform the uniformity test using the kernel MMD method (Theorem \ref{thm:mmd_eval}).
	\item KS (Kolmogorov-Smirnov). Again construct $\hat \nu$ from $\sigma_1, \ldots \sigma_n$ and perform the one-sample Kolmogorov-Smirnov test against the uniform distribution over $[0,1]$.
\end{enumerate}

\begin{figure}
\centering
    \includegraphics[width=0.21\textwidth]{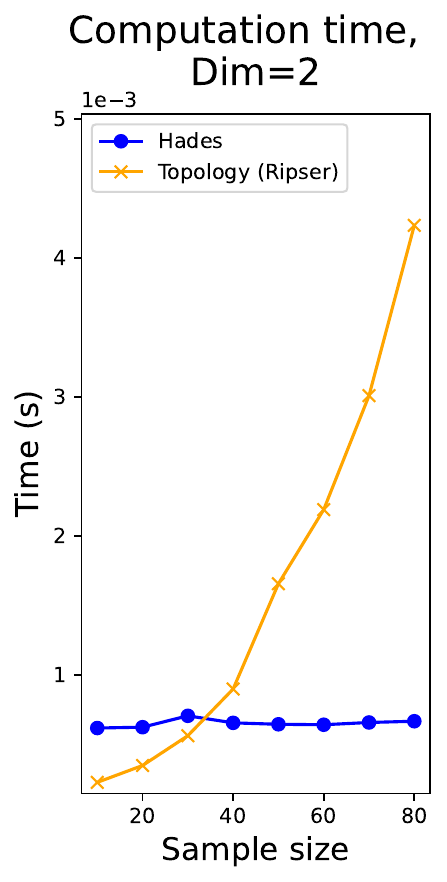}
    \includegraphics[width=0.21\textwidth]{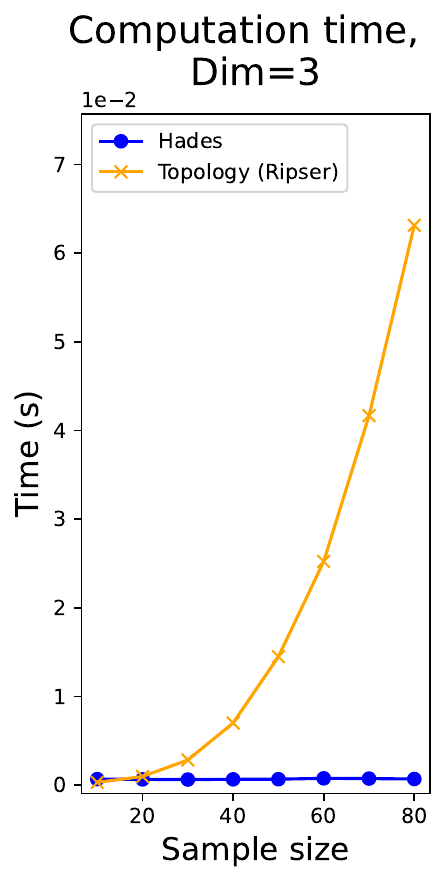}
    \includegraphics[width=0.22\textwidth]{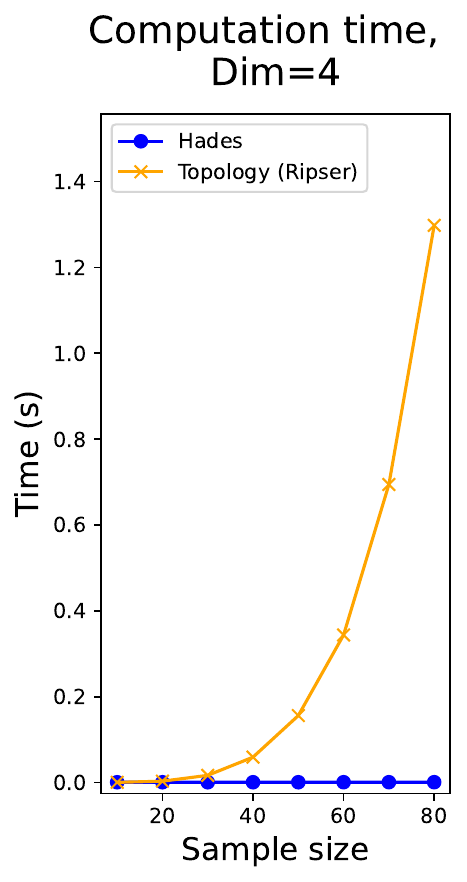}
    \includegraphics[width=0.215\textwidth]{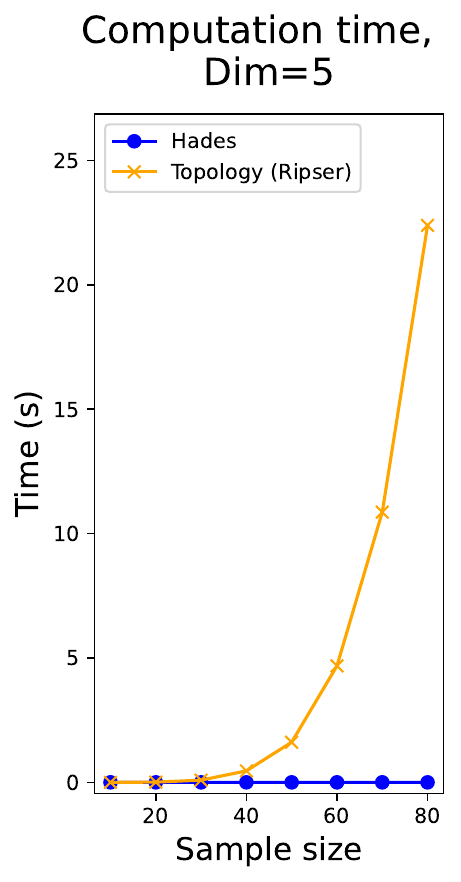} \\
    \vspace{2mm}
    \includegraphics[trim = {1cm 0cm 1cm 0cm}, width=0.43\textwidth]{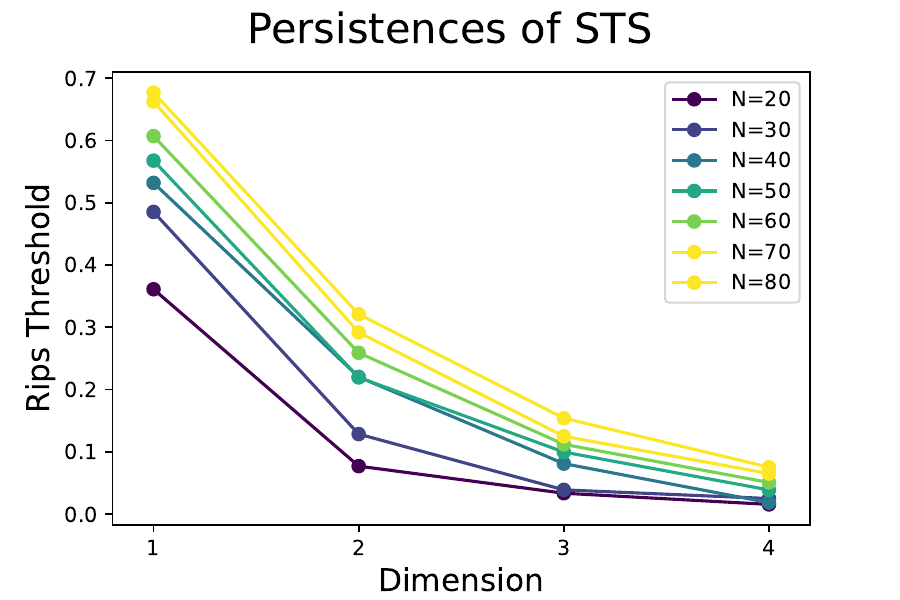}
    \includegraphics[trim = {1cm 0cm 1cm 0cm}, width=0.43\textwidth]{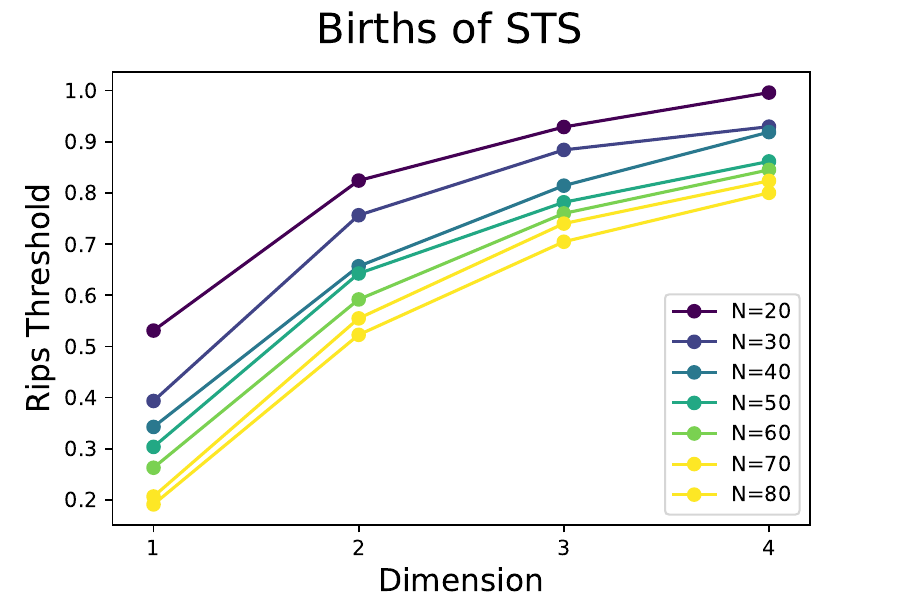}
    \caption[Comparison of \Hades and Ripser]{Top Row: Comparison of computation time of local shape analysis in \Hades (blue) versus Ripser (orange), a highly optimised library for computing persistent homology. Bottom Row: Persistence of the significant topological signature (STS) in high-dimensional spheres decay significantly across dimensions.}
    \label{fig:tda compare}
\end{figure}

\section{Comparison with topological methods} 

We demonstrate significantly improved time complexity and statistical foundation of \Hades in the singularity detection task, compared to the previous topological methods. Topological methods of singularity detection are based on persistent homology, a prominent tool from \textit{topological data analysis} \cite{tardis, gadata, bendich1, bendich2, bendich3, waas, bokor1, bokor2}. Persistent homology computes topological features at varying scales of data, and the main idea behind topological methods for singularity detection is to compute persistent homology on local neighborhoods of data. In particular, the recent algorithms in \cite{tardis, gadata} uses the fact that a small annular neighborhood of a point on a manifold has the topology of a sphere, whose topology is well-understood.

\paragraph{Time complexity.} A major advantage of \Hades over singularity detection algorithms based on persistent homology is that \Hades scales much better to high-dimensional data. We firstly have the following:
\begin{thm}
    The time complexity of the Uniformity Test on $n$ points in $\RR^D$ is $O(n^2 D)$.
\end{thm}
\begin{proof}
    The dimensionality reduction step is an application of PCA. SVD is performed on a rectangular matrix of dimension $k \times D$, which has the time complexity of $O(k^2 D)$ \cite{svd_time}. With the estimated dimension of $\hat d$, a matrix multiplication between a rectangular matrix of size $(k \times \hat d)$ and a diagonal matrix of size $(\hat d \times \hat d)$ is performed, for which the time complexity is $O(k \hat d)$, which is less than $O(k^2 D)$. This step thus amounts to the time complexity of:
    \[ T_1 = O(k^2 D) \]
    
    The goodness-of-fit step computes the MMD of a $\hat d$-dimensional point set of size $k$. Following the expression computed in Theorem \ref{thm:mmd_eval}, the time complexity for this step is:
    \[ T_2 = O(k^2 d + k + kd) = O(k^2 d) \]
\end{proof}

In comparison, the time complexity of persistent homology increases exponentially in the intrinsic dimension of data. The computational complexity of Ripser \cite{ripser}, a highly optimised Python package for computing persistent homology, is $O(s^3)$ where $s$ is the number of simplices constructed. However a dataset of $k$ points has a total of $s = \binom k{d+1} = O(k^{d+1})$ simplices of dimension $d$. A small annular local neighborhood of a $d$-dimensional manifold is topologically a $(d-1)$-sphere, and requires computationally constructing $d$-simplices. Therefore, the computational complexity of the $(d-1)$-th persistent homology group is $O(k^{3d+3})$. Persistent homology computation corresponds to the Uniformity Test, where in our algorithm we instead use PCA and kernel MMD. Using the computational complexity of the Uniformity Test given above, we have the following comparison of computational complexity incurred by local shape analysis:
\begin{align*}
    \text{Uniformity test: } &\quad O(k^2 D) \\
    \text{Persistent homology: } &\quad O(k^{3d+3})
\end{align*}
Thus we observe an exponential dependence of persistent homology computation on the intrinsic dimension $d$ of data, whereas the Uniformity Test has a linear dependence on the ambient dimension $D$. When $D\gg d$, preprocessing data by dimensionality reduction allows us to circumvent the dependence of \Hades on the ambient dimension $D$.

In Figure \ref{fig:tda compare}, we compare computation times of \Hades (blue curve) and Ripser (orange curve). On each of the five plot shows computation times for a fixed dimension\footnote{For $d$-dimensional data, we use samples of the unit $d$-dimensional ball for \Hades and samples of the unit $(d-1)$-dimensional sphere for Ripser.}, but varying sample size. We observe that while \Hades shows poorer performance than Ripser in low-dimensional data, the situation is quickly reversed in high-dimensional data.

\paragraph{Diminishing persistence.} We observe from computational experiments that the topological signature of a high-dimensional sphere has a small persistence. This appears to present problem in applying the standard practice in topological data analysis, which declares a point on the persistent diagram as a genuine signal only if the point has a high persistence. In the case of the $d$-dimensional sphere, one seeks one highly persistent point on the $d$-dimensional persistence diagram, since the $d$-dimensional sphere has a 1-dimensional $d$-th homology group, and all other $k$-th homology groups of are zero for $k>0$.

As such we define the \textit{significant topological signature} (STS) at $(d, n)$ to be the most persistent point of $\on{PD}_d(\mathbf X_n)$, where $\on{PD}_d(\mathbf X_n)$ is the $d$-th persistence diagram of the Rips filtration on $\mathbf X_n$, and $\mathbf X_n$ is an independently and identically distributed sample of size $n$ from the $d$-dimensional sphere. Figure \ref{fig:tda compare} tabulates birth times and persistences (y-axis) of the STS at $(d, n)$ for varying sample size $n$ (x-axis) the dimension $d$ (different curves, colour-coded). The STS is significant because it is the main signal sought by the standard practice of topological data analysis.

Figure \ref{fig:tda compare} indicates that the STS of a high-dimensional sphere has a small persistence and a large birth time. The small persistence tells us that STS becomes increasingly unreliable in high dimensions, due to it resembling "topological noise". This appears to defy the current paradigm of topological data analysis where highly persistent topological features are to be seen as genuine signal and other topological features are to be seen as noise. The large birth time tells us that one cannot use small connectivity threshold to detect STS, and therefore that it is difficult to reduce the number of high-dimensional simplices appearing in the full filtration of a point cloud. 

This situation may be improved by using low-dimensional topological signal of high-dimensional spheres, which runs on smaller time complexity. In fact, even the 1-dimensional sphere (circle) exhibits systematic high-dimensional topological signals in large connectivity thresholds \cite{uzu_circle, vr_circle}, and high-dimensional spheres exhibit systematic low-dimensional topological signals.

\section{Theoretical Guarantee}

In this section, we will define a mathematically precise version of the singularity detection algorithm, and state a theorem guaranteeing that the algorithm detects singularities correctly. In the following, we fix the ambient dimension $D$, threhsold parameter $\eta \in (0, 1)$. We first define the PCA dimension estimator and projector precisely:
\begin{defn}
    Given $\mu$, a Borel probability measure on $\RR^D$, its \textit{estimated dimension} $\hat d(\mu)$ and \textit{linear regression} $\tse(\mu)$ are defined as:
    \begin{align*}
        \hat d (\mu) =& \min\bigg\{ k \>\bigg|\> \frac{\lambda_{k+1} + \cdots + \lambda_D}{\lambda_1 + \cdots + \lambda_D} \le \eta \bigg\} \\
        \tse(\mu) =& \on{span}\bigg( \mathcal{E}(\mu, \lambda_1), \ldots , \mathcal{E}(\mu, \lambda_{\hat d(\mu)}) \bigg)
    \end{align*}
    where $(\lambda_1, \ldots \lambda_D)$ are eigenvalues of $\Sigma[\mu]$, $\mathcal{E}(\mu, \lambda)$ is the $\lambda$-eigenspace of $\Sigma[\mu]$, and $\Sigma[\mu]$ is the covariance matrix of $\mu$. The parameter $\eta$ is implicit from the notations $\hat{d}(\mu), \tse(\mu)$.
\end{defn}
% We generally have $\dim \mc L(\mu) \ge \hat d(\mu)$. Equality holds iff $\lambda_{\hat d} > \lambda_{\hat d + 1}$, where $\hat d = \hat d(\mu)$ and $\vec\lambda\Sigma[\mu] = (\lambda_1, \ldots \lambda_D)$.

Using $\hat d$ and $\tse$, we define the mathematically precise version of the \textit{singularity score}. In the following let $\MMD(\mu, \nu)$ denote the kernel MMD associated to the Gaussian kernel $\kappa(x,y) = \exp(-\gamma \cdot \|x-y\|^2)$ for some fixed $\gamma > 0$. Also denote by $\unif_d$ the uniform measure over the unit $d$-dimensional disk centered at the origin. We first define the \textit{abstract singularity score}, and use this for empirical measures to define the \textit{empirical singularity score}.

\begin{defn}
    The \textit{abstract singularity score} is defined as:
    \begin{align*}
        \sigma(\mu) =& \MMD ( \mu_\perp, \unif_{\hat d} )
    \end{align*}
    where $\hat d = \hat d (\mu)$ and $\mu_\perp = \Pi(\mu, \tse \mu)$ is the pushforward of $\mu$ along the projection to $\tse \mu$.
    
    Let $\mathbf x = \{x_1, \ldots x_n\} \subset \RR^D$ and let $r>0$. Denote $\mathbf x(z) = \mathbf x \cap \Ball(z, r) \backslash \{z\}$, where $\Ball(z,r) \subseteq \RR^D$ is the open ball of radius $r$, centred at $z$\footnote{The point $z$ is excluded for a technical reason concerning Wasserstein concentration inequalities, although the proofs can be modified to be included if necessary.}. The \textit{local empirical measure} of $\mathbf x$ at $z$ is:
    \[ \hat \mu(z) = g_{z, r}\bigg( \frac1{\# \mathbf x(z)} \sum_{y \in \mathbf x(z)} \delta_y \bigg)\]
    where $g_{z,r}(\nu)$ is the pushforward of a measure $\nu$ by the affine map $x \mapsto r^{-1}(x-z)$. The $i$-th \textit{empirical singularity score} of $\mathbf x$ is defined as:
    \[ \hat \sigma_i (\mathbf x, r, \eta) = \sigma(\hat \mu(x_i)) \]
\end{defn}

Note that all of $\hat d, \tse, \sigma$ depend on the choice of dimension estimation threshold $\eta$. We now state the setup and the main theorem.

\paragraph{Setup.} Let $M = M_1 \cup M_2$, where $M_1, M_2 \subseteq \RR^D$ are smooth compact $d$-dimensional manifolds embedded in $\RR^D$. Suppose there exist $d_0, \phi >0$ such that the following holds for every $x \in M_1 \cap M_2$: the tangent spaces $T_x M_1$ and $T_x M_2$ intersect at a $d_0$-dimensional subspace, and all principal angles of the pair are $\ge \phi$. Let $\mu$ be the uniform measure over $M$, and let $\mathbf X_n = (X_1, \ldots X_n)$ be an iid\footnote{independently and identically distributed} sample of size $n$ drawn from $\mu$. 

\begin{thm}[Theoretical guarantee]\label{thm:main}
    There exist constants $\xi, \eta_-, \eta_+ , c_A, c_B, r_0 > 0$ depending only on $M$ such that the following holds. Given $\eta \in [\eta_-, \eta_+]$, $r \le r_0$, and $q \in (0,1)$, the following implications both hold for all $i$ with probability at least $q$, when $n$ is large enough:
    \begin{enumerate}
        \item When the distance of $X_i$ from $M_1 \cap M_2$ is less than $c_A r$, then $\hat\sigma_i > 2\xi$. 
        \item When the distance of $X_i$ from $M_1 \cap M_2$ is greater than $c_B r$, then $\hat\sigma_i < \xi$. 
    \end{enumerate}
    where $\hat \sigma_i = \hat \sigma_i(\mathbf X, r, \eta)$.
\end{thm}

The proof of the theorem requires much work, and it is presented in Appendix \ref{app:guarantee}. One main tool for the theorem is the Wasserstein distance, instead of the kernel MMD, which is possible since $\Delta(\mu, \nu) \le \sqrt{2\gamma} \cdot \Wass(\mu, \nu)$ for the Gaussian kernel $\kappa(x,y) = e^{-\gamma\|x-y\|^2}$ (Lemma \ref{mmd_wass_comparison}). The advantage of the Wasserstein distance is that it is intuitively easy to prove geometric claims.

It has the following key ingredients:
\begin{enumerate}
    \item For a fixed $z \in M$ and as $r \rightarrow 0, n \rightarrow \infty$, the empirical measure $\hat \mu(z)$ converges to the uniform distribution over $T_zM^\circ := T_z M \cap \Ball(0, 1)$, where $\Ball(0, 1) \subseteq \RR^D$ is the unit ball of radius $1$. Convergence is quantified using the Wasserstein distance. (Proposition \ref{main_wass_conc})
    \item The singularity score function $\mu \mapsto \sigma(\mu)$ is a Lipschitz continuous function in $\mu$, where Lipschitz continuity is quantified using the Wasserstein distance. (Proposition \ref{sing lipschitz})
    \item The singularity score of the limiting measure at each point as $r \rightarrow 0, n \rightarrow \infty$ is zero at smooth points and positive at singular points (Propositions \ref{sing limit smooth}, \ref{sing limit singular}).
    \item By moving sufficiently far away from the singularities, the local neighborhood of a point only isolates one manifold $M_i$ at a time, instead of cutting through both $M_1$ and $M_2$ (Proposition \ref{separation}).
\end{enumerate}
To understand the proof, the reader is advised to start from the last part, Subsection \ref{sect: proof_main_outline}, and work backwards to identify the components used in the proof.

We remark that the constants $c_A, c_B$ appearing in the theorem are unfortunately intrinsic features of the singularity detection. Suppose that $x \in M$, the ball of radius $r$ is used to isolate local neighborhood of $x$, and that the distance of $x$ to the singularities of $M$ is $c\cdot r$ where $c \in \RR^+$. Then there is an inherent ambiguity in choosing $c_0$ such that whenever $c > c_0$, $x$ is declared non-singular, and whenever $c < c_0$, $x$ is declared singular.

\section{Experiments}\label{sect:exp}
We implemented \Hades in Python and performed various computational experiments. Singularity detection lacks a ground truth for most real-world datasets, and is an \textit{unsupervised learning} algorithm. We follow the standard 2-step approach to assess the performance of a singularity detection algorithm:
\begin{enumerate}[label=(\roman*)]
    \item \textbf{Synthetic data.} We plot singularities detected from 2- and 3-dimensional datasets and visually inspect that the singularities are detected correctly. Then we detect singularities from families of high-dimensional synthetic datasets whose singularities are completely understood by construction, and use receiver-operating-characteristic (ROC) curve to quantitatively assess accuracy of the algorithm.
    \item \textbf{Real data.} We study datasets of road networks, cyclo-octane conformation, images of handwritten digits, and images of clothing items. For the road network and cyclo-octane conformation datasets, we recover the already-known locations of the singularities. For the image datasets whose geometry are not well-understood, we observe that images with high singularity score are anomalous from visual inspection.
\end{enumerate}
For details of the experiments, see Appendix \ref{app:experiments}.

\subsection{Synthetic data: Visualisation and ROC Curves} 

\begin{figure}
\centering
    \includegraphics[trim={1.5cm 1.5cm 1.5cm 1.5cm}, width=0.45\textwidth]{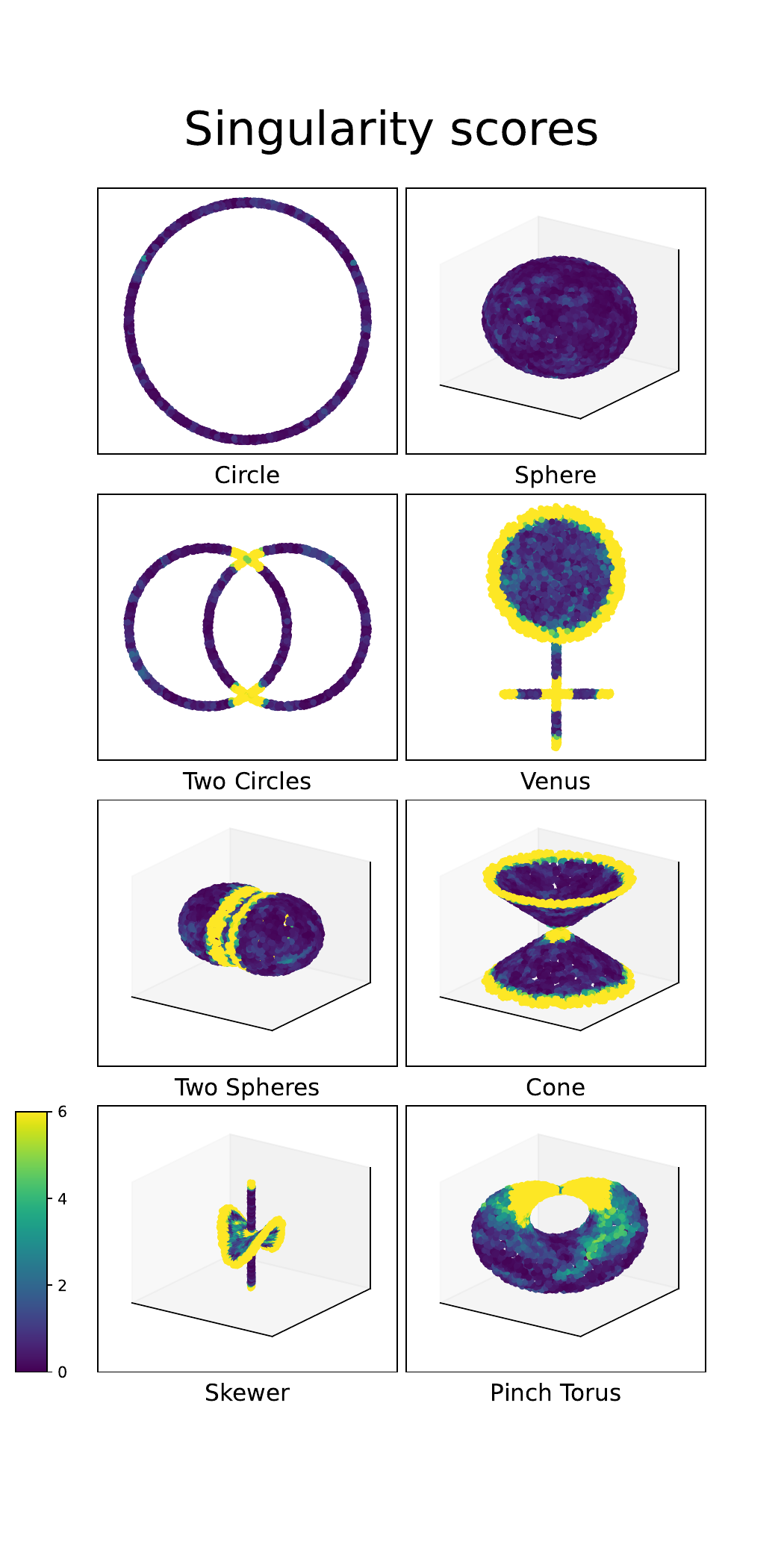}
    \includegraphics[trim={1.5cm 1.5cm 1.5cm 1.5cm}, width=0.45\textwidth]{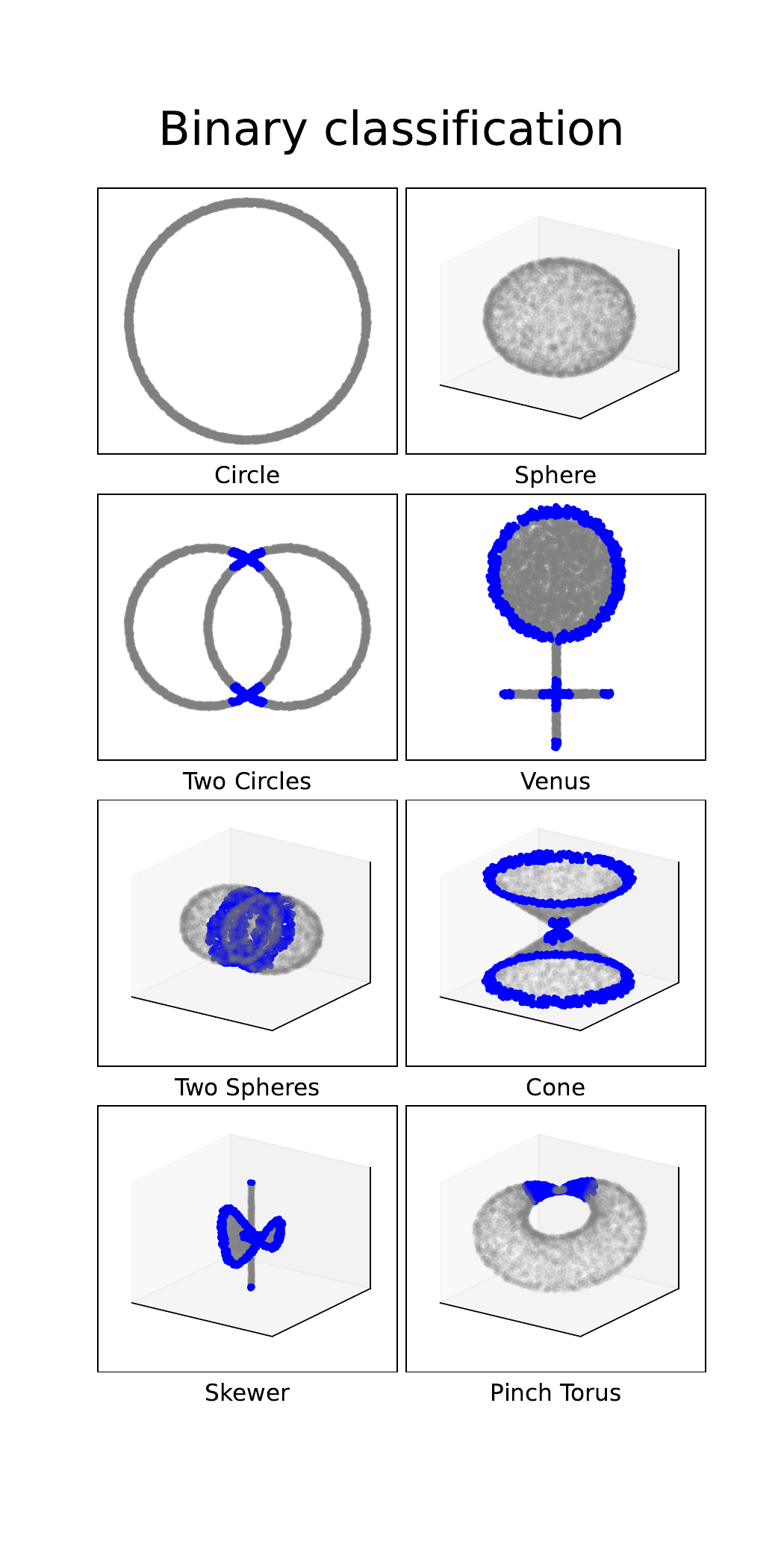}
    \caption[\Hades applied to synthetic datasets, visualised]{Singularities discovered by \Hades marked blue in synthetic datasets.}
    \label{fig:synthetic lowdim}
\end{figure}

We first apply \Hades to the 2- and 3-dimensional point clouds in Figure \ref{fig:synthetic lowdim}, where singular points detected by the algorithm are marked blue. 
These synthetic datasets are generated from known data distributions of various geometric shapes, and uniform noise has been added to the datasets. They demonstrate that the algorithm is robust to noise and curvature. The algorithm simultaneously detects multiple types of singularities such as intersections, branching points, sharp corners, and cones. We also observe that no singularities are detected for the first row, which consist entirely of manifolds. This is enabled by the manifold hypothesis testing algorithm SUPC described in \ref{sect:mh_explanation}. The sizes of datasets range from 5,000 to 10,000. The time taken to extract singularities was about 3 minutes per dataset.

\begin{figure}
\centering
    \includegraphics[width=0.9\textwidth]{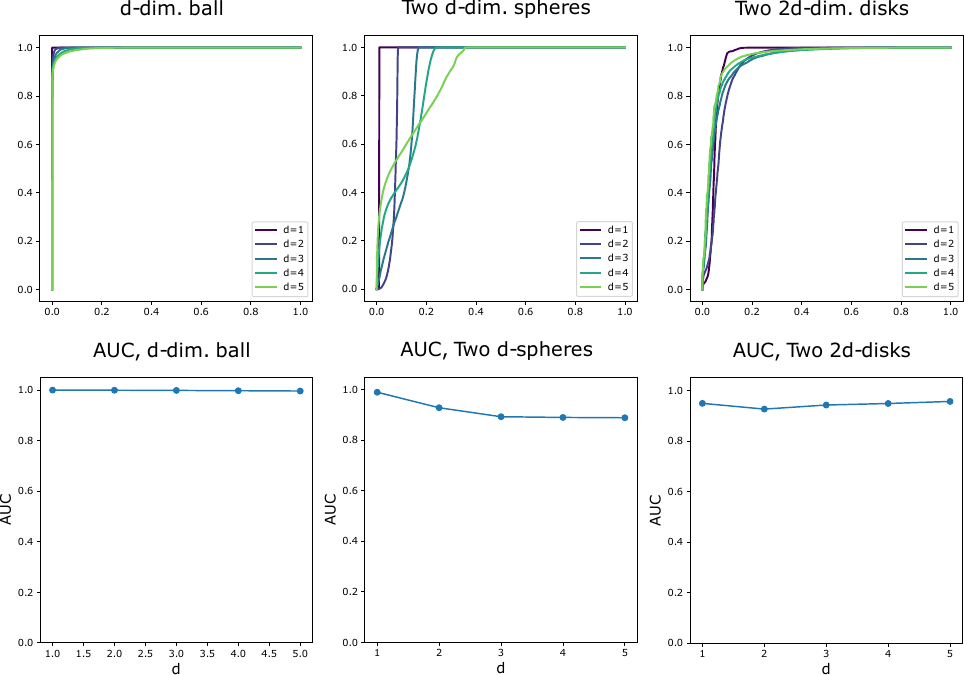}
    \caption[\Hades applied to synthetic datasets, quantified with ROC and AUC]{ROC curve and AUC scores of singularities discovered by \Hades in synthetic datasets.}
    \label{fig:synthetic highdim}
\end{figure}

Going beyond visual inspection, we quantify accuracy of \Hades on three families of geometric spaces: 
\begin{enumerate}
    \item One solid $d$-dimensional ball. (Singularity at boundary)
    \item Two $d$-dim. spheres intersecting at a $(d-1)$-dim. sphere. (Singularity at intersection)
    \item Two $2d$-dim. disks intersecting orthogonally at a $d$-dim. disk. (Singularity at intersection and boundary)
\end{enumerate} 
Visual inspection is inadequate for inspecting high-dimensional singularities, so we use receiver-operating characteristic (ROC) curve and its area-under-curve (AUC) to assess the performance. The AUC scores we obtain were all $\ge 0.89$. The ROC curves and the AUC values are shown in Figures \ref{fig:synthetic highdim}.

\begin{comment}
    More precisely, the AUC are the following for $d=1, \ldots 5$:
    \begin{align*}
        \text{One $d$-dimensional solid ball: } & \text{AUC} = (1.00, 1.00, 1.00, 1.00, 1.00) \\
        \text{Two $d$-dimensional spheres: } & \text{AUC} = (0.99, 0.93, 0.89, 0.89, 0.89) \\
        \text{Two $2d$-dimensional disks: } & \text{AUC} = (0.95, 0.93, 0.94, 0.95, 0.96)
    \end{align*}
\end{comment}

\subsection{Synthetic data: Manifold hypothesis}

\begin{figure}
\centering
    \vspace{5mm}
    \includegraphics[width=0.9\textwidth]{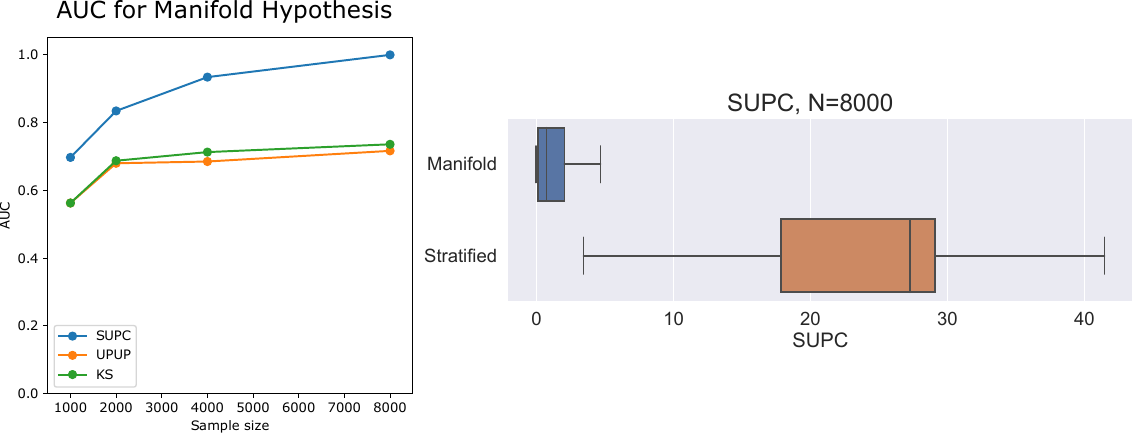}
    \caption[\Hades applied to test the manifold hypothesis]{Testing the manifold hypothesis with \Hades. Left: AUC of manifold hypothesis performed with three different scores: SUPC, UPUP, KS with synthetic datasets. Right: Boxplots of SUPC for manifolds vs. stratified spaces at sample size 8000.}
    \label{fig:synthetic mh}
\end{figure}

We test the manifold hypothesis with \Hades using the methods described in Subsection \ref{sect:mh_explanation}. Unlike the \textit{local} tests of detecting singularities, we perform a \textit{global} test, on one dataset at a time. For this, datasets consisting of synthetically generated point clouds were created, with a \textit{binary} label on whether each point cloud was a stratified space (with singularities) or a manifold (without singularities). Figure \ref{fig:synthetic mh} shows the results. As sample sizes increase from $1000$ to $8000$, AUC values for all of SUPC, UPUP, KS increase, with UPUP and KS reaching just about 0.7 and SUPC reaching the AUC score $1.00$. The boxplot on the right shows the distribution of SUPC scores for the manifolds and stratified spaces at sample size $8000$, demonstrating a clean separation between the two types of data. This indicates that the manifold hypothesis can be effectively tested with SUPC.

\pagebreak

\subsection{Real data: Road network}

\begin{figure}[H]
\centering
    \includegraphics[trim={2cm 2cm 2cm 2cm}, width=0.7\textwidth]{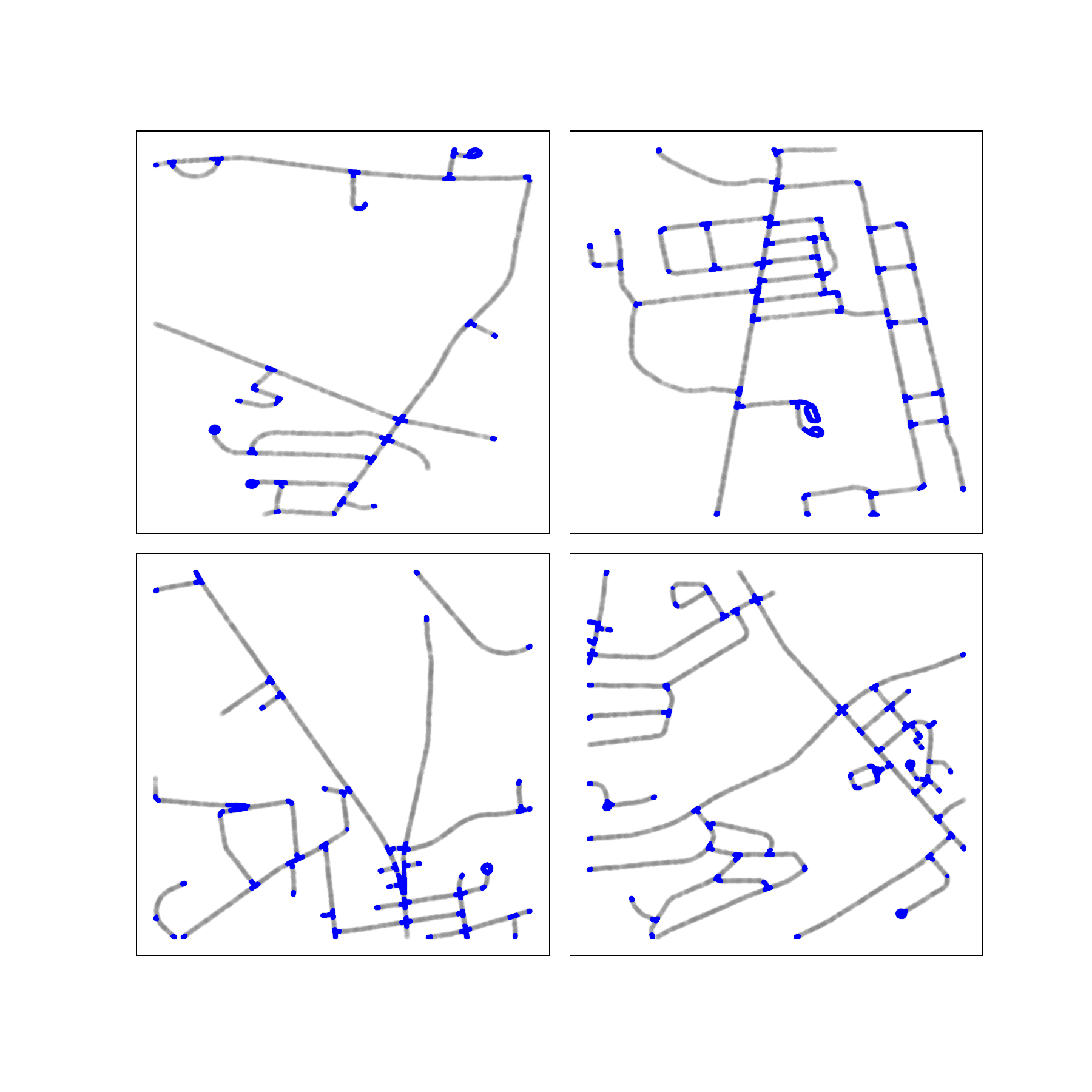}
    \caption[\Hades applied to road network datasets]{Singularities discovered by \Hades marked blue in the Massachusetts Roads Dataset.}
    \label{fig:real road}
\end{figure}

We apply \Hades to the Massachusetts Roads Dataset \cite{massroad}, a dataset consisting of pixelised images of road networks in Massachusetts. Each road network is mathematically a planar embedding of a graph. Intersections and sharp corners of the road are singular points, and everything else is locally a straight line, and thus are smooth points. From Figure \ref{fig:real road}, visual inspection reveals that singularities are accurately detected. Each image had $1500 \times 1500$ resolution, containing 45,000 to 200,000 pixels with non-zero brightness values. The time taken to run each dataset ranges from 6 to 31 seconds. Expanding this analysis, the same computational experiment can be performed to other datasets that can be modeled as (1-dimensional) graphs, including images of neurons, and filamentary structures formed by galaxies.

\subsection{Real data: Cyclo-octane conformation}

\begin{figure}
\centering
    \includegraphics[trim={3.3cm 1.5cm 3.3cm 1.5cm}, width=0.32\textwidth]{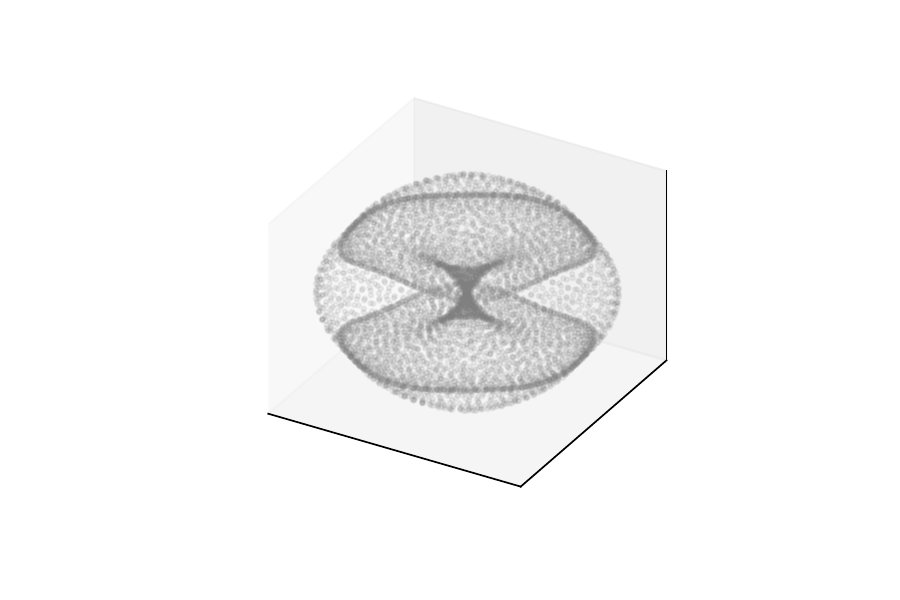}
    \includegraphics[trim={3.3cm 1.5cm 3.3cm 1.5cm}, width=0.32\textwidth]{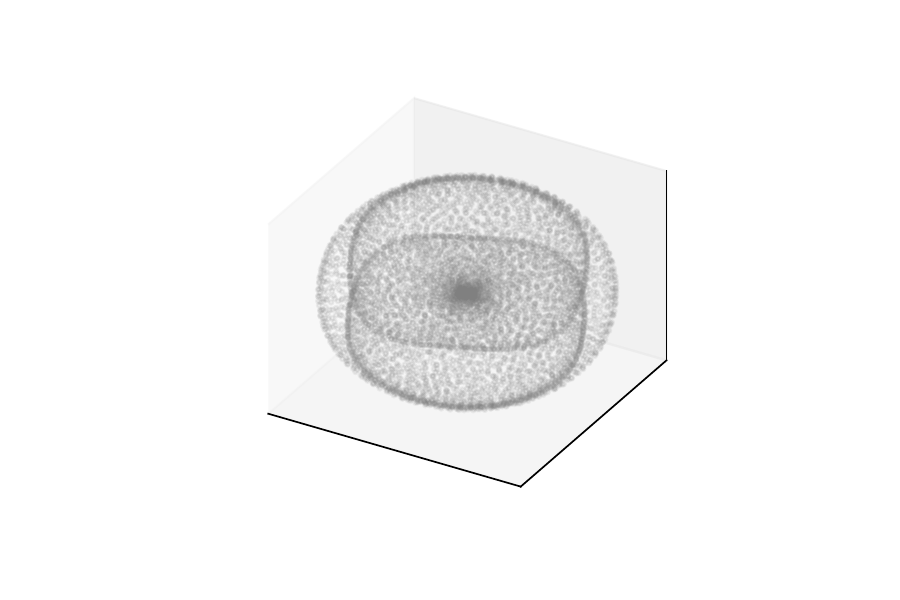}\\
    \includegraphics[trim={3.3cm 1.5cm 3.3cm 1.5cm}, width=0.32\textwidth]{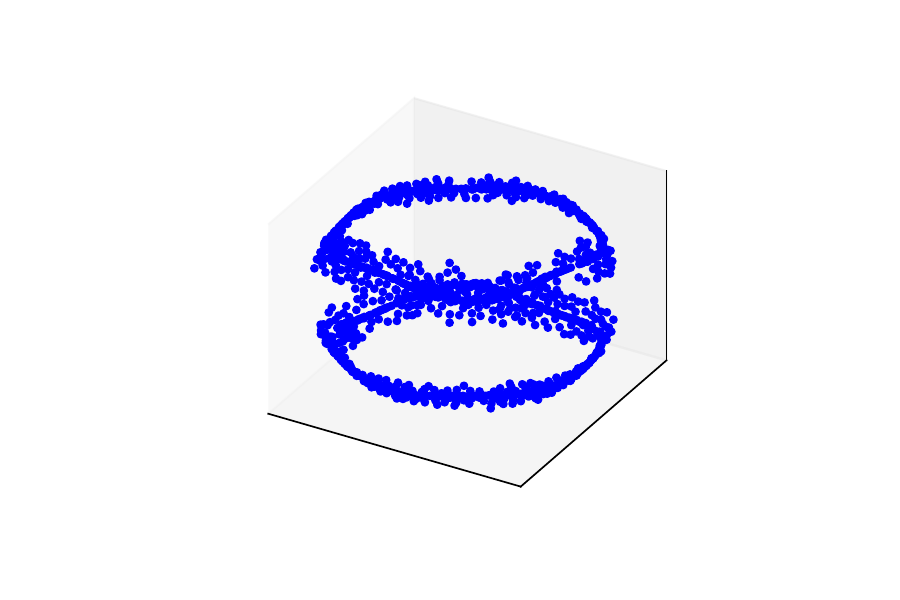}
    \includegraphics[trim={3.3cm 1.5cm 3.3cm 1.5cm}, width=0.32\textwidth]{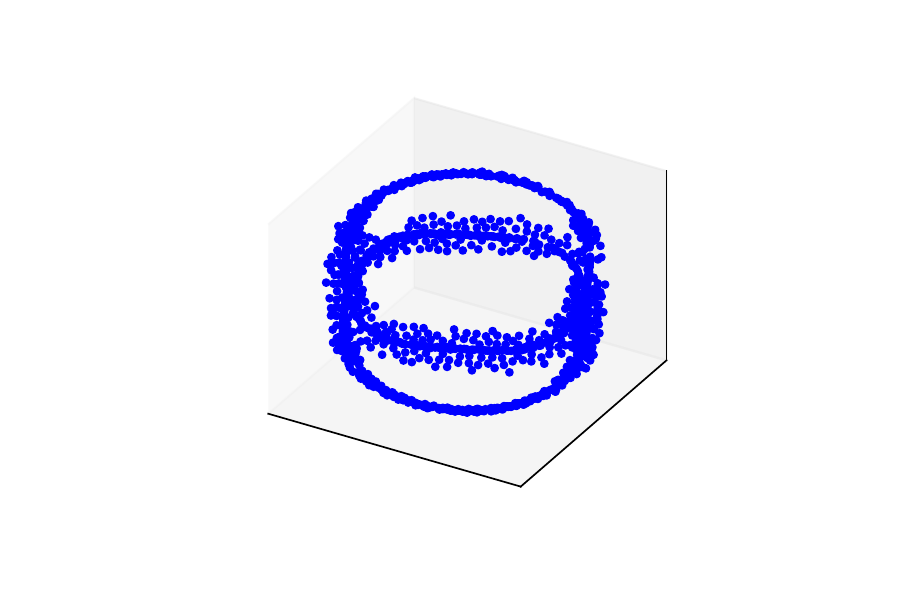}
    \caption[\Hades applied to the cyclo-octane conformation dataset]{Singularities discovered by \Hades marked blue in a cyclo-octane conformation dataset, which are union of two circles. Each row shows rotations of the same 3D Isomap projection of the 24-dimensional dataset. The first row shows the whole dataset and the second row shows singularities.}
    \label{fig:real cyclo}
\end{figure}

We apply \Hades to the dataset of cyclo-octane conformations. This dataset, introduced in \cite{cyclo_octane}, consists of 6040 points on the 24-dimensional space $\RR^{24}$ that parametrises 3D positions of 8 carbon molecules in the cyclo-octane $\text{C}_{8} \text{H}_{16}$. The space of cyclo-octane was previously identified to be the union of a Klein bottle and a sphere, intersecting at two circles \cite{cyclo_octane}. These two circles are singularities of the space of conformations, and indeed they are correctly detected by \Hades, as seen in Figure \ref{fig:real cyclo}. The 3D projections of the conformation dataset, obtained using the dimensionality reduction algorithm Isomap \cite{isomap}, is displayed in Figure \ref{fig:real cyclo}; we emphasise that the computation wasn't done on the 3D projection, and instead done directly on the original 24-dimensional data.

Running \Hades on the entire conformation dataset took 5 seconds on a standard laptop. This shows great improvement from the previous benchmark for this dataset, in \cite{gadata}, in which their singularity detection algorithm \textit{Geometric Anomaly Detection} took at least several hours on parallel processing, as informed by the first author on private communication.

\subsection{Real data: Images of handwritten digits and clothings}

We apply \Hades to image datasets, of handwritten digits (MNIST) and clothing items (Fashion-MNIST), and find that images with high singularity scores are visibly more anomalous. MNIST is a standard dataset of images of handwritten digits \cite{mnist} consisting of 60,000 data points, where there are 6,000 data points for each digit from $0, 1, \ldots 9$. Each data point is a $28 \times 28 = 784$-dimensional vector of brightness values between $0$ and $1$, where each entry of the vector indicates the brightness value of each pixel in the image. Similarly, the Fashion-MNIST dataset consists of $28 \times 28$ images of 10 classes of clothing items\footnote{T-shirt, Trouser, Pullover, Dress, Coat, Sandal, Shirt, Sneaker, Bag, Ankle boot}, where there are 6,000 data points per class.

\begin{figure}[H]
\centering
    \includegraphics[scale=0.63]{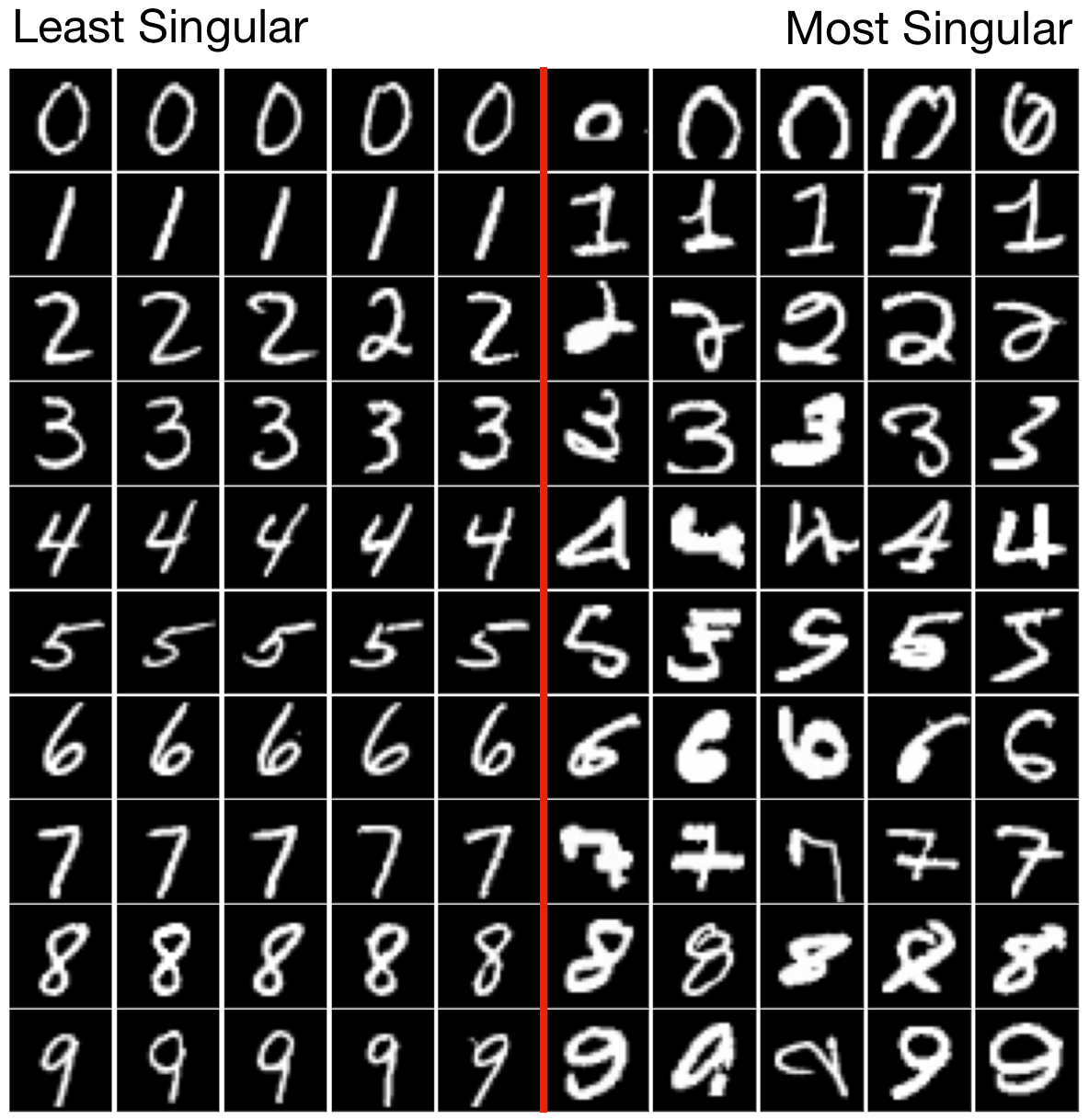}
    \caption[\Hades applied to the MNIST dataset]{Images with the lowest singularity scores (left half) and the highest singularity scores (right half), upon applying \Hades to the MNIST hand-written digits dataset.}
    \label{fig:real mnist}
\end{figure}

\begin{figure}[H]
\centering
    \includegraphics[scale=0.63]{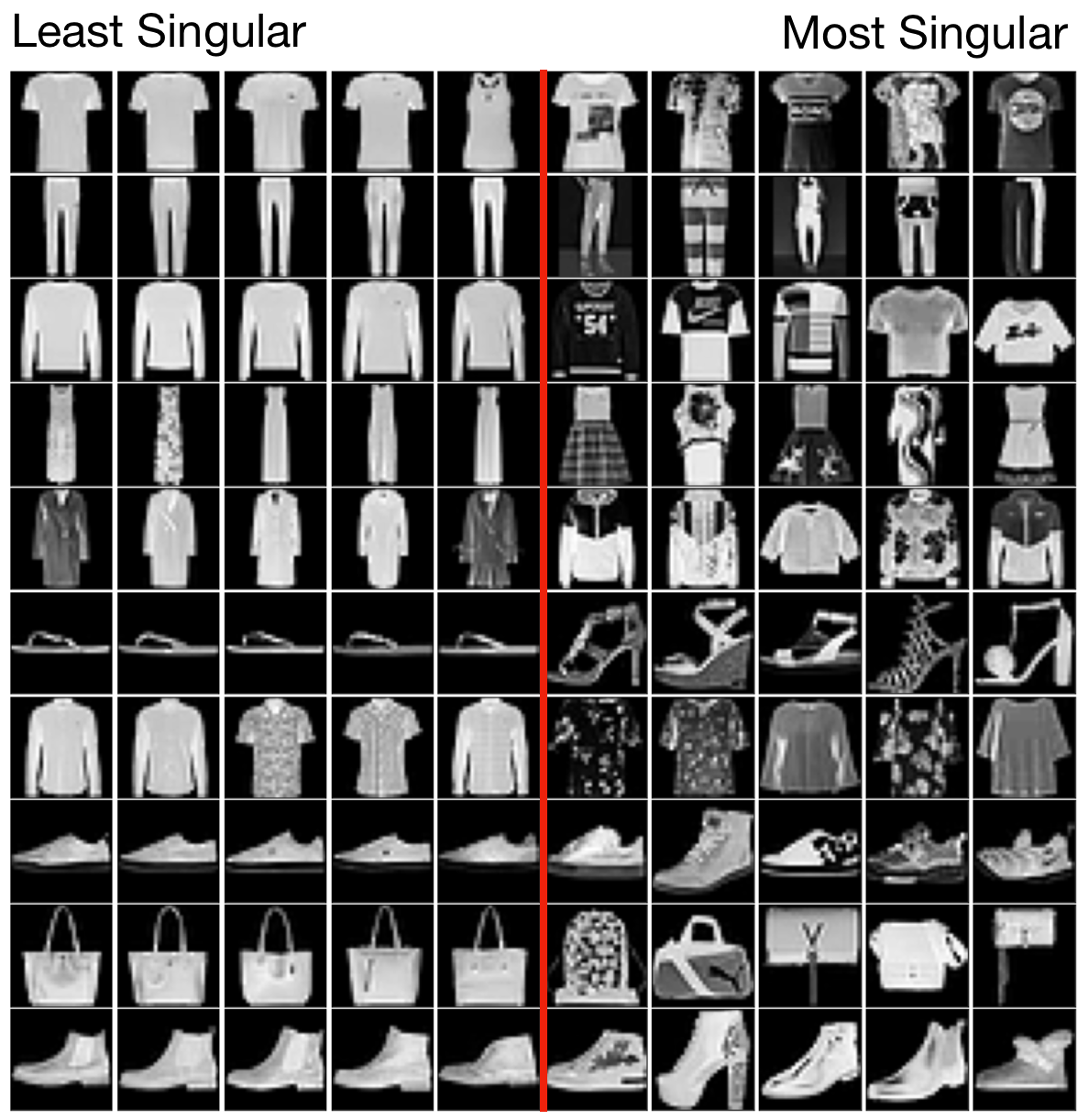}
    \caption[\Hades applied to the Fashion-MNIST dataset]{Images with the lowest singularity scores (left half) and the highest singularity scores (right half), upon applying \Hades to the Fashion-MNIST dataset.}
    \label{fig:real fmnist}
\end{figure}

We applied \Hades on MNIST and Fashion-MNIST datasets on each class of 6,000 images\footnote{Similar results were obtained from running \Hades on the entire dataset of 60,000 datasets.}, and sorted the images according to their singularity scores. Prior to applying \Hades, each 784-dimensional image vector was reduced to 100-dimensional vector by applying Discrete Cosine Transform. Figure \ref{fig:real mnist} (MNIST) and Figure \ref{fig:real fmnist} (Fashion-MNIST) show the result, where the left half of each Figure displays images with the lowest singularity scores and the right half displays images with the highest singularity scores. 

Images on the right half have irregular characteristics when compared to images on the left. This is explained from the fact that \Hades assesses \textit{local uniformity}. Indeed, images on the left look similar to each other, indicating that there are a lot more of similar images of small, subtle variations, thus locally constituting a more uniform distribution with a well-behaved variation. On the other hand, images on the right arise from irregular handwritings and clothing items. This means that there wouldn't be a uniform distribution of similar variations of the images, and thus picked up by \Hades as highly singular. The computation time for running \Hades on 6,000 images corresponding to each digit spanned 30 seconds to 45 seconds.

\pagebreak

\subsection{Anomaly detection}

\begin{figure}[H]
\centering
    \includegraphics[width=0.23\textwidth]{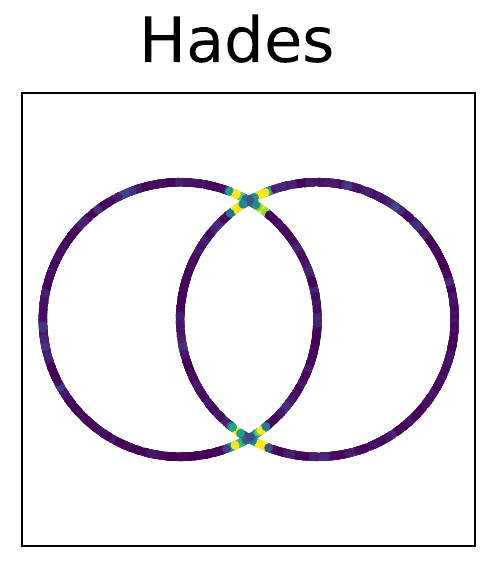}
    \includegraphics[width=0.23\textwidth]{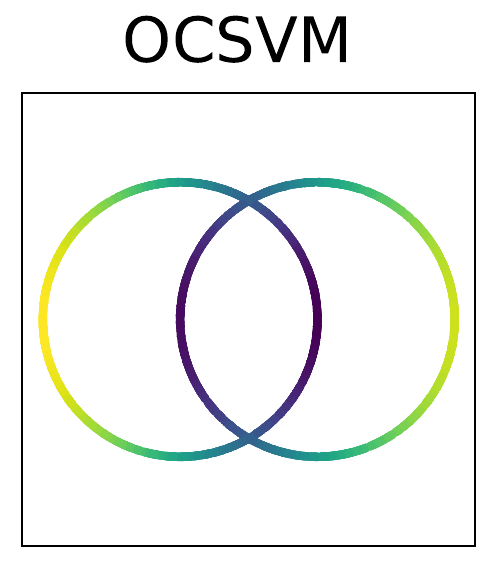}
    \includegraphics[width=0.23\textwidth]{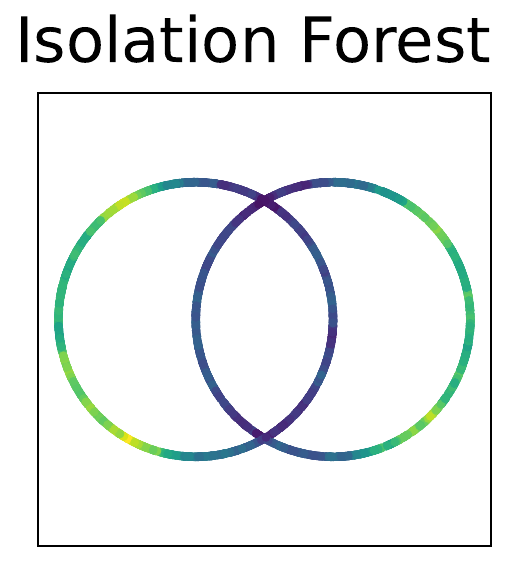}
    \includegraphics[width=0.23\textwidth]{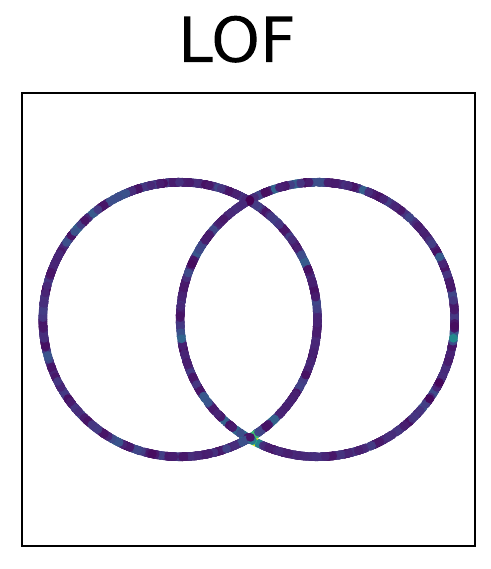}
    \caption[Comparison of \Hades and anomaly detection algorithms]{\Hades is different from the \textit{anomaly detection} algorithms; points marked as highly anomalous are marked in yellow.}
    \label{fig:outlier compare}
\end{figure}

We remark that \Hades has a different objective to existing anomlay detection algorithms. Whereas \Hades detects anomaly in local geometry, existing anomaly detection algorithms detect outliers. Along with \Hades, three anomaly detection algorithms were tesed in Figure \ref{fig:outlier compare} (One-Class SVM \cite{ocsvm}, Isolation Forest \cite{isolation_forest}, Local Outlier Factor \cite{lof}). The points with high anomaly score are marked in yellow (viridis colormap).

\section{Conclusion}
We introduced and studied \Hades, an unsupervised learning algorithm that assigns a singularity score to data points.
This is done by measuring how much the local geometry deviates from a manifold using a goodness-of-fit. 
The strengths of the algorithm are firstly its speed, in particular compared with recent topological approaches, and secondly that it can be seen as first step toward learning the full stratified space. The main disadvantage is that the goodness-of-fit algorithm simply detects what is \textit{not} like a disk, and doesn't give a further details about the local geometry. This is where future research may blossom by using the richer information of local geometry provided by topological methods; for example one may compute persistent homology only at points declared to be singular by \Hades. These research works together aim to create a computational toolbox for modeling general data using stratified spaces.

\pagebreak
\typeout{} %Uncomment when bib not working
\bibliographystyle{abbrv}
\bibliography{refs}

\pagebreak
\appendix
\section*{Appendix}

\section{Experimental details}\label{app:experiments}

We implemented \Hades in Python. Computational experiments were done with a standard laptop: Macbook Pro 2018 with the 2.6 GHz Intel Core i7 processor and the 16 GB 2400 MHz DDR4 memory.

\paragraph{Synthetic data: Low-dimensional.} Details of the datasets used are in Table \ref{table: lowdim}. Hyperparameters were not specified to run each dataset; they were identified automatically by the algorithm. This means that only the point clouds were inputted to \Hades to produce Figure \ref{fig:synthetic lowdim}.

\begin{table}[H]
\centering
\begin{tabular}{lllll} 
\hline
\textbf{Dataset} & \textbf{Description} & \textbf{Size} & \textbf{Time} & \textbf{T/S}  \\ 
\hline\hline
Circle & One circle & 5000 & 48 & 0.010 \\ \hline
Sphere & One 2-dimensional sphere & 10000 & 140 & 0.014 \\ \hline
Two circles & Two intersecting circles & 5000 & 78 & 0.016 \\ \hline
Venus & One disk and two intersecting line segments attached & 10000 & 147 & 0.015 \\ \hline
Two spheres & Two intersecting spheres & 10000 & 195 & 0.020 \\ \hline
Cone & Two joined cones that are cut along top and bottom & 10000 & 181 & 0.018 \\ \hline
Skewer & One saddle surface skewered by a line segment & 10000 & 175 & 0.018 \\ \hline
Pinch torus & One torus, pinched along a neck & 10000 & 184 & 0.018 \\ \hline
\end{tabular}
\vspace{3mm}
\caption{Details of synthetic low-dimensional datasets, with the number of data points (size) and the time taken to run \Hades on each data, and the time taken divided by size, in seconds (T/S).}
\label{table: lowdim}
\end{table}

\paragraph{Synthetic data: High-dimensional.} There are three families of datasets here, constructed for values of $d=1, 2, 3, 4, 5$:
\begin{enumerate}
	\item \textbf{Solid ball.} One $d$-dimensional unit ball (with filled interior). 
    \item \textbf{Two Disks.} Two $2d$-dimensional unit disks intersecting at a $d$-dimensional disk. Constructed by taking two $2d$-dimensional unit disks in $\RR^{3d}$, where the first disk spans axes $1, 2, \ldots 2d$ and the second disk spans axes $d+1, d+2, \ldots 3d$.
    \item \textbf{Two Spheres.} Two $d$-dimensional spheres intersecting at a $(d-1)$-dimensional sphere. Constructed by taking two $d$-dimensional unit spheres in $\RR^{d+1}$, whose centres are spaced apart by distance 1.
\end{enumerate}

To calculate ROC curves, binary labels constituting ground truth are required. We define the ground truth label to depend on the local radius used for neighborhood isolation. This is because singular locus in a stratified space has measure zero, so that there is in fact 0 probability that a randomly sampled point from a stratified space is singular. However, the measure is positive when we thicken the singular locus by a radius, which is relevant to experimental setting where local neighborhoods used for data analysis may intersect the singular locus sufficiently closely. 

We therefore define a binary label on a stratified space $M$ with singular locus $M_{\on{sing}}$ by declaring that $x \in M$ is $s$-close to singularity if the distance from $x$ to $M_{\on{sing}}$ is within $s$. Furthermore, when a local radius parameter $r$ is used for \Hades, we set $s = r/2$, so that a data point is declared singular iff it is within the distance $r/2$ from the singular locus. The scores used for \Hades' classification is $\log(1/p_i)$, where $p_i$ is the goodness-of-fit p-value of the $i$-th data point.

The volume of a $d$-dimensional disk of radius $r$ is $\omega_d r^d$, where $\omega_d$ is a constant. Therefore, when $r < 1$, the volume diminishes exponentially in $d$. To account for this, we increased the radius parameter and the sample size as the dimension increased. We used the radius parameters $r_d = r_0^{1/d}$ and sample sizes $N_d = N_0 \alpha^d$ for constants $r_0, N_0, \alpha$. For the Solid Ball dataset, we used $(r_0, N_0, \alpha) = (0.02, 15000, 1.5)$. For the Two Disks dataset, we used $(r_0, N_0, \alpha) = (0.1, 15000, 1.5)$ and for the Two Spheres dataset, we used $(r_0, N_0, \alpha) = (0.03, 15000, 1.5)$. The threshold parameter $\eta$ was fixed at $\eta = 0.95$.

\paragraph{Synthetic data: Manifold hypothesis.} The manifolds consisted of spheres, ellipsoids, Cartesian products of spheres, and torus. The stratified spaces consisted of unions of two spheres, cones, hollow cubes, and a union of three disks. For each specified sample size $N \in \{1000, 2000, 4000, 8000\}$, 20 copies of each point cloud were randomly generated. Thus for each sample size, a synthetic dataset consisting of 160 point clouds sampled from various manifolds and 120 point clouds sampled from various stratified spaces were generated. Then SUPC, UPUP, KS scores were calculated for each point cloud, and we tested their efficacy in distinguishing manifolds from stratified spaces.

\paragraph{Road networks.} Starting from $1500 \times 1500$ images of aerial photographs of road networks, we extracted pixels that contain non-zero brightness values. Due to the uniform, clean nature of the images, we used fixed hyperparameters $(r, \eta) = (0.012, 0.8)$ (each image was normalised to fit in a unit square). Due to the large number of pixels in the images, only 10\% of the pixels were used for singularity score calculations, and the singularity scores obtained here were extrapolated to the rest of the pixels. This significantly reduced computation time while still cleanly detecting singularities. The number of points contained in the images ranged from 45,000 to 200,000, and the time taken to run \Hades on each image ranged from 6 to 31 seconds.

\paragraph{Cyclo-octane conformation.} The consists of 6040 points on the 24-dimensional space $\RR^{24}$ that parametrises 3D positions of 8 carbon molecules in the cyclo-octane $\text{C}_{8} \text{H}_{16}$. This was taken from the publicly available repository of \cite{gadata}. \Hades was run directly on the 24-dimensional dataset, with the fixed hyperparameters $(r, \eta) = (0.35, 0.95)$. This took 6 seconds to run. 

\paragraph{Image datasets.} We first applied Discrete Cosine Transform to each image and reduced the 784-dimensional image vector into a 100-dimensional vector. This reduces the data dimension whilst retaining shape information of each digit. On this transformed dataset of 100-dimensional vectors, we ran \Hades with the fixed hyperparameters of $(k, \eta) = (200, 0.95)$, where $k$ is the number of nearest neighboring points used for local neighborhood isolation, and $\eta$ is the PCA threshold parameter. Nevertheless the algorithm returns similar results when we change the hyperparameters. \Hades was run separately on each class of images, although we observed similar results when we ran the algorithm on the whole dataset.

\pagebreak
\section{Explicit computation of the kernel MMD}\label{app:mmd_formula}

The content of this section is the proof of Theorem \ref{thm:mmd_eval}, which is the formula we use to evaluate the MMD used in the Uniformity Test.

The following formula holds in general for kernel MMD:
\begin{align*}
    \MMD_\kappa(\mu, \nu) =& \iint \kappa(x,y) \d (\mu - \nu)(x) \d (\mu - \nu)(y) \\
    =& \iint \kappa(x,y)\d\mu(x)\d\mu(y) + \iint \kappa(x,y)\d\nu(x)\d\nu(y) - 2 \iint \kappa(x,y)\d\mu(x)\d\nu(y)
\end{align*}

\begin{prop}
    Suppose $\kappa$ is a kernel on $\RR^d$. Let $U_d = \omega_d^{-1} \mc H^d|_{\mathbb{D}_d}$ be the uniform distirbution over the unit $d$-dimensional ball where $\omega_d = \pi^{d/2} / \Gamma\left(1 + \frac d2 \right)$ is the volume of the unit $d$-dimensional ball. Let $\hat\mu_n = \frac1n (\delta_{X_1} + \cdots + \delta_{X_n})$ be the iid sample drawn uniformly from $U_d$. Then we have the following.
    \begin{align*}
        \MMD_\kappa(\mu, \nu) =& \frac1{\omega_d^2} \bar F_{\kappa, d} + \frac1{n^2} \sum_{i,j} \kappa(x_i, x_j) - \frac2{n\omega_d } \sum_{i=1}^n F_{\kappa, d}(x_i)
    \end{align*}
    where
    \begin{align*}
    F_{\kappa, d}(x) =& \int_{\|y\| \le 1} \kappa(x, y) \d y, \quad \bar F_{\kappa, d} = \int_{\|x\|\le 1} F_{\kappa, d}(x) \d x
    \end{align*}
\end{prop}
\begin{proof}
    We separately evaluate the terms:
    \begin{align*}
        \iint \kappa(x,y)\d U_d(x) \d U_d(y) =& \frac1{\omega_d^2} \iint_{\|x\| \le 1, \|y \| \le 1 } \kappa(x,y) \d x \d y = \frac1{\omega_d^2} \int_{\|x\|\le 1} F_{\kappa, d}(x) \d x \\
        \iint \kappa(x,y)\d \hat \mu_n (x) \d \hat \mu_n (y) =& \frac1{n^2} \sum_{i,j} \kappa(x_i, x_j) \\
        \iint \kappa(x,y)\d \hat \mu_n (x) \d U_d (y) =& \frac1{n\omega_d } \sum_{i=1}^n F_{\kappa, d}(x_i)
    \end{align*}
\end{proof}

The volume of $\SS^{d-1}$ is equal to $d \omega_d$. The volume of a $d$-dimensional ball is equal to $\pi^{d/2} / \Gamma(\frac d2 + 1)$.

\begin{lem}
    Let $\kappa(x,y)$ be a kernel on $\RR^d$ invariant under rotation, i.e. for any orthogonal transform $A \in O(d)$, we have that $\kappa(Ax, Ay) = \kappa(x, y)$. Then whenever $\|x_1 \| = \|x_2 \|$, we have $F_{\kappa, d}(x_1) = F_{\kappa, d}(x_2)$. Furthermore, the following identity holds:
    \[ \bar F_{k, d} = d \omega_d \int_0^1  F_{\kappa, d}(r)  r^{d-1} \d r \]
\end{lem}
\begin{proof}
    Rotational invariance of the unit ball directly implies that $\|x_1 \| = \|x_2 \|$ gives $F_{\kappa, d}(x_1) = F_{\kappa, d}(x_2)$. We also evaluate:
    \begin{align*}
        \bar F_{k, d} =& \int_{\|x\|\le 1} F_{\kappa, d}(x) \d x = \mc H^{d-1}(\mathbb{S}^{d-1}) \int_0^1  F_{\kappa, d}(r)  r^{d-1} \d r = d \omega_d \int_0^1  F_{\kappa, d}(r)  r^{d-1} \d r
    \end{align*}
\end{proof}

We also note the simple expression for the expected value.
\begin{prop}
    We have:
    \[ \EE_{\hat \mu_n} \MMD^2_\kappa(\hat\mu_n, \mu) = \frac1n \cdot \left( \int \kappa(x,x) \d \mu(x) - \iint \kappa(x,y) \d \mu(x) \d \mu(y) \right) \]
\end{prop}
\begin{proof}
    \begin{align*}
         \EE_{\hat \mu_n} \MMD^2_\kappa(\hat\mu_n, \mu) =& \iint \kappa \d x \d y - 2 \iint \kappa \d x \d y + \frac Cn + \frac{n(n-1)}{n^2} \EE \sum_{i \neq j} \kappa(X_i X_j) \\
         =& \frac Cn - \frac1n \EE \iint \kappa \d x \d y
    \end{align*}
\end{proof}

\begin{prop}
    For $k \ge 0$, let $\kappa(x,y) = \langle x,y\rangle^k$ be a monomial kernel. Then we have that:
    \begin{align*}
        F_{\kappa, d}(r) =& \omega_{d-1} \on{B}\left( \frac{k+1}2, \frac{d+1}2 \right) = \frac{\pi^{(d-1)/2} \Gamma(\frac{k+1}2)}{\Gamma(\frac{k+d}2 + 1)} \cdot r^k & \text{, if $k$ is even}\\
        \bar F_{\kappa, d} =& \frac d{d+k} \omega_d \omega_{d-1} \on{B}\left( \frac{k+1}2, \frac{d+1}2 \right) = \frac{2^d \pi^{d-1}}{(d-1)!} \cdot \frac1{d+k} \on{B}\left( \frac{k+1}2, \frac{d+1}2 \right) & \text{, if $k$ is even}
    \end{align*}
    and both expressions are zero if $k$ is odd.
\end{prop}
\begin{proof}
    We directly evaluate:
    \begin{align*}
        F_{\kappa, d}(r) =& \int_{\|y\|\le 1} \kappa(r \cdot e_1,y)^k \d y \\
        =& \int_{-1}^{+1} \int_{\| z \| \le \sqrt{1-s^2}, z \in \RR^{d-1}} (rs)^k \d z \d s \\
        =& \omega_{d-1} r^k \int_{-1}^{+1} s^k (1-s^2)^{(d-1)/2} \d s
    \end{align*}
    By symmetry, odd $k$ implies that $F_{\kappa, d}(r) = 0$. For even $k$, we may write:
    \begin{align*}
        \int_{-1}^{+1} s^k (1-s^2)^{(d-1)/2} \d s =& 2 \int_{0}^{1} (s^2)^{k/2} (1-s^2)^{(d-1)/2} \d s \\
        =& 2 \int_{0}^{1} t^{k/2} (1-t)^{(d-1)/2} (2\sqrt{t})^{-1} \d t \\
        =& \int_{0}^{1} t^{(k-1)/2} (1-t)^{(d-1)/2} \d t \\
        =& \on{B}\left( \frac{k+1}2, \frac{d+1}2 \right)
    \end{align*}
    where $\on{B}(u,v) = \int_0^1 t^{u-1}(1-t)^{v-1} \d t = \Gamma(u)\Gamma(v)/\Gamma(u+v)$ is the Beta function. Thus for even $k$ we get that:
    \[ F_{\kappa, d}(r) / r^k = \omega_{d-1} \on{B}\left( \frac{k+1}2, \frac{d+1}2 \right) = \frac{\pi^{(d-1)/2}}{\Gamma(1 + \frac{d-1}2)} \frac{\Gamma(\frac{k+1}2) \Gamma( \frac{d+1}2 )}{\Gamma(\frac{k+d}2 + 1  )} = \frac{\pi^{(d-1)/2} \Gamma(\frac{k+1}2) }{\Gamma(\frac{k+d}2 + 1)} \]
    For $\bar F_{\kappa, d}$ and even $k$,
    \begin{align*}
        \bar F_{\kappa, d} =& d \omega_d \int_0^1  F_{\kappa, d}(r)  r^{d-1} \d r \\
        =& d\omega_d \omega_{d-1} \on{B}\left( \frac{k+1}2, \frac{d+1}2 \right) \int_0^1 r^{k+d-1} \d r \\
        =& \frac{d}{d+k} \cdot \omega_d \omega_{d-1} \cdot \on{B}\left( \frac{k+1}2, \frac{d+1}2 \right)
    \end{align*}
    We note that:
    \[ \omega_d \omega_{d-1} = \frac{\pi^{d - (1/2)}}{\Gamma(\frac {d+2}2) \Gamma(\frac {d+1}2)} = \frac{\pi^{d - (1/2)}}{2^{1-(d+1)} \sqrt{\pi} \Gamma(d+1)} = \frac{2^d \pi^{d-1} }{d!} \]
    where we used the Lagrange duplication formula $\Gamma(z) \Gamma(z+\frac12 ) = 2^{1-2z} \sqrt{\pi} \Gamma(2z)$ for $z = (d+1)/2$. This gives:
    \[ \bar F_{\kappa, d} = \frac{2^d \pi^{d-1}}{(d-1)!} \cdot \frac1{d+k} \on{B}\left( \frac{k+1}2, \frac{d+1}2 \right) \]

\end{proof}

\begin{prop}
    Let $\kappa(x,y) = \sum_{k=0}^\infty a_k \langle x,y \rangle^k$ be a power series kernel with $a_k \ge 0$. Then we have that:
    \begin{align*}
        F_{\kappa, d}(r) =& \omega_{d-1} \sum_{k=0}^\infty \on{B}\left( k+\frac12, \frac{d+1}2 \right) a_{2k}r^{2k} \\
        =& \pi^{(d-1)/2} \sum_{k=0}^\infty \frac{\Gamma(k + \frac{1}2)}{\Gamma(k + \frac{d}2 + 1)} \cdot a_{2k} r^{2k} \\
        \bar F_{\kappa, d} =& d\omega_d \omega_{d-1} \sum_{k=0}^\infty \on{B}\left( k+\frac12, \frac{d+1}2 \right) \frac{a_{2k}}{d+2k} \\
        =& \frac{2^d \pi^{d-1}}{(d-1)!} \sum_{k=0}^\infty \on{B}\left( k + \frac12, \frac{d+1}2 \right) \frac{a_{2k}}{d+2k} 
    \end{align*}
\end{prop}
\begin{proof}
    This is proven by interchanging sum and integral, using the previous proposition for monomial kernel, and noting that only the even indexed terms survive. The exchange of sum and integral is justified because of absolute convergence, following from $a_k \ge 0$.
\end{proof}

We now prove Theorem \ref{thm:mmd_eval}.

\begin{thm}\label{thm:mmd_eval2}
    Let $\hat \mu_n = \frac1n(\delta_{x_1} + \cdots + \delta_{x_n})$ be a discrete (non-random) measure and let $\unif_d$ be the uniform distribution over the unit $d$-dimensional disk in $\RR^d$. Let $\kappa$ be a kernel given by $\kappa(x,y) = \sum_{k=0}^\infty a_k \langle x,y\rangle^k$. Then we have:
    \begin{align*}
        \MMD^2_\kappa(\hat\mu_n, \unif_d) = \frac1{n^2}\sum_{i=1}^n \sum_{j=1}^n \kappa(x_i, x_j) + \sum_{k=0}^\infty a_{2k} \beta_{d, k} \left( \frac d{d+2k} - \frac2n \sum_{i=1}^n \|x_i\|^{2k} \right)
    \end{align*}
    where
    \begin{align*}
        \beta_{d, k} = \frac1{\sqrt \pi} \frac{\Gamma(\frac d2 + 1) \Gamma(k + \frac12)}{\Gamma(k + \frac d2 + 1)}
    \end{align*}
    and $\Gamma(t)$ is the Gamma function.
\end{thm}
\begin{proof}
    The claim can be restated as $\MMD_\kappa(\hat\mu_n, U_d) = A + B - C$, where 
    \begin{align*}
        A =& \frac 1{n^2}\sum_{i,j}\kappa(x_i, x_j) \\
        B =& \frac{\Gamma(\frac d2 + 1)}{\sqrt \pi} \sum_{k=0}^\infty  \frac{\Gamma(k+\frac12) a_{2k}}{\Gamma(k+\frac d2 + 1)} \frac d{d+2k} \\
        C =& \frac2n \cdot \frac{\Gamma(\frac d2 + 1)}{\sqrt \pi} \sum_{k=1}^n \frac{\Gamma(k+\frac12) a_{2k}}{\Gamma(k+\frac d2 + 1)}  \sum_{i=1}^n \|x_i\|^{2k}
    \end{align*}
    We firstly know that:
    \[ \MMD_\kappa(\hat\mu_n, U_d) = A + B - C \]
    \begin{align*}
        A =& \frac1{n^2} \sum_{i,j} \kappa(x_i, x_j) \\
        B =& \frac{\omega_{d-1}}{\omega_d} \sum_{k=0}^\infty \frac d{d+2k} \on{B}\left( k+\frac12, \frac{d+1}2 \right) a_{2k} \\
        C =& \frac 2n \frac{\omega_{d-1}}{\omega_d} \sum_{k=1}^n \on{B}\left( k+\frac12, \frac{d+1}2 \right) a_{2k} \sum_{i=1}^n \|x_i\|^{2k}
    \end{align*}
    We compute:
    \begin{align*}
        B =& \frac{\omega_{d-1}}{\omega_d} \sum_{k=0}^\infty \frac d{d+2k} \on{B}\left( k+\frac12, \frac{d+1}2 \right) a_{2k} \\
        =& \pi^{-1/2} \frac{\Gamma(\frac{d+2}2)}{\Gamma(\frac{d+1}2)} \sum_{k=0}^\infty \frac d{d+2k} \frac{\Gamma(k+\frac12) \Gamma(\frac{d+1}2)}{\Gamma(k+\frac d2 + 1)} a_{2k} \\
        =& \pi^{-1/2} \Gamma\left(\frac d2 + 1\right) \sum_{k=0}^\infty \frac d{d+2k} \frac{\Gamma(k+\frac12) a_{2k}}{\Gamma(k+\frac d2 + 1)} 
    \end{align*}
    and
    \begin{align*}
        C =& \frac 2n \frac{\omega_{d-1}}{\omega_d} \sum_{k=1}^n \on{B}\left( k+\frac12, \frac{d+1}2 \right) a_{2k} \sum_{i=1}^n \|x_i\|^{2k} \\
        =& \frac 2n \pi^{-1/2} \Gamma\left( \frac d2 + 1 \right) \sum_{k=1}^n \frac{\Gamma(k+\frac12) a_{2k}}{\Gamma(k+\frac d2 + 1)}  \sum_{i=1}^n \|x_i\|^{2k}
    \end{align*}
\end{proof}

\paragraph{Error rate.} The MMD expression above involves the following infinite series:
\begin{align*}
    B =& \frac{\Gamma(\frac d2 + 1)}{\sqrt \pi} \sum_{k=0}^\infty  \frac{\Gamma(k+\frac12) a_{2k}}{\Gamma(k+\frac d2 + 1)} \frac d{d+2k} \\
    C =& \frac2n \cdot \frac{\Gamma(\frac d2 + 1)}{\sqrt \pi} \sum_{k=1}^n \frac{\Gamma(k+\frac12) a_{2k}}{\Gamma(k+\frac d2 + 1)}  \sum_{i=1}^n \|x_i\|^{2k}
\end{align*}
Up to constant, the series are:
\begin{align*}
    \sum_{k=0}^\infty  \frac{\Gamma(k+\frac12) a_{2k}}{\Gamma(k+\frac d2 + 1)} \frac d{d+2k}, \quad \sum_{k=1}^\infty \frac{\Gamma(k+\frac12) a_{2k}}{\Gamma(k+\frac d2 + 1)}  \sum_{i=1}^n \|x_i\|^{2k}
\end{align*}
Observe that the summands $d/(d+2k)$ and $\sum_i \|x_i\|^{2k}$ are non-increasing in $k$. Also, we substitute in $a_{2k} = \gamma^{2k}$. We are interested in the relative error of estimation, so that we are then further simply interested in:
\begin{align*}
    \sum_{k=0}^\infty  \frac{\Gamma(k+\frac12) \gamma^{2k}}{\Gamma(k+\frac d2 + 1)}
\end{align*}
At $d=1$, this is
\begin{align*}
    \sum_{k=0}^\infty  \frac{\Gamma(k+\frac12) \gamma^{2k}}{\Gamma(k+\frac 32)} \le \sum_{k=1}^\infty \frac{\gamma^{2k}}k = -\log(1-\gamma^2)
\end{align*}
Thus a crude error bound is given by considering the convergence rate of the function $\log(1-\gamma^2)$. 

By direct evaluation, the evaluation up to $k=10$ of the Taylor series and $\gamma \le 0.9$ gives relative error $\le 0.03$ and $\gamma \le 0.5$ gives relative error $\le 10^{-6}$.

\pagebreak
\section{Proof of the theoretical guarantee} \label{app:guarantee}

This section, spanning across the rest of the paper, contains the proof of our main theoretical guarantee, Theorem \ref{thm:main}. We develop several technical tools step-by-step and proceed towards the final subsection. As such, the reader is advised to start from the last subsection and work backwards to understand details of the proof.

\subsection{Wasserstein inequalities}

We prove several inequalities involving the Wasserstein distance.

\paragraph{Concentration}

In this subsection we study how the empirical measure approximates the underlying measure, in the sense of Wasserstein distance. The main objective of this subsection is Proposition \ref{main_wass_conc}, which is derived by simplifying Corollary 1.2 from \cite{boissard_le_gouic}. We use the notion of \textit{covering number} for this:
\[N_{\on{cover}}(M, r) = \min \bigg \{ m \>\bigg |\> \exists x_1, \ldots x_m \in M, \> \cup_{i=1}^m \Ball(x_i, r) \supseteq M \bigg \}\]
\begin{thm}[Boissard-Le Gouic]
    Let $(M, d, \mu)$ be a measured Polish space of a finite diameter $R$. Suppose that there exist $\alpha > 2p, \beta > 0$ so that the following holds for for every $0 < r < R/4$:
    \begin{align*}
        & N_{\on{cover}}(M, r) \le \beta \left (\frac R r \right)^\alpha
    \end{align*}
    Then the following holds:
    \begin{align*}
        \EE \bigg[ \Wass_p(\hat \mu_m, \mu) \bigg ] \le \frac{64R}3 \cdot \left(\frac{ 2p }{\alpha - 2p} \right)^{2p/\alpha} \cdot  \left( \frac \beta m \right)^{1/\alpha}
    \end{align*}
\end{thm}

To apply this to compact subsets of a Euclidean space, we use a lemma from \cite{bartlett_notes}:
% \footnote{It appears in Lecture 12 of Peter Barlett's Berkeley lecture slides on theoretical statistics (find a better reference, \url{https://www.stat.berkeley.edu/~bartlett/courses/2013spring-stat210b/})}:
\begin{lem}
    The $D$-dimensional unit ball $\Ball_D$ satisfies:
    \begin{align*}
        N_{\on{cover}}(\Ball_D, r) \le (1 + 2r^{-1})^D 
    \end{align*}
\end{lem}
\begin{proof}
    It is easy to see that a \textit{maximal packing} by $N'$ balls of radii $r/2$ is also a covering by balls of radii $r$\footnote{Suppose that balls of radii $r/2$ centered at $x_1, \ldots x_{N'}$ is a maximal packing, but it's not a covering if we chose radii $r$. Then there exists a point $y$ that is away by the distance $r$ from $x_1, \ldots x_{N'}$, which means that balls of radii $r/2$ centered at $N'+1$ points $\{y, x_1, \ldots x_{N'}\}$ is also a packing. This contradicts maximality.}, and so we have $N_{\on{cover}}(\Ball_D, r) \le N'$. Now consider a maximal packing by balls of radii $r/2$ centered at $x_1, \ldots x_{N'}$. Then,
    \begin{align*}
        & \cup_i \Ball(x_i, r/2) \subseteq (1 + r/2) \cdot \Ball_D \\
        \implies & N' \cdot (r/2)^D \le (1 + r/2)^D
    \end{align*}
\end{proof}

\begin{cor}\label{wass_exp_cor}
    Let $\mu$ be a Borel probability measure valued in $\RR^D$ whose support has the diameter of $R$, and suppose $D \ge 3$. Then we have:
    \begin{align*}
        & \EE \bigg[ \Wass(\hat \mu_m, \mu) \bigg ] \le \frac{c_1}{m^{1/D}} \le \frac{c_2}{m^{1/D}} \\
        \text{where } & c_1 = 32 R \cdot \left( \frac{2}{D-2}\right)^{2/D}, \> c_2 = 51 R
    \end{align*}
    Also, if $D \ge 4$, we may take $c_2 = 32R$. 
\end{cor}
\begin{proof}
    Whenever $r < R/2$, the following holds:
    \begin{align*}
        & N_{\on{cover}}(M, r) \le N_{\on{cover}}( (R/2) \cdot \Ball_D, r) = N_{\on{cover}}(\Ball_D, 2r/R) \le \left( 1 + \frac{2}{2r/R} \right)^D \le (1.5 R / r)^D
    \end{align*}
    (The assumption $r < R/2$ is only used in the last inequality above) Thus we may apply the previous theorem by taking $\alpha = D$, $\beta = 1.5^D$, and $p=1$, from which we get that:
    \[ \EE \bigg[ \Wass(\hat \mu_m, \mu) \bigg ] \le 32 R \cdot f(D/2) \cdot m^{-1/D} \text{, where } f(t) = (t-1)^{-1/t} \]
    The derivative of $f(t)$ has the same sign as $(t-1)\log(t-1) - t$\footnote{The derivative is $\frac{(t-1)\log(t-1) - t}{t^2 \cdot (t-1)^{1 + 1/t}}$.}, which is an increasing function that takes a zero value at some $t \in (4.5, 5)$ and nowhere else. Thus $f(t)$ at $[1.5, \infty)$ is bounded above by $f(1.5) = 2^{2/3} \le 1.6$ and the limit value of $f$ at infinity, which is $\le 1$; we have $\lim_{t \rightarrow \infty} (t-1)^{1/t} \le \lim_{t \rightarrow \infty} t^{1/t} = \exp(\lim_{t\rightarrow \infty} (\log t)/t ) = 1$. Therefore, we get:
    \[ 32R \cdot f(D/2) \cdot m^{-1/D} \le 32 R \cdot 2^{2/3} \cdot m^{-1/D} \le 51 R \cdot m^{-1/D} \]
    Note also that $f(2) = 1$, so that $D \ge 4$ implies the tighter bound.
\end{proof}

To obtain a concentration inequality, we use the Proposition A2 from \cite{boissard_le_gouic}:
\begin{prop}\label{boiss_a2}
    Let $(E, d, \mu)$ be a measured Polish metric space of a finite diameter $R$ and suppose that $\mu$ has a finite $p$-th moment. Then we have:
    \begin{align*}
        & \Prob\bigg( \Wass_p(\hat\mu_m, \mu) \ge t + \EE[\Wass_p (\hat\mu_m, \mu) ] \bigg) \le \exp\left( - \frac{ m t^{2p} }{2R^{2p} } \right)
    \end{align*}
\end{prop}

Combining the above with Corollary \ref{wass_exp_cor}, we obtain that:
\begin{prop}[Global concentration]\label{wass vanilla}
	Let $\mu$ be a Borel probability measure valued in $\RR^D$ whose support has the diameter of $R$, and suppose $D \ge 3$. For any $t > 0$, the following holds whenever $m \ge f(t)$:
	\begin{align*}
		\Prob \bigg( \Wass(\hat \mu_m, \mu) \ge t  \bigg) \le \exp\left( - \frac{mt^2}{8R^2} \right)
	\end{align*}
	where $f(t) = (102 R/t)^D$.
\end{prop}
\begin{proof}
	Follows directly by using Proposition \ref{boiss_a2} and letting the expected value of the Wasserstein distance be $\le t/2$ in Corollary \ref{wass_exp_cor}.
\end{proof}

We modify the above Proposition to study local behaviour of measures.

\begin{prop}[Local concentration]\label{wass fixed U}
	Let $\mathbf X = (X_1, \ldots X_m)$ be an i.i.d. sample drawn from $\mu \in \mc P(\RR^D)$, where $D \ge 3$. Let $\hat \mu_m = \frac1m \sum_i \delta_{X_i}$ be the empirical measure constructed from $\mathbf X$. Let $U \subseteq \RR^D$ be a Borel set which is contained in a ball of radius $r$. Denote $u = \mu(U)$. For any error level $\epsilon > 0$, the following holds whenever $m \ge \max(N, 2u^{-1} )$:
	\[ \Prob\bigg( \Wass(\hat \mu_m |_U, \mu|_U)  \ge t \bigg) \le c\cdot m^N \gamma^m \]
where
\begin{align*}
	c = \left( \frac u{1-u} \right)^N, \quad N = \lceil (204r/t)^D \rceil, \quad \gamma = 1 - u (1- \exp(-t^2/8r^2))
\end{align*}
In particular, $\gamma \in (0,1)$ and the probability of error decays exponentially in $m$.
\end{prop}
\begin{proof}
    We condition over points of $\mathbf X$ falling into $U$. Denote by $\mc{S}_I$ the event $(X_i \in U \iff i \in I)$, $|I|$ for the cardinality of each $I$, and $u := \mu (U)$, we have
    \begin{align}
    	\Prob\bigg( \Wass(\hat \mu_m |_U, \mu|_U)  \ge t \bigg)
    	&= \sum_{I \subseteq \{1, \ldots m\} } \Prob(\mc S_I) \cdot \Prob\bigg( \Wass(\hat \mu_m |_U, \mu|_U)  \ge t \>\bigg|\> \mc S_I \bigg) \nonumber \\
    	&= \sum_{I \subseteq \{1, \ldots m\} } u^{|I|} (1-u)^{m - |I|} \Prob\bigg( \Wass(\hat \mu_m |_U, \mu|_U)  \ge t \>\bigg|\> \mc{S}_I \bigg) \nonumber \\
    	&= \sum_{k=0}^m \binom m{k} u^k (1-u)^{m-k}  \Prob\bigg( \Wass(\hat \mu_m |_U, \mu|_U)  \ge t \>\bigg|\> \mc{S}_{\{1, \ldots k\} }\bigg) \label{local conc eq 1}
    \end{align}
    Now we apply Proposition \ref{wass vanilla} to the conditional probabilities above. This only applies to $k \ge N = \lceil (204r/t)^D \rceil$, thus we split the sum for $k < N$ and $k \ge N$. Writing $\xi = \exp(-t^2/8r^2)$, Equation \eqref{local conc eq 1} is thus bounded by:
    \begin{align*}
        \le & \sum_{k=0}^{N-1} \binom m{k} u^k (1-u)^{m-k} + \sum_{k=N}^{m} \binom m{k} u^k (1-u)^{m-k} \xi^k  \\
        \le & (1-u)^m \cdot \sum_{k=0}^{N-1} \left( \frac{m u}{1-u} \right)^k + \sum_{k=0}^m \binom mk (u \xi) ^k (1-u)^{m-k} \\
        = & (1-u)^m \frac{\left( \frac{mu}{1-u} \right)^N - 1}{\frac{mu}{1-u} - 1} + (1 - u + u \xi)^m \\
        \le & (1-u)^{m} \bigg( \bigg( \frac{mu}{1-u} \bigg)^N -1 \bigg)+ (1 - u + u \xi)^m \\
        \le & \bigg( \frac{mu}{1-u} \bigg)^N \cdot (1 - u + u \xi)^m
    \end{align*}
    where in the second to last inequality we used the assumption $mu/(1-u) \ge 2$ and in the last inequality we used $(1-u)^m \le (1-u + u\xi)^m$.
\end{proof}

We further modify the above into a simultaneous concentration inequality, which is the main result of the section.

\begin{prop}[Local simultaneous concentration]\label{main_wass_conc}
    Let $\mathbf X = (X_1, \ldots X_m)$ be an i.i.d. sample drawn from $\mu \in \mc P(\RR^D)$, where $D \ge 3$. Let $\hat \mu_m = \frac1m \sum_i \delta_{X_i}$ be the empirical measure constructed from $\mathbf X$. Also let $r, t > 0$ and $U_i = \Ball(X_i, r) \backslash \{X_i\}$. Then the following holds whenever $m \ge \max(N, 2/u_-)$:
	\[ \Prob  \bigg( \max_i \Wass(\hat \mu_m|_{U_i}, \mu|_{U_i}) \le t \bigg) \ge 1 - \delta_m \]
	where $\lim_{m \rightarrow \infty} \delta_m = 0$ exponentially fast, given explicitly as:
	\[ \delta = c \cdot m^{N+1} \gamma^m \]
    where
    \begin{align*}
        & c = \bigg(\frac{u_+}{1 - u_+}\bigg)^N, \quad N = \bigg\lceil\bigg( \frac{204r}{t} \bigg)^D\bigg\rceil, \quad \gamma = 1 - u_-(1 - \xi), \quad \xi = \exp\bigg( \frac{-t^2}{8r^2} \bigg) \\
        & \quad u_- = \inf_{x \in \on{supp}\mu} \mu(\Ball(x, r)), \quad u_+ = \sup_{x \in \on{supp}\mu} \mu(\Ball(x, r))
    \end{align*}
\end{prop}
\begin{proof}
	We use union bound for different $i=1, \ldots m$. Let $\mu^m = \mu \times \cdots \times \mu$ be the product measure on $(\RR^D) \times \cdots \times (\RR^D)$. Define the set $E_i \subseteq (\RR^D)^m$ as the set where the exception event occurs for $U_i$:
    \[ E_i = \bigg\{ \mathbf x = (x_1, \ldots x_m) \bigg| \Wass(\delta_{\mathbf x}|_{V_i}, \mu|_{V_i}) \ge t \bigg\} \text{, where } V_i = \Ball(x_i, r) \]
    Then we have:
    \[  \Prob  \bigg( \max_i \Wass(\hat \mu_m|_{U_i}, \mu|_{U_i}) \le t \bigg) = 1 - \mu^m(E_1 \cup \cdots \cup E_m) \]
    We then apply the union bound:
	\begin{align*}
		\mu^m(E_1 \cup \cdots \cup E_m) \le & \mu^m(E_1) + \cdots + \mu^m(E_m) \\
        = & m \cdot \int \mu^{m-1}\bigg( (x_2, \ldots x_m) \bigg| (x_1, x_2, \ldots x_m) \in E_1 \bigg) \d x_1 \\
		\le & m \cdot \int \bigg(\frac{u_x}{1-u_x}\bigg)^N (m-1)^N \gamma_x^{m-1} \d x
	\end{align*}
	where $u_x = \mu(\Ball(x,r))$ and $\gamma_x = 1 - u_x (1 - \xi)$. Then $u_x \le u_+$ and $\gamma_x \le \gamma$, so that we have:
    \[ \mu^m(E_1 \cup \cdots \cup E_m) \le c \cdot m^{N+1} \gamma^{m} \]
    and the claim is shown.
\end{proof}

\paragraph{Lipschitz continuity}

\begin{lem}\label{mproj same}
    Let $\mu_1, \mu_2 \in \mc P(\RR^D)$ and $\pi \in \on{Gr}(k, D)$. Denoting $\mu_i' = \Pi(\mu_i, \pi)$, the following holds:
    \begin{align*}
        \Wass(\mu_1', \mu_2') \le \Wass(\mu_1, \mu_2)
    \end{align*}
\end{lem}
\begin{proof}
    For every transportation plan from $\mu_1$ to $\mu_2$, we can construct a less costly transportation plan from $\Pi(\mu_1, \pi)$ to $\Pi(\mu_2, \pi)$ by simply pushforwarding across projection. 

    Denote by $\mu_1^\perp = \Pi(\mu_1, \pi)$ and similarly $\mu_2^\perp$. Denote by $p_\pi$ the orthogonal projection map to $\pi$. By definition we have $\mu_1^\perp(U) = \mu(p_\pi^{-1} U)$ for every open $U \subseteq \pi$ and similarly for $\mu_2^\perp$.
    
    Given $\mu_{12}$, a coupling of $\mu_1$ and $\mu_2$, we may define $\mu_{12}^\perp$ as the pushforward along $p_\pi \times p_\pi$, as follows:
    \[\mu_{12}^\perp(U \times V) = \mu_{12}(p_\pi^{-1} U \times p_\pi^{-1} V) \]
    for each open $U, V \subseteq \pi$. $\mu_{12}^\perp$ is a coupling of $\mu_1^\perp, \mu_2^\perp$ because:
    \[ \mu_{12}^\perp(U \times \pi) = \mu_{12}(p_\pi^{-1} U \times p_\pi^{-1} \pi) = \mu_{12}(p_\pi^{-1} U \times \RR^D) = \mu_1(p_\pi^{-1} U) = \mu_1^\perp(U) \]
    and similarly for $\mu_2$. Now,
    \begin{align*}
        \int_{\pi \times \pi} \|x-y\| \d \mu_{12}^\perp (x, y) = \int_{\RR^D \times \RR^D} \| p_\pi(x)-p_\pi(y)\| \d \mu_{12}(x, y) \le \int_{\RR^D \times \RR^D} \|x-y\| \d \mu_{12}(x,y)
    \end{align*}
    where the first equality is due to the general fact that, for $f: X \rightarrow Y$,
    \[ \int_Y \phi(y) \d f_* \mu(y) = \int_X \phi(f(x)) \d \mu(x) \]
    where in our case, $\phi(x,y) = \|x-y\|$, $f(x,y) = (p_\pi(x), p_\pi(y))$, $\mu = \mu_{12}$, and $f_* \mu = \mu_{12}^\perp$.
\end{proof}

\begin{lem}\label{mproj tilt}
    Let $\mu \in \mc P(\RR^D)$ and $\pi_1, \pi_2 \in \on{Gr}(k, D)$. Assume that the support of $\mu$ is bounded in the ball of radius $1$, centered at the origin. Denoting $\mu_i = \Pi(\mu, \pi_i)$, we have:
    \begin{align*}
        \Wass(\mu_1, \mu_2) \le \sqrt{\sin^2 \theta_1 + \cdots + \sin^2 \theta_d}
    \end{align*}
    where $(\theta_1, \ldots \theta_d)$ are the principal angles between $(\pi_1, \pi_2)$. 
\end{lem}
\begin{proof}
    Denote the orthogonal projection map to $\pi_i$ by $p_i$. We define a coupling $\mu_{12}$ of $(\mu_1, \mu_2)$:
    \[ \mu_{12}(U \times V) = \mu(p_1^{-1} U \cap p_2^{-1} V) \]
    It is a coupling since $\mu(U \times \RR^D) = \mu(p_1^{-1} U \cap \RR^D) = \mu(p_1^{-1} U) = \mu_1(U)$ and similarly for $\mu_2$.
    
    For each $x$, consider the following sets:
    \begin{align*}
        S_x = p_2(p_1^{-1}(x) \cap \mc B_1), \quad & S = \{ (x, y) \>|\> x\in \pi_1, y \in S_x \} \subseteq \pi_1 \times \pi_2 \\
        S_x' = (\pi_2 \cap \mc B_1) \backslash S_x, \quad & S' = \{ (x,y) \>|\> x\in\pi_1,  y \in S_x' \} \subseteq \pi_1 \times \pi_2
    \end{align*}
    where $\mc B_1 = \mc B(0, 1) \subseteq \RR^D$ is the unit ball centered at origin. Also let $\theta = \angle(\pi_1, \pi_2)$. We claim that $\mu_{12}|_{S'} \equiv 0$ and $S_x \subseteq \mc B(x, \sin \theta)$. The proposition follows from these assumptions:
    \[ \int_{\pi_1 \times \pi_2} \|x-y\| \d \mu_{12}(x,y) = \int_{S} \|x-y\| \d \mu_{12}(x,y) \\
    \le  \int_{S} \sin \theta \d \mu_{12}(x,y) \\
    = \sin \theta \]
    It remains to prove the postponed claims. First we show $\mu_{12}|_{S'} \equiv 0$. Suppose $U \times V \subseteq S'$. By definition of $S'$, for each $(x, y) \in U \times V$, we have $y \notin p_2(p_1^{-1}(x) \cap \Ball_1)$, i.e. $p_1^{-1}(x) \cap p_2^{-1}(y) \cap \Ball_1 = \emptyset$. Therefore we have $p_1^{-1}(U) \cap p_2^{-1}(V) \cap \Ball_1 = \emptyset$. Since the support of $\mu$ is in $\Ball_1$, we have that $\mu_{12}(U \times V) = \mu(p_1^{-1}(U) \cap p_2^{-1}(V)) = \mu(\emptyset) = 0$. Therefore $\mu_{12}|_{S'} \equiv 0$.

    Now we show $S_x \subseteq \mc B(x, \sin \theta)$. Suppose $y \in S_x = p_2(p_1^{-1}(x) \cap \Ball_1)$, so that there is $z \in \Ball_1$ such that $p_1(z) = x, p_2(z) = y$. Let $d_0 = \dim (\pi_1\cap \pi_2)$. Define $\pi_i^\perp \subseteq \pi_i$ to be the orthogonal complement of $\pi_1 \cap \pi_2$, so that $\pi_i = \pi_i^\perp + (\pi_1 \cap \pi_2)$ is an orthogonal decomposition for $i=1, 2$. By Corollary \ref{two subspace standard matrix}, we obtain an orthonormal basis $\{ u_1, \ldots u_{d_0} \} \cup \{v_1, w_1, \ldots v_{d-d_0}, w_{d-d_0}\}$ of $\on{span}(\pi_1, \pi_2)$ so that:
    \begin{align*}
        \pi_1 \cap \pi_2 =& \on{span}(u_1, \ldots u_{d_0}) \\
        \pi_1^\perp =& \on{span}(v_1, \ldots v_{d-d_0}) \\
        \pi_2^\perp =& \on{span}(v_1', \ldots v_{d-d_0}') \\
        \text{where } v_i' =& (\cos \theta_i) v_i + (\sin \theta_i) w_i
    \end{align*}
    with $(\theta_1, \ldots \theta_{d-d_0})$ being the nonzero principal angles of $(\pi_1, \pi_2)$. We now attempt to understand $(x, y)$ through their 2-dimensional projections. For each $i$, define $\rho_i = \on{span}(v_i, w_i)$ and $z_i = \Pi(z, \rho_i)$. Then for any $u \in \rho_i$, we have that $\Pi(z, u) = \Pi(z_i, u)$ \footnote{More generally, for any pair of subspaces $\pi' \subseteq \pi$, we have that $\Pi(\Pi(z, \pi), \pi') = \Pi(z, \pi')$, as it can be checked by directly writing down the projection matrices.}. This gives an orthogonal decomposition:
    \begin{align*}
        x =& \Pi(z, (\pi_1 \cap \pi_2) + \pi_1^\perp) = \Pi(z, \pi_1 \cap \pi_2) + \sum_{i=1}^{d-d_0} \Pi(z_i, v_i)  \\
        y =& \Pi(z, (\pi_1 \cap \pi_2) + \pi_1^\perp) = \Pi(z, \pi_1 \cap \pi_2) + \sum_{i=1}^{d-d_0} \Pi(z_i, v_i')
    \end{align*}
    Therefore,
    \begin{align*}
        \|x-y\|^2 = \sum_{i=1}^{d-d_0} \| \Pi(z_i, v_i) - \Pi(z_i, v_i') \|^2 = \sum_{i=1}^{d-d_0} (\sin^2 \theta_i) \cdot \|z_i\|^2 \le \sum_{i=1}^{d-d_0} \sin^2 \theta_i
    \end{align*}
    where the second equality follows from elementary Euclidean geometry on each 2-dimensional plane $\rho_i$. The claim thus follows (by padding the zero principal angles back in, for which $\sin 0 = 0$.) 
\end{proof}

\begin{prop}\label{wass lipschitz 2}
    Let $M = M_1 \cup M_2 \subset \RR^D$ be the union of two $d$-dimension submanifolds. Let $\tau = \min(\tau_1, \tau_2)$, where $\tau_i$ is the reach of the manifold $M_i$. Let $x \in M_1 \cap M_2$ and $y \in \RR^D$. Let $r>0$ be a number and let $s = \|y-x\|/r$. Then there exist constants $c_5, c_6 > 0$ depending only on $d$ such that the following holds:
        \[ \rho, \>  s \le c_5 \implies \Wass(\mu_{x,r}, \mu_{y,r}) \le c_6 (\rho + s ) \text{, where } \rho = r/\tau \]
\end{prop}
\begin{proof}
    We construct a 3-step transportation plan for the bound. The first two steps redistribute masses, where in the first step the claim reduces to the case of a single manifold. The third step moves two parts of mass through translation and relocation. We begin by defining the following notations:
    \begin{align*}
        B_x = \Ball(x, r), \> \alpha_{i,x} = \mc H^d(M_i \cap B_x)
    \end{align*}

    \vspace{5mm}
    \textbf{Step 1.} We redistribute mass equally for two manifolds. For $i = 1, 2$, define $\mu_{x, r}^{(i)}$ to be the normalised restriction of $\mu_{x, r}$ to $g_{x, r}(M_i)$, where $g_{x, r}(z) = (z-x)/r$. Then we see that:
    \begin{align*}
        \mu_{x, r} = \frac{\alpha_{1, x} \cdot \mu_{x, r}^{(1)} + \alpha_{2, x} \cdot \mu_{x, r}^{(2)}}{\alpha_{1,x} + \alpha_{2, x}}
    \end{align*}
    Define the following:
    \[ \mu_{x, r}' = \frac12 \bigg( \mu_{x, r}^{(1)} + \mu_{y, r}^{(2)} \bigg) \]
    so that the masses are equally distributed on the two manifolds. We have:
    \begin{align*}
        \Wass(\mu_{x,r}, \mu_{y,r}) \le & \bigg( \Wass(\mu_{x, r}, \mu_{x, r}') + \Wass(\mu_{y, r}, \mu_{y, r}') \bigg) + \Wass(\mu_{x, r}', \mu_{y, r}') \\
        \le & \bigg( \Wass(\mu_{x, r}, \mu_{x, r}') + \Wass(\mu_{y, r}, \mu_{y, r}') \bigg) + \frac12 \sum_{i=1}^2 \Wass( \mu_{x, r}^{(i)}, \mu_{y, r}^{(i)} ) 
    \end{align*}
    where the first inequality follows from the triangle inequality and the second inequality is due to the fact that the support of $\mu$ lies on $M_1 \cup M_2$. 

    If $\alpha_{1,x} \ge \alpha_{2, x}$, then a transportation plan from $\mu_{x, r}$ to $\mu_{x, r}'$ can be constructed by moving $u \cdot \mu_{x, r}^{(1)}$ to the origin and back to $u \cdot \mu_{x, r}^{(2)}$, where $u = |\alpha_{1,x} - \alpha_{2, x}| / (2\alpha_{1,x} + 2\alpha_{2, x})$. Distance of masses moved by the transportation plan is at most $2$. If $\alpha_{1,x} \le \alpha_{2, x}$, there is a completely analogous transportation plan. The transportation cost is thus bounded by:
    \[ \Wass(\mu_{x, r}, \mu_{x, r}') \le 2 u = \frac{|\alpha_{1,x} - \alpha_{2, x}|}{\alpha_{1,x} + \alpha_{2, x}} \]
    and similarly for $\Wass(\mu_{y, r}, \mu_{y, r}')$. 

    \vspace{5mm}
    \textbf{Step 2.} From the previous step, we are interested in bounding $\Wass(\mu_{x, r}^{(i)}, \mu_{y, r}^{(i)})$ for $i=1, 2$. Due to symmetry in consideration of $M_1$ and $M_2$, let us simply write $N = M_1$ and also:
    \[ \nu_x = \mu_{x, r}^{(1)}, \> \nu_y = \mu_{y, r}^{(1)} \]
    We are interested in bounding $\Wass(\nu_x, \nu_y) = \Wass(\mu_{x,r}^{(1)}, \mu_{y,r}^{(1)})$. Define:
    \[ U_x = g_{x,r}(B_x \backslash B_y), \> V_x = g_{x,r}(B_x \cap B_y) \]
    where $g_{x, r}(z) = (z-x)/r$. Since $g_{x,r}(B_x)$ is the unit ball of radius $1$ at the origin of $\RR^D$, we see that $(U_x, V_x)$ partition that ball into two regions. We will use them to divide $\nu_x$ into two parts. Also define:
    \[ \beta_x = \mc H^d(N \cap B_x \backslash B_y), \> \gamma = \mc H^d(N \cap B_x \cap B_y), \> \alpha_x = \alpha_{1, x} \]
    so that $\beta_x + \gamma = \alpha_x$. With this notation we have:
    \[ \nu_x = \frac{\beta_x \nu_x' + \gamma \nu_x''}{\beta_x + \gamma} \text{, where } \nu_x' = \nu_x \|_{U_x}, \> \nu_x'' = \nu_x \|_{V_x} \]
    and similarly for $\nu_y$. We'd like to compare the pairs $(\nu_x', \nu_y')$ and $(\nu_x'', \nu_y'')$ separately, but the ratios $\beta_x / \gamma$ and $\beta_y / \gamma$ are different. To match the ratios, define $\nu_y^\dagger$ as the linear combination of $(\nu_y', \nu_y'')$ that has the same ratio as that of $(\nu_x', \nu_x'')$ in $\nu_x$:
    \[ \nu_y^\dagger = \frac{\beta_x \nu_y' + \gamma \nu_y''}{\beta_x + \gamma} \]
    Now we construct a transportation plan from $\nu_y$ to $\nu_y^\dagger$; they are both linear combinations of $\nu_y'$ and $\nu_y''$ with different ratios. If $\beta_x \ge \beta_y$, then we transport $\nu_y$ into $\nu_y^\dagger$ by moving $u_2 \cdot \nu_y''$ to the origin and then to $u_2 \cdot \nu_y'$, where $u_2 = \gamma \cdot |(\beta_y + \gamma)^{-1} - (\beta_x + \gamma)^{-1} |$. If $\beta_x \le \beta_y$, we move $u_2 \cdot \nu_y'$ to the origin and then to $u_2 \cdot \nu_y''$. Distance of masses moved around in this process is at most $2$. Therefore,
    \[ \Wass(\nu_y, \nu_y^\dagger) \le 2 \cdot u_2 = 2\gamma \cdot \bigg| \frac1{\beta_y + \gamma} - \frac1{\beta_x + \gamma} \bigg| = 2\gamma \cdot \bigg| \frac1{\alpha_y} - \frac1{\alpha_x} \bigg| \le 2 \cdot \bigg| \frac{\alpha_y}{\alpha_x} - 1 \bigg| \]
    Therefore,
    \begin{align*}
        \Wass(\nu_x, \nu_y) \le & \Wass(\nu_x, \nu_y^\dagger) + \Wass(\nu_y^\dagger, \nu_y) \\
        = & \Wass\bigg( \frac{\beta_x \nu_x' + \gamma \nu_x''}{\beta_x + \gamma}, \frac{\beta_x \nu_y' + \gamma \nu_y''}{\beta_x + \gamma} \bigg) + \Wass(\nu_y^\dagger, \nu_y) \\
        \le & \frac{\beta_x \Wass(\nu_x', \nu_y') + \gamma \Wass(\nu_x'', \nu_y'') }{\beta_x + \gamma} + 2 \cdot \bigg| \frac{\alpha_y}{\alpha_x} - 1 \bigg|
    \end{align*}

    \vspace{5mm}
    \textbf{Step 3.} At this point our task is reduced to bounding both $\Wass(\nu_x', \nu_y')$ and $\Wass(\nu_x'', \nu_y'')$. We simply relocate all mass of $\nu_x'$ to the origin and bring it back to $\nu_y'$, so that we use the trivial bound $\Wass(\nu_x', \nu_y') \le 2$. To bound $\Wass(\nu_x'', \nu_y'')$, we observe that:
    \[ \nu_x'' = g(\mc H^d)\|_{g(M_1) \cap g(B_x \cap B_y)} = g(\mc H^d\|_{M_1 \cap B_x \cap B_y}) \]
    where $g = g_{x, r}$. Therefore,
    \[ \Wass(\nu_x'', \nu_y'') = \Wass\bigg( g_{x,r}(\mc H^d\|_{M_1 \cap B_x \cap B_y}), g_{y,r}(\mc H^d\|_{M_1 \cap B_x \cap B_y}) \bigg) \]
    Since for any $w$, $g_{y,r}(w) - g_{x, r}(w) = (x-y)/r$, we obtain $\nu_y''$ from $\nu_x''$ by pushforwarding the measure by translation through $(x-y)/r$. Thus $\Wass(\nu_x'', \nu_y'') \le s := \|x-y\|/r$. Therefore,
    \[ \frac{\beta_x \Wass(\nu_x', \nu_y') + \gamma \Wass(\nu_x'', \nu_y'') }{\beta_x + \gamma} \le \frac{\beta_x \cdot 2 + \gamma \cdot s}{\beta_x + \gamma} \le \frac{2\beta_x}{\alpha_x} + s \]
    This thus shows that:
    \[ \Wass(\nu_x, \nu_y) \le \frac{2\beta_x}{\alpha_x} + \frac{2 |\alpha_y - \alpha_x| }{\alpha_x} + s\]

    \vspace{5mm}
    \textbf{Total bound.} We now collect the terms from above and bound them using $s$ and $r$. The main tool here is Corollary \ref{lift ball volume}, which gives bounds for the volume of a manifold cut out by a ball. To apply it to balls of radii $r$ centered at $x$ and $y$, we assume that $r/\tau < c_3$ and $s<1$. Collecting the terms from above, we get the following inequality:
    \begin{align*}
        \Wass(\mu_{x,r}, \mu_{y, r}) & \le E_1 + E_2 + E_3 \\
        E_1 &= \frac{|\alpha_{1,x} - \alpha_{2, x}|}{\alpha_{1,x} + \alpha_{2,x}} + \frac{|\alpha_{1,y} - \alpha_{2, y}|}{\alpha_{1,y} + \alpha_{2,y}} \\
        E_2 &= \frac{|\alpha_{1,y} - \alpha_{1, x}|}{\alpha_{1, x}} + \frac{|\alpha_{2,y} - \alpha_{2, x}|}{\alpha_{2, x}} \\
        E_3 &= \frac{\beta_{1, x}}{\alpha_{1, x}} + \frac{\beta_{2, x}}{\alpha_{2, x}} + s
    \end{align*}
    where $\beta_{i, x} = \mc H^d(M_i \cap B_x \backslash B_y)$. Corollary \ref{lift ball volume} implies that for $i = 1, 2$, we have:
    \begin{align*}
        \frac{\alpha_{i, x}}{\omega_d r^d} \in \bigg[ 1 - c_4 \rho, 1 + c_4 \rho \bigg], \frac{\alpha_{i, y}}{\omega_d r^d} \in \bigg[ 1 - c_4 (\rho + s), 1 + c_4 (\rho + s) \bigg]
    \end{align*}
    where $\rho = r/\tau$ \footnote{Defining $\tau = \min(\tau_1, \tau_2)$, where $\tau_i$ is the reach of $M_i$, makes these bounds work simultaneously for a single value of $\tau$.}. Here we used the fact that $\d(y, M_i) \le \|x-y\| \le s \cdot r$. To work with $\beta_x$, we use the triangle inequality to see that $\Ball(x, r-sr) \subseteq \Ball(y, r) = B_y$, and thus:
    \[ \beta_{i, x} = \mc H^d(M_i \cap B_x \backslash B_y) \le \mc H^d(M_i \cap B_x \backslash \Ball(x, r-sr) ) \]
    Thus Corollary \ref{lift ball volume} again implies:
    \begin{align*}
        \beta_{i, x} \le & \omega_d r^d (1 + c_4 \rho) - \omega_d (r-sr)^d (1 - c_4 (r - sr)/\tau ) \\
        \le & \omega_d r^d \bigg( 1 + c_4 \rho - (1-s)^d (1 - c_4 \rho) \bigg) \\
        \le & \omega_d r^d \bigg( 1 - (1-s)^d + 2 c_4 \rho \bigg) \\
        \le & \omega_d r^d (d \cdot s + 2 c_4 \rho)
    \end{align*}
    where in the last inequality we used the fact that $(1-t)^d \ge 1 - d \cdot t$ for $t \in [0,1]$ and $d \ge 1$ \footnote{This is because the second derivative of $(1-t)^d$ is non-negative.} Therefore, we see that:
    \begin{align*}
        E_1 \le & \frac{c_4 \rho}{1 - c_4 \rho} + \frac{c_4 (\rho + s)}{1 - c_4(\rho + s)}, \quad  E_2 \le 2 \cdot \frac{c_4 (2\rho + s)}{1 - c_4 \rho}, \quad E_3 \le 2 \cdot \frac{d \cdot s + 2c_4 \rho}{1 - c_4 \rho} + s
    \end{align*}
    Therefore, assuming that $\rho + s \le 1/(2c_4)$ and thus $1-c_4(\rho+s) \ge 1/2$, we produce the following linear bound:
    \begin{align*}
        & \Wass(\mu_{x, r}, \mu_{y, r}) \\
        \le & E_1 + E_2 + E_3 \\
        \le & \frac{c_4 (2\rho + s) + 2c_4 (2\rho + s) + 2d \cdot s + 4c_4 \rho }{1 - c_4 (\rho + s) } + s \\
        \le & \frac{ 10c_4 \rho + (3c_4 + 2d) s }{1 - c_4 (\rho + s) } + s \\
        \le & 20c_4 \rho + (6c_4 + 4d + 1) s 
    \end{align*}
    Therefore we may set $c_5 = \min(\frac1{4c_4}, c_3, 1)$ and $c_6 = \max(20c_4, 6c_4 + 4d + 1)$ to obtain our claim.
\end{proof}

The following is a simplified version of the main result in Section 4 of \cite{uzu_pca}:
\begin{prop}\label{wass lipschitz 1}
    Let $\mu$ be the uniform distribution over $M \subset \RR^D$, which is a $d$-dimensional submanifold. Then for every $x \in M$, $\mu_{x, 0}$ is the uniform distribution over\footnote{To be precise, it is the intersection $T_x' M \cap \Ball(0, 1)$, where $T_x'M = T_x M - x$ is the linear subspace of $\RR^D$ obtained by translating $T_x M$ by $(-x)$. The $\Ball(0,1)$ here refers to the unit ball centred at the origin in $\RR^D$.} $T_x M \cap \Ball(0, 1)$. Furthermore, we have the following bound:
    \[ r/\tau \le c_7 \implies \Wass(\mu_{x, r}, \mu_{x, 0}) \le c_8 r/\tau \]
\end{prop}

Using the above and repeating the mass-redistribution argument in Step 1 of the proof in Proposition \ref{wass lipschitz 2} verbatim, we obtain the following:
\begin{prop}\label{wass lipschitz 3}
    Let $\mu$ be the uniform distribution over $M = M_1 \cup M_2 \subset \RR^D$, which is a union of two $d$-dimensional submanifolds. Then for every $x \in M_1 \cap M_2$, $\mu_{x, 0}$ is the uniform distribution over $(T_x M_1 \cup T_x M_2) \cap \Ball(0, 1)$. Furthermore, we have the following bound:
    \[ r/\tau \le c_9 \implies \Wass(\mu_{x, r}, \mu_{x, 0}) \le c_{10} r/\tau \]
\end{prop}

We prove a result that allows us to work with one manifold at a time when dealing with the smooth region of $M_1 \cup M_2 \cup \RR^D$, where each $M_i$ is a submanifold.

\begin{prop}\label{separation}
    Let $M = M_1 \cup M_2 \subset \RR^D$ be a union of two $d$-dimensional submanifolds, such that $M_1 \cap M_2$ is nonempty. Suppose that for every $x \in M_1 \cap M_2$, we have $\dim(T_x M_1 \cap T_x M_2) = d_0$ for a fixed $d_0$, and that principal angles of $(T_x M_1, T_x M_2)$ are all $\ge \phi$. Then we have:
    \begin{align*}
        h(M_1, M_2) = \inf_{x \in M_1} \frac{\d(x, M_2)}{\d(x, M_1 \cap M_2)} \in (0, 1]
    \end{align*}
    In particular, for any $r > 0$ and $x \in M_1$, we have:
    \[ x \notin \Ball(M_1 \cap M_2, h^{-1} \cdot r) \implies \Ball(x, r) \cap M_2 = \emptyset \]
    where $h = h(M_1, M_2)$.
\end{prop}
\begin{proof}
    Firstly $h \le 1$ holds trivially since $\d(x, M_2) \le \d(x, M_1 \cap M_2)$. Let $r>0$ be a number satisfying:
    \begin{align}
        \frac r{\tau} < \min\bigg(\frac{\sqrt 2 - 1}2, \> \frac{1-\cos\phi}{12} \bigg) \label{rassump}
    \end{align}
    where $\tau = \min(\tau_1, \tau_2)$ and $\tau_i$ is the reach of $M_i$. The angle condition involving $\cos\phi$ will be used in the final steps of the proof. Since $M_1$ partitions into the disjoint union of $M_1 \cap B$ and $M_1 \backslash B$ where $B = \Ball(M_1 \cap M_2, r)$, we may write:
    \[ h = \min \bigg( \inf_{x \in M_1 \cap B} \frac{\d(x, M_2)}{\d(x, M_1 \cap M_2)}, \quad \inf_{x \in M_1 \backslash B} \frac{\d(x, M_2)}{\d(x, M_1 \cap M_2)} \bigg) \]
    The second term is easily seen to be positive:
    \[ \inf_{x \in M_1 \backslash B} \frac{\d(x, M_2)}{\d(x, M_1 \cap M_2)}  \ge \frac{\inf_{x \in M_1 \backslash B} \d(x, M_2)}{\sup_{x \in M_1 \backslash B} \d(x, M_1 \cap M_2)} > 0 \]
    where the numerator is positive since $M_1 \backslash B$ is a compact set and $x \mapsto \d(x, M_1)$ is positive and continuous, and the denominator is finite since $M_1, M_2$ are bounded.

    \vspace{5mm}
    \textbf{Reduction to linear algebra.} We now examine the fraction $\d(x, M_2)/\d(x, M_1 \cap M_2)$ when $x \in M_1 \cap B$. Denote $r_0 = \d(x, M_1 \cap M_2)$. By compactness of $M_1 \cap M_2$, we see that there is a point $x_0 \in M_1 \cap M_2$ such that $r_0 = \d(x, x_0)$. Also denote $\pi_i = T_{x_0} M_i$ and:
        \[ x_1 = \Pi(x, \pi_1), \quad x_2 = \Pi(x_1, M_2) \]
    Then we apply Lemma \ref{geodesic bound} due Equation \ref{rassump} with $\|x - x_0\| = r_0 < (\sqrt 2 - 1) \tau$ and \footnote{In detail: $\|x_1 - x_2\| = \inf_{y \in M_2} \|x_1 - y\| \le \|x_1 - x_0\| \le \|x - x_0\| = r_0$ and thus $\|x_2 - x_0\| \le \|x_2 - x_1 \| + \|x_1 - x_0\| \le 2r_0$.} $\|x_2 - x_0\| \le 2r_0 < (\sqrt 2 - 1)\tau$,
        \[ \d(x, \pi_1) \le r_0^2/\tau, \quad \d(x_2, \pi_2) \le (2r_0)^2/\tau \]
    Therefore:
    \begin{align}
        \d(x, M_2) \ge & \d(x_1, M_2) - \d(x, x_1) \nonumber \\
        =& \d(x_1, x_2) - \d(x, \pi_1) \nonumber \\
        \ge & \d(x_1, \pi_2) - \d(x_2, \pi_2) - \d(x, \pi_1) \nonumber \\
        \ge & \d(x_1, \pi_2) - 5\tau^{-1} r_0^2 \label{dxm2}
    \end{align}
    
    \vspace{5mm}
    \textbf{Linear algebra.} Now we are interested in controlling $\d(x_1, \pi_2)$, and this is an exercise of linear algebra since $x_1 \in \pi_1$. Write $x_1^\perp = \Pi(x_1, \pi_2)$ and $(z, z_\perp) = (x_1 - x_0, x_1^\perp - x_0)$, so that $\d(x_1, \pi_2) = \d(x_1, x_1^\perp) = \d(z, z_\perp)$. Let $\pi_i' = \pi_i - x_0$ be a vector space (satisfying $0 \in \pi_i'$), and let $A_i \in \RR^{D \times d}$ be a matrix whose columns are an orthonormal basis of $\pi_i'$. Then $z \in \pi_1'$ and $z_\perp = \Pi(z, \pi_2') = A_2A_2^\top z$ and
    \begin{align}
        \d(x_1, \pi_2) = \d(z, z_\perp)^2 = \|z\|^2 - \|z_\perp\|^2 \label{eqz1}
    \end{align}
    by Pythagoras' theorem. Now for some $u \in \RR^d$ we may write $z = A_1 u$, so that $\|z_\perp\| = \|A_2 A_2^\top z\| = \|A_2 A_2^\top A_1 u \|$. Using the fact that the map $w \mapsto A_i w$ is distance-preserving and the assumption that principal angles between tangent spaces $(\pi_1, \pi_2)$ are $\ge \phi$, we get:
    \begin{align}
        \| z_\perp \| = \|A_2 A_2^\top A_1 u \| = \|A_2^\top A_1 u \| \le \|A_2^\top A_1\| \cdot \|u\| = \|A_2^\top A_1\| \cdot \|z\| \le (\cos \phi) \cdot \|z\| \label{eqz2}
    \end{align}
    and we also note that:
    \begin{align}
        \|z\| = \d(x_1, x_0) \ge \d(x, x_0) - \d(x, x_1) = \d(x, x_0) - \d(x, \pi_1) \ge r_0 - r_0^2/\tau \label{eqz3}
    \end{align}

    \vspace{5mm}
    \textbf{Combining the bound.} We plug Equations \eqref{eqz1}, \eqref{eqz2}, \eqref{eqz3} into Equation \eqref{dxm2}:
    \begin{align*}
        \d(x, M_2) \ge & \d(x_1, \pi_2) - 5r_0^2/\tau \\
        \ge & \sqrt{1 - \cos^2\phi} \cdot \|z\| - 5r_0^2/\tau \\
        \ge & (1-\cos\phi)(r_0 - r_0^2/\tau) - 5r_0^2/\tau \\
        \ge & r_0 - \bigg( (\cos \phi) r_0 + 6r_0^2/\tau \bigg) \\
        \ge & \frac{1-\cos\phi}2 \cdot r_0 > 0 
    \end{align*}
    where in the last equality, we used the assumption $r_0/\tau \le (1-\cos \phi)/12$ in Equation \eqref{rassump}. Therefore, recalling that $r_0 = \d(x, M_1 \cap M_2)$, the following holds for all $x \in M_1 \backslash B$:
    \[ \frac{\d(x,M_2)}{\d(x, M_1 \cap M_2)} \ge \frac{1-\cos\phi}2 > 0\]
    and the claim is proven.
\end{proof}

The MMD for the Gaussian kernel can be controlled with the Wasserstein distance as follows:

\begin{lem}\label{mmd_wass_comparison}
    Let $\kappa(x,y) = e^{-\gamma \|x-y\|^2}$ be a Gaussian kernel. Then whenever $\|f\|_\kappa \le 1$, $f$ is a $\sqrt{2 \gamma}$-Lipschitz function. Therefore, for any probability measures $\mu, \nu$ valued in $\RR^D$ with finite first moment, we have:
    \[ \MMD_\kappa(\mu, \nu) \le \sqrt{2 \gamma} \cdot \Wass (\mu, \nu) \]
\end{lem}
\begin{proof}
    \begin{align*}
        |f(x) - f(y)| =& |\langle f, \kappa(x,-) - \kappa(y,-) \rangle| \\
        \le & \|f\|_\kappa \cdot \| \kappa(x,-) - \kappa(y,-) \|_\kappa \\
        =& \|f\|_\kappa \cdot \sqrt{\kappa(x,x) + \kappa(y,y) - 2\kappa(x,y)} \\
        =& \sqrt{2} \|f\|_\kappa \cdot \sqrt{1 - e^{-\gamma \|x-y\|^2}} \\
        \le & \sqrt{2 \gamma} \|f\|_\kappa \cdot \| x-y \|
    \end{align*}
    where in the last inequality we used $\sqrt{1-e^{-s^2}} \le s$. Therefore any function with $\|f\|_\kappa \le 1$ is also a $\sqrt{2\gamma}$-Lipschitz function. The conclusion follows by the integral probability metric definition.
\end{proof}

%%%%%%%%%%%%%%%%%%%%%%%%%%%%%%%%%%%%%%%%%%%%%%%%%%%%%%%%%
%%%%%%%%%%%%%%%%%%%%%%%%%%%%%%%%%%%%%%%%%%%%%%%%%%%%%%%%%
\subsection{Measures on Manifolds}

We use the following lemma from \cite{uzu_pca}, which is a simple extension of Proposition 6.3 of \cite{nsw}. It controls the deviation of geodesic from the first order approximation:
\begin{lem}\label{geodesic bound}
	Let $M$ be a smooth compact $n$-manifold embedded in $\RR^D$ with reach $\tau$. Suppose that $x,y$ are connected by a (unit speed) geodesic $\gamma: [0, \tilde r] \rightarrow M$ of length $\tilde r$ with $\gamma(0) = x, \gamma(\tilde r) = y$, and denote $r = \|x-y\|$. Then the following inequalities hold:
	\[ \tilde r - \frac{\tilde r^2}{2\tau} \le r \le \tilde r \]
	If $r \le 0.5 \tau$, then the following hold:
	\begin{align*}	
	\frac{\tilde r}\tau \le 1 - \sqrt{1 - \frac{2r}{\tau} } \text{, and } \| y - (x + \tilde r \dot \gamma(0)) \| \le \frac{\tilde r^2}{2\tau}
	\end{align*}	
If $r \le (\sqrt{2} - 1)\tau \approx 0.4 \tau$, then the following also hold:
	\begin{align*}	
	\tilde r \le r + \frac{r^2}{\tau} \text{, and } \| y - (x + \tilde r \dot \gamma(0)) \| \le \frac{r^2}{\tau}
	\end{align*}	
\end{lem}

\begin{lem}
    Let $M \subset \RR^D$ be a compact set and let $\tau$ be its reach. Let $\pi_M$ be the projection map to $M$, such that for any $x \in \RR^D$, $\pi_M(x)$ is the set of points on $M$ that minimises the distance to $M$. The following hold:
    \begin{enumerate}
        \item The distance function $x \mapsto \d(x, M) = \inf \{\|y-x\| \>|\> y \in M\}$ is continuous.
        \item For $0 < r < \tau$, $\pi_M|_{\Ball(M, r)}$ is a single-valued continuous function.
    \end{enumerate}
\end{lem}
\begin{proof}
    (1) From the definition it easily follows that $\d(-, M)$ is a Lipschitz function; we have that: $|\d(x, M) - \d(x', M)| \le \|x-x'\|$.

    (2) Let's write $\pi = \pi_M|_{\Ball(M,r)}$ for the moment. Let $x \in \Ball(M, r)$. Suppose that $x_n \rightarrow x$ but $\pi(x_n)$ doesn't converge to $\pi(x)$. Then there exists $s>0$ such that $\pi(x_n) \notin \Ball(\pi(x), s)$.

    Since $\d(y, M) = \|y - \pi(y)\|$ for each $y \in \Ball(M, r)$, the continuity of $\d(-, M)$ implies that there is a convergence $\|x_n - \pi(x_n)\| \rightarrow \|x - \pi(x)\|$. Since we also have $x_n \rightarrow x$, we have $\|x - \pi(x_n)\| \rightarrow \|x - \pi(x)\|$. Thus $\inf \{ \|x - y\| \>|\> y \in M \backslash \Ball(\pi(x), s) \} = \|x - \pi(x)\| = \d(x, M)$.
    
    This is a contradiction. Since $M \backslash \Ball(\pi(x), s)$ is a compact set, the distance function $y \mapsto \|y-x\|$ attains a minimum on some $z \in M \backslash \Ball(\pi(x), s)$. This violates the definition of reach, which requires a unique nearest point of $x$ on $M$, which can't be simultaneously $\pi(x)$ and $z$.
\end{proof}

\begin{lem}
    Let $M \subset \RR^D$ be a compact path-connected set and let $\tau$ be its reach. If $x, y \in M$ satisfies $\|x-y\| < \tau$, then there exists a continuous path on $M$ that connects $(x,y)$ such that every point on the path has distance at most $\|x-y\|$ from both $x$ and $y$.
\end{lem}
\begin{proof}

    Define a path $\bar\gamma:[0,1] \rightarrow M$ by $\bar \gamma(t) = (1-t)x + ty$, the line segment connecting $(x,y)$. Since $\|x-y\| < \tau$, every point on $\bar\gamma$ is within distance $\tau$ from $x$, and thus $\pi_M \circ \bar\gamma: [0,1] \rightarrow M$ is a (single-valued) continuous function. Let's write $\gamma = \pi_M \circ \bar\gamma$.

    Let $t_0 \in [0,1]$ and write $z = \gamma(t_0)$ and $\bar z = \bar \gamma(t_0)$. Then we have:
    \[ \|z - x \| \le \|z - \bar z \| + \|\bar z - x\| \le \|y - \bar z \| + \|\bar z - x\| = \|y - x\| \]
    where the first inequality is the triangle inequality, the second inequality is due to the definition of $\gamma$, and the last equality is due to $(x, \bar z, y)$ lying on one line. Therefore $\|z - x \| \le \|y-x\|$, and by symmetry of the argument in $(x,y)$, we also get $\|z-y\| \le \|y-x\|$.
\end{proof}

\begin{prop}
    Let $M$ be a $d$-dimensional submanifold. Let $\pi_x: \RR^D \rightarrow T_x M$ be the projection map to $T_x M$, and let $\tilde \pi_x := \pi_x|_M : M \rightarrow T_x M$ and $\tilde \pi_{x, r} := \pi_x|_{M \cap \Ball(x, r)}$. The following hold:
    \begin{enumerate}
        \item When $r < \tau / 2$, $\tilde \pi_{x, r}$ has nonsingular derivatives and is a diffeomorphism.
        \item For any $y \in M$, we have $J_y \tilde \pi_x = \det(A_x^\top A_y)$, where $A_x \in \RR^{D \times d}$ is any orthonormal frame of $T_x M$.
        \item For any $y \in M$, the following bound holds:
        \[ \cos \theta_{x, y} \ge 1 - \frac{\d_M(x, y)}{\tau} \]
        where $\theta_{x,y} = \measuredangle_{\max}(T_x M, T_y M)$.
        \item For any $y \in M$, the following bound holds:
        \[ J_y \tilde \pi_x \in [(\cos \theta_{x, y} )^d, 1] \]
        \item Suppose that $r < (\sqrt 2 - 1)\tau$. We have 
        \[ \frac{\mc H^d( M \cap \Ball(x, r))}{\omega_d r^d } \in \bigg[ (1-\rho^2/4)^{d/2}, (1-\rho - \rho^2)^{-d} \bigg]\]
        where $\rho = r / \tau$.
    \end{enumerate}
\end{prop}
\begin{proof}
    (1) The nonsingularity is Lemma 5.4 from \cite{nsw}. By applying the inverse function theorem locally at each point where the derivative is non-singular, we see that $\tilde \pi_{x, r}$ is a diffeomorphism.

    (2) This is because $\d \tilde {\pi_x} (v) = A_x^\top v$ for each (embedded) tangent vector $v \in T_y M$. 

    (3) This is Proposition 6.2 from \cite{nsw}. 

    (4) This folllows from (2) and the definition of principal angles.

    (5) The lower bound is Lemma 5.3 from \cite{nsw}. To see the upper bound, we note by the Area Formula of geometric measure theory that the volume $\mc H^d(M \cap \Ball(x,r))$ is the integral of Jacobian of inverse-projection in $\pi(M \cap \Ball(x,r))$, i.e.
    \[ \mc H^d(M \cap \Ball(x,r)) = \int_{\pi(M \cap \Ball(x,r))} ( J_{z'} \tilde{\pi_x} )^{-1} \d z \]
    where $z' \in M \cap \Ball(x, r)$ is the unique point such that $\pi_x(z') = z$. Now note that $\pi(M \cap \Ball(x, r))$ is contained in a ball of radius $r$ in $T_x M$, so that its measure is at most $\omega_d r^d$. Furthermore, the inverse of Jacobian in the integrand is at most $(1 - \d_M(x, y) / \tau)^{-d}$ by (3) and (4). By the bound on geodesic length (\ref{geodesic bound}), we have $\d_M(x,y) / \tau \le \rho + \rho^2$ and thus obtain the claim.
\end{proof}

\begin{cor}
    Let $M \subset \RR^D$ be a $d$-dimensional submanifold. There exist constants $c_1 >0 , c_2 \ge 0$ depending only on $d$ such that the following hold.
    \[ r < c_1 \tau \implies \frac{\mc H^d( M \cap \Ball(x, r))}{\omega_d r^d } \in \bigg[ 1 - \frac {c_2r}\tau, 1 + \frac {c_2 r} \tau \bigg] \]
\end{cor}
\begin{proof}
    Let's first assume that $r < (\sqrt 2 - 1)\tau$. Then we can relax the upper bound of (5) of the previous Proposition into $(1-\sqrt 2 \rho)^{-d}$. Now we further relax our lower and upper bounds, which are given by:
    \[ f_1(t) = (1-t^2/4)^{d/2}, \> f_2(t) = (1-\sqrt 2 t)^{-d} \]
    Their second derivatives are given by:
    \[ f_1''(t) = d \cdot (1-t^2/4)^{d/2} \cdot \frac{(d-1)t - 4}{(t^2-4)^2}, \> f_2''(t) = 2d \cdot (d + 1) \cdot (1-\sqrt2 t)^{- d - 2} \]
    Then we see that $f_1''(t) \le 0$ for $t \in [0, 4/(d-1)]$ and $f_2''(t) \ge 0$ for $t \in [0, 1/\sqrt{2}]$. 

    Therefore, if we let $c_1 = \min(\sqrt2 - 1, 4/(d-1))$, then we see that $1 - c_2 t \le f_1(t)$ and $1 + c_2 t \ge f_2 (t)$, where $c_2 \ge 0$ is given by:
    \[ c_2 = \max \bigg( 1 - (3/4)^{d/2}, (\sqrt 2 - 1)^{-d} \bigg) \]
    which are values obtained from slopes of $f_1(t), f_2(t)$ by plugging in $t = 1$ and $t = \sqrt 2 - 1$ respectively.
\end{proof}

\begin{cor}\label{lift ball volume}
    Let $M \subset \RR^D$ be a $d$-dimensional submanifold, and let $r > 0$. Suppose $x \in \RR^D$ is a point satisfying $\d(x, M) = s \cdot r$. There exists constants $c_3, c_4 \ge 0$ depending only on $d$ such that the following holds.
    \[ r < c_3 \tau , \> s < 1 \implies \frac{\mc H^d( M \cap \Ball(x, r))}{\omega_d r^d } \in \bigg[ 1-c_4 (s + r / \tau), 1 + c_4 (s + r / \tau)  \bigg]   \]
\end{cor}
\begin{proof}
    We start by assuming that $s < 1$ and $r < \tau$, so that there is a unique point of projection $y \in M$ minimising distance from $x$, so that $\|x-y\| = sr$. By the triangle inequality, we have the inclusions:
    \[ \Ball(y, r-sr) \subseteq \Ball(x, r) \subseteq \Ball(y, r + sr) \]
    which implies:
    \[ \mc H^d(M \cap \Ball(y, r-sr)) \le \mc H^d(M \cap \Ball(x, r)) \le \mc H^d(M \cap \Ball(y, r+sr)) \]
    Applying the previous Corollary, we get the lower bound:
    \[ \mc H^d(M \cap \Ball(y, r-sr)) \ge (1 - c_2 (r-sr)/\tau) \cdot \omega_d (r-sr)^d \ge \omega_d r^d \cdot (1-c_2 r/\tau)(1-s)^d \]
    and the upper bound:
    \[ \mc H^d(M \cap \Ball(y, r+sr)) \le (1 + c_2(r+sr)/\tau) \cdot \omega_d (r+sr)^d \le \omega_d r^d \cdot (1 + 2c_2 r/\tau)(1+s)^d \]
    where we are assuming that $r < (c_1/2) \tau$, so that $r-sr \le r+sr \le c_1 \tau$ and the previous Corollary applies. Assuming $d \ge 1$ and $t \in [0, 1]$, the functions $t \mapsto (1-t)^d$ and $t \mapsto (1+t)^d$ both have non-negative second derivative, so that we have $(1-t)^d \ge 1-d \cdot t$ and $(1+t)^d \le 1 + 2^d \cdot t$. Letting $c_4' = \max(2^d, 2c_2)$, we get:
    \[ \frac{\mc H^d(M \cap \Ball(x, r))}{\omega_d r^d} \in \bigg[ (1 - c_4' s)(1 - c_4' r/\tau), (1 + c_4' s)(1 + c_4' r/\tau) \bigg] \]
    Expanding the brackets, we get:
    \[(1 + c_4's)(1 + c_4' r/\tau) = 1 + c_4' s + c_4' r/\tau + c_4' s r / \tau \le 1 + c_4' s + c_4' \cdot 2r / \tau \le 1 + 2 c_4' (s + r/\tau) \]
    and similarly $(1-c_4's)(1-c_4'r/\tau) \ge 1 - 2c_4'(s+r/\tau)$. We thus obtain the claim by setting $c_3 = c_1/2$ and $c_4 = 2c_4'$.
\end{proof}

\subsection{Principal angle}

To work with local behaviour of the space given by a union of two manifolds, we must understand the space of pair of subspaces. Indeed, every pair of linear subspaces of the same dimension can be characterised by \textit{principal angles}, up to (simultaneous) rigid motion. 

\begin{defn}
    Given $\pi_1, \pi_2 \in \text{\rm Gr}(d, D)$, let $A_i \in \RR^{D \times d}$ be a matrix with orthonormal columns that span $\pi_i$. Denote by $\pran(\pi_1, \pi_2) \in [0, 1]^d$ the singular values of the matrix $A_1^\top A_2$, arranged in the descending order. The \textit{principal angles} of $(\pi_1, \pi_2)$ are defined as the angles $(\theta_1, \ldots \theta_d) \in [0, \pi/2]^d$ such that $(\cos \theta_1, \ldots \cos\theta_d) = \pran(\pi_1, \pi_2)$, which satisfy $\theta_1 \le \cdots \le \theta_d$.
\end{defn}

We note in particular that $\theta_1 = \cdots = \theta_{d_0} = 0 < \theta_{d_0 + 1}$, where $d_0 = \dim(\pi_1) = \dim(\pi_2)$.

The largest principal angle has a simple interpretation:
\begin{lem}
    If $\pran(\pi_1, \pi_2) = (\cos \theta_1, \ldots \cos \theta_d)$ for $\pi_1, \pi_2 \in \on{Gr}(d, D)$, then:
    \[ \theta_d = \max_{x \in \pi_1} \min_{y \in \pi_2} \angle(x,y) = \d_H( \pi_1 \cap \SS, \pi_2 \cap \SS) \]
    Here $\angle(x,y) = \cos^{-1}(\langle x,y\rangle / (\| x \| \cdot  \| y \| ))$, $\d_H(A, B) = \inf \{ r \>|\> \Ball(A, r) \supseteq B, \Ball(B, r) \supseteq A \}$ is the Hausdorff distance between two sets $A, B$, and $\SS$ is the unit $(D-1)$-dimensional sphere.
\end{lem}
\begin{proof}
    Let $A_i \in \RR^{D \times d}$ be a matrix whose columns form an orthonormal basis of $\pi_i$. We have:
    \begin{align*}
        \cos \theta_D = \min_{\|z\|=1} \|A_1^\top A_2 z\| = \min_{\|y\|=1, y \in \Pi_2} \|A_1^\top y\| = \min_{\|y\|=1, y \in \Pi_2} \langle y_1,  y \rangle
    \end{align*}
    where $y_1$ is the unit vector in the direction of $A_1 A_1^\top y$. Noting that $\langle y_1,y\rangle = \max_{\|x\|=1, x \in \pi_1} \langle x,y \rangle$, we have $\cos \theta_D = \min_{\|y\|=1, y\in\pi_2} \max_{\|x\|=1, x\in\pi_1} \langle x,y\rangle$.
\end{proof}

Principal angles characterise pairs of subspaces up to rotation.

\begin{prop}
    $\pran$ induces the following bijection:
    \[ \pran : \frac{\on{Gr}(d, D) \times \on{Gr}(d, D)}{O(D)} \longrightarrow S(d, \max(0, 2d - D)) \]
    where $S(k, j) = \{ (t_1, \ldots t_k) \>|\> 1 \ge t_1 \ge \cdots \ge t_k \ge 0 , \> t_1 = \cdots = t_j = 1 \}$, which is a set homeomorphic to the standard $(k-j)$-simplex. 
    
    Explicitly, we have the following. If $(\pi_1, \pi_2), (\pi_1', \pi_2') \in \on{Gr}(d, D) \times \on{Gr}(d, D)$ satisfy $\pran(\pi_1, \pi_2) = \pran(\pi_1', \pi_2')$, then there exists an element $A \in O(D)$ such that $(A \pi_1, A \pi_2) = (\pi_1', \pi_2')$. Furthermore, if $(t_1, \ldots t_d) \in [0, 1]^d$ satisfies $t_1 \ge \cdots \ge t_d$ and $t_1 = \cdots = t_j = 1$ with $j = d - \max(0, 2d - D)$, then there exists $(\pi_1, \pi_2) \in \on{Gr}(d, D) \times \on{Gr}(d, D)$ such that $\pran(\pi_1, \pi_2) = (t_1, \ldots t_d)$.
\end{prop}
\begin{proof}
    We prove the explicit version. Suppose that $(\pi_1, \pi_2), (\pi_1', \pi_2') \in \on{Gr}(d, D) \times \on{Gr}(d, D)$ with $\pran(\pi_1, \pi_2) = \pran(\pi_1', \pi_2')$. Let $A_i \in \RR^{D \times d}$ be a matrix with orthonormal columns spanning $\pi_i$, and similarly define $A_i'$. Without loss of generality, we may assume that $A_1 = A_1' = J$, since by Gram-Schmidt there are matrices $H, H' \in O(D)$ such that $H A_1 = H' A_1' = J$, where $J = [I_d, 0_{D-d, d}] \in \RR^{D \times d}$ has $1$ on the diagonal and zero elsewhere. Let's relabel $B = A_2, B' = A_2'$. Also write $B^\top = [B_1^\top, B_2^\top]$ and $(B')^\top = [(B_1')^\top, (B_2')^\top]$, where $B_1, B_1'$ are both $(d\times d)$-matrices.
    
    Since $\pran(\pi_1, \pi_2) = \pran(\pi_1', \pi_2')$, the singular values of $(d \times d)$-matrices $J^\top B = B_1$ and $J^\top B' = B_1'$ are equal. Therefore there exist $U, V \in O(d)$ such that $B_1' = U B_1 V^\top$. Then:
    \begin{align*}
        \begin{bmatrix} U & 0 \\ 0 & I \end{bmatrix} \begin{bmatrix} B_1 \\ B_2 \end{bmatrix} V^\top = \begin{bmatrix} U B_1 V^\top \\ B_2 V^\top \end{bmatrix} = \begin{bmatrix} B_1' \\ B_3 \end{bmatrix} \text{, where } B_3 = B_2 V^\top
    \end{align*}
    The right hand side also has orthonormal columns, so that we have $(B_1')^\top B_1' + B_3^\top B_3 = I_d$. Since $B'$ also have orthonormal columns, we also have $(B_1')^\top B_1' + (B_2')^\top B_2' = I_d$. Therefore, $B_3^\top B_3 = (B_2')^\top B_2'$. This guarantees the existence of $W \in O(D-d)$ such that $W B_3 = B_2'$. Therefore, for $Z = \smallbmat{ U & 0 \\ 0 & W }$, we have:
    % $Z = \begin{smallbmat} U & 0 \\ 0 & W \end{smallbmat}$
    \begin{align*}
        ZBV^\top = \begin{bmatrix} U & 0 \\ 0 & W \end{bmatrix} \begin{bmatrix} B_1 \\ B_2 \end{bmatrix} V^\top = \begin{bmatrix} U B_1 V^\top \\ W B_2 V^\top \end{bmatrix} = \begin{bmatrix} B_1' \\ W B_3 \end{bmatrix} = \begin{bmatrix} B_1' \\ B_2' \end{bmatrix} = B'
    \end{align*}
    Therefore $Z \pi_2 = \pi_2'$. The block diagonal form of $Z$ also ensures that $Z$ leaves $\pi_1 = \pi_1' = \RR^k$ invariant. Therefore, we have $(Z \pi_1, Z \pi_2) = (\pi_1', \pi_2')$ as desired.
\end{proof}

\begin{cor}\label{two subspace standard matrix}
    Given $\pi_1, \pi_2 \in \on{Gr}(d, D)$, suppose that $A_i \in \RR^{D \times d}$ has columns forming an orthonormal basis of $\pi_i$. Let $d_0 = \dim (\pi_1 \cap \pi_2) \ge 2d - D$ and let $d_1 = D-2d + d_0$. Then there exist matrices $U \in O(D)$ and $V_1, V_2 \in O(d)$ such that $UA_1 V_1 = \tilde A_1$ and $U A_2 V_2 = \tilde A_2$, where
    \begin{align*}
        \tilde A_1 = \begin{bmatrix} I_{d} \\ 0 \end{bmatrix}, \quad \tilde A_2 = \begin{bmatrix} I_{d_0} & 0 \\ 0 & \cos \Theta \\ 0 & \sin \Theta \\ 0 & 0
        \end{bmatrix} \in \RR^{D \times d}
    \end{align*}
    where $\Theta = \on{diag}(\theta_{d_0+1}, \ldots \theta_d) \in \RR^{(d-d_0) \times (d-d_0)}$ is the diagonal matrix of nonzero principal angles $\pran(\pi_1, \pi_2) = (\theta_1, \ldots \theta_d)$.
\end{cor}
\begin{proof}
    $\tilde A_1, \tilde A_2$ have orthonormal columns and furthermore $\tilde A_1^\top \tilde A_2$ and $A_1^\top A_2$ have the same singular values. Therefore the previous proposition applies, and the claim follows.
\end{proof}

We recall the following Lemma from \cite{spectral_clustering_local_pca}, dealing with covariance of one disk:

\begin{lem} \label{covariance of disk}
    Let $\mu$ be the uniform distribution over $\pi \cap \Ball(0, 1)$, where $\pi$ is a $d$-dimensional subspace $\pi$ of $\RR^D$. Then,
    \[ \vec\lambda \Sigma[\mu] = \frac1{d+2} (\underbrace{1, \ldots 1}_d , \underbrace{0, \ldots 0}_{D-d} ) \]
\end{lem}

We also need the covariance of \textit{two} disks:
\begin{prop} \label{covariance of two disks}
    Let $\pi_1, \pi_2 \in \on{Gr}(d, D)$ with $\dim(\pi_1 \cap \pi_2) = d_0 \ge 2d - D$. Define $\mu = \frac12 (\mu_1 + \mu_2)$, where $\mu_i$ is the uniform measure over $\pi_i \cap \Ball(0, 1)$. Then eigenvalues of the covariance of $\mu$ are:
    \[ \vec\lambda\Sigma[\mu] = \frac1{(d+2)} ( \underbrace{1, \ldots 1}_{d_0}, \cos^2 (\theta_{d_0+1}/2), \ldots \cos^2 (\theta_{d}/2) , \sin^2 (\theta_{d}/2) , \ldots \sin^2 (\theta_{d_0 + 1} / 2), \underbrace{0, \ldots 0}_{D-2d + d_0} )\]
    % \[ \vec\lambda\Sigma[\mu] = \frac1{(d+2)} \bigg(\cos^2 \frac{\theta_{1}}{2}, \ldots \cos^2 \frac{\theta_{d}} 2 , \sin^2 \frac{\theta_{d}}2 , \ldots \sin^2 \frac{\theta_{1}}2, \underbrace{0, \ldots 0}_{D-2d} \bigg) \in \RR^D \]
    where $\theta_1 \le \cdots \le \theta_d$ are the principal angles between $(\pi_1, \pi_2)$ with $\theta_1 = \ldots = \theta_{d_0} = 0 < \theta_{d_0 + 1}$. 
\end{prop}
\begin{proof}
    By starting from the matrix form in Corollary \ref{two subspace standard matrix} and then by applying multiple 2-dimensional rotations to the standard matrix form of $(\pi_1, \pi_2)$, the following can be proven. There exists an orthogonal matrix $V \in O(D)$ and matrices $A_1, A_2$ such that columns of each $V A_i$ is an orthonormal basis of $\pi_i$:
    \begin{align*}
        A_1 = \begin{bmatrix} I_{d_0} & 0 \\ 0 &  \cos \frac12 \Theta \\ 0 & \sin \frac12 \Theta \\ 0 & 0 \end{bmatrix}, A_2 = \begin{bmatrix} I_{d_0} & 0 \\ 0 & \cos \frac12 \Theta \\ 0 & -  \sin \frac12 \Theta \\ 0 & 0 \end{bmatrix}
    \end{align*}
    where $\Theta \in \RR^{(d-d_0) \times (d-d_0)}$ is the diagonal matrix of nonzero principal angles. Therefore $A_1 = U_1 J, A_2 = U_2 J$ where $J$ is given by $J^\top = [I_d, 0] \in \RR^{d \times D}$, and
    \begin{align*}
        U_1 = \begin{bmatrix} 
            I_{d_0} & & & \\
            & \cos \frac12 \Theta & - \sin \frac12  \Theta & \\
            & \sin \frac12 \Theta & \cos \frac12  \Theta & \\
            & & & I_{D-2d+d_0}
        \end{bmatrix}, \>
        U_2 = \begin{bmatrix} 
            I_{d_0} & & & \\
            & \cos \frac12  \Theta & \sin \frac12  \Theta & \\
            & - \sin \frac12  \Theta & \cos \frac12  \Theta & \\
            & & & I_{D-2d+d_0}
        \end{bmatrix}
    \end{align*}
    Here the matrix $V \in O(D)$ simply plays the role of an orthonormal coordinate transform and can be safely ignored in calculating the eigenvalues of $\Sigma[\mu]$. Indeed, orthonormal coordinate transformation induces a conjugation on the covariance matrix, and leaves its eigenvalues invariant. Thus without loss of generality, assume that columns of each $A_i$ is an orthonormal basis of $\pi_i$. 
    
    Let $Z \in \RR^D$ be a random vector, drawn from the uniform distribution over the unit $d$-dimensional disk that spans the first $d$ canonical basis vectors of $\RR^D$. Then for each $X_i \sim \mu_i$, we have $X_i = U_i Z$. This implies that:
    \[ \Sigma[\mu_i] = \EE[X_iX_i^\top] = U_i \EE[ZZ^\top] U_i^\top = \frac1{(d+2)} U_i \begin{bmatrix} I_d & 0 \\ 0 & 0 \end{bmatrix} U_i^\top \]
    Thus we write $U_1, U_2$ in block diagonal forms:
    \[ U_i = \begin{bmatrix} U_i^{(11)} & U_i^{(12)} \\ U_i^{(21)} & U_i^{(22)} \end{bmatrix} \]
    where
    \begin{align*}
        U_i^{(11)} =& \begin{bmatrix} I_{d_0} & 0 \\ 0 & \cos\frac12 \Theta \end{bmatrix}, 
        U_i^{(12)} = \begin{bmatrix} 0 & 0 \\ (-1)^i \sin\frac12 \Theta & 0 \end{bmatrix} \\
        U_i^{(21)} =& \begin{bmatrix} 0 & (-1)^{i+1} \sin\frac12 \Theta \\ 0 & 0 \end{bmatrix}, 
        U_i^{(22)} = \begin{bmatrix} \cos\frac12\Theta & 0 \\ 0 & I_{D-2d+d_0} \end{bmatrix}
    \end{align*}
    Thus we compute:
    \begin{align*}
        \Sigma[\mu_1] = \begin{bmatrix} U_{11} & U_{12} \\ U_{21} & U_{22} \end{bmatrix} \begin{bmatrix} I_d & 0 \\ 0 & 0 \end{bmatrix} \begin{bmatrix} U_{11}^\top & U_{21}^\top \\ U_{12}^\top & U_{22}^\top \end{bmatrix} = \begin{bmatrix} U_{11} & 0 \\ U_{21} & 0 \end{bmatrix} \begin{bmatrix} U_{11}^\top & U_{21}^\top \\ U_{12}^\top & U_{22}^\top \end{bmatrix} = \begin{bmatrix} U_{11}U_{11}^\top & U_{11} U_{21}^\top \\ U_{21} U_{11}^\top & U_{21}U_{21}^\top \end{bmatrix}
    \end{align*}
    Thus
    \begin{align*}
        U_{11}U_{11}^\top = \begin{bmatrix} I_{d_0} & 0 \\ 0 & \cos^2 \frac12 \Theta \end{bmatrix}, U_{11}U_{21}^\top = \begin{bmatrix} 0 & 0 \\ \cos\frac12\Theta \sin\frac12 \Theta & 0 \end{bmatrix}, U_{21}U_{21}^\top = \begin{bmatrix} \sin^2\frac12 \Theta & 0 \\ 0 & 0 \end{bmatrix}
    \end{align*}
    and
    \begin{align*}
        \Sigma[\mu_1] = \frac1{(d+2)} \begin{bmatrix}
        I_{d_0} & 0 & 0 & 0 \\
        0 & \cos^2\frac12\Theta & \cos\frac12\Theta\sin\frac12\Theta & 0 \\
        0 & \cos\frac12\Theta\sin\frac12\Theta & \sin^2\frac12\Theta & 0 \\
        0 & 0 & 0 & 0
        \end{bmatrix}
    \end{align*}
    Doing the calculation verbatim for $\Sigma[\mu_2]$ gives flipped sign for off-diagonl entries. Thus:
    \begin{align*}
        \Sigma[\mu] = \Sigma[\frac12( \mu_1 + \mu_2) ] = \frac1{(d+2)} \begin{bmatrix}
        I_{d_0} & 0 & 0 & 0 \\
        0 & \cos^2\frac12\Theta & 0 & 0 \\
        0& 0 & \sin^2\frac12\Theta & 0 \\
        0 & 0 & 0 & 0
        \end{bmatrix}
    \end{align*}
    This is already a diagonal matrix, and we directly take the diagonal entry to obtain the claim.
\end{proof}

%%%%%%%%%%%%%%%%%%%%%%%%%%%%%%%%%%%%%%%%%%%%%%%%%%%%%%%%%
%%%%%%%%%%%%%%%%%%%%%%%%%%%%%%%%%%%%%%%%%%%%%%%%%%%%%%%%%
\subsection{Singularity Score}

\paragraph{Eigenvalue control}

Before deriving results on singularity score, we first need to derive results on dimension estimation and linear approximation, which are used to define the singularity score. In this section we derive results for controlling the change of eigenvalues of a real symmetric matrix. The real symmetric matrix of interest for us is the covariance matrix, from which we get eigenvalues for dimension estimation. We introduce the following notations. 

\begin{defn}
    Given a symmetric real matrix $A \in \RR^{D \times D}$, we use the following notation for the vector of eigenvalues of $A$, arranged in the decreasing order:
    \[ \vec \lambda A = (\lambda_1 A, \> \ldots \> \lambda_D A) \in \RR^D \]
    We also denote:
    \begin{align*}
        \specgap_k A =& \lambda_k A - \lambda_{k+1} A \\
        \Tail_k A =& \lambda_{k+1} A + \cdots + \lambda_D A \\
        \TQ_k A =& \frac{\Tail_k A}{\Tail_0 A}
    \end{align*}
    where TQ stands for \textit{tail quotient}. For a measure $\mu$ on $\RR^D$, we will use a slight abuse of notation, and denote the spectral gap of its covariance matrix as:
    \[ \specgap_k \mu = \specgap_k \Sigma \mu \]
\end{defn}

The Hoffman-Wielandt theorem \cite{hoffman_wielandt} will be our main tool of controlling the eigenvalues:

\begin{thm}[Hoffman-Wielandt]\label{hoffman_wielandt}
	For normal matrices $A, A'$ of dimension $D \times D$, there is an enumeration of eigenvalues $(\lambda_1, \ldots \lambda_D)$ of $A$ and $(\lambda_1', \ldots \lambda_D')$ of $A'$ such that
	\[ \sum_{i=1}^D |\lambda_i - \lambda_i'|^2 \le \|A - A'\|_{\on{F}}^2\]
	where $\|A\|_{\on{F}} := \sqrt{\on{Tr}(A^\top A})$ denotes the Frobenius norm. In particular, if $A, A'$ are real symmetric matrices, then:
	\[ \|\vec\lambda(A)  - \vec\lambda(A')\| \le \|A-A'\|_{\on{F}} \]
\end{thm}

We will also use the following Lipschitz continuity relation from \cite{uzu_pca}:
\begin{prop}\label{covariance lipschitz}
	Let $\mu, \nu \in \mc P$ such that the support of each measure is contained in a ball of radius $r$.  Then,
	\[ \|\Sigma \mu - \Sigma \nu \| \le 8 r \cdot \Wass(\mu, \nu) \]
\end{prop}

We then get the following bound on the variation of the spectral gap:
\begin{lem}\label{specgap lipschitz}
    Let $\mu, \nu \in \mc P$ be such that the support of each measure is contained in a ball of radius $1$. Then,
    \[ |\specgap_k \mu - \specgap_k \nu | \le 16D \cdot \Wass(\mu, \nu) \]
    % Let $A, B \in \RR^{D \times D}$ be real symmetric matrices. Then for all $k$,
    % \[ |\specgap_k (A) - \specgap_k (A') | \le 2\sqrt{D} \cdot \|A-A'\|_{\on F} \]
\end{lem}
\begin{proof}
    Let $A = \Sigma \mu, A' = \Sigma \nu$. The Hoffman-Wielandt theorem implies the following for all $k$:
    \[ D^{-1/2} \cdot |\lambda_k(A) - \lambda_k(A')| \le D^{-1/2} \cdot \|\vec \lambda(A) - \vec\lambda(A')\|_1 \le \|\vec \lambda(A) - \vec\lambda(A')\|_2 \le \|A - A'\|_{\on F} \]
    where the second inequality is due to the fact that $D^{-1/2} \cdot \|x\|_1 \le \|x\|_2$ for any $x \in \RR^D$. The triangle inequality then implies:
    \[ |\specgap_k (A) - \specgap_k (A') | \le | \lambda_k(A) - \lambda_k(A')| + |\lambda_{k+1}(A) - \lambda_{k+1}(A') | \le 2\sqrt{D} \cdot \|A-A'\|_{\on F} \]
    Now the claim follows by applying Proposition \ref{covariance lipschitz} and the fact that  Frobenius norm satisfies $\|B\|_{\on F} \le \sqrt{D} \cdot \|B\|$ generally for any $B \in \RR^{D \times D}$.
    \[ \|A-A'\|_{\on F} \le \sqrt{D} \cdot \|A - A'\| \le 8 \sqrt{D} \cdot \Wass(\mu, \nu) \]
\end{proof}

The variation of tail quotient can be controlled as follows:

\begin{lem}\label{tq lipschitz}
    Let $\mu, \nu \in \mc P$ be such that the support of each measure is contained in a ball of radius $1$. Assume that $\Wass(\mu, \nu) \le \beta / (16D)$, where $\beta = \| \vec\lambda \Sigma \mu \|_1$. Then the following holds for all $k$:
    \[ |\TQ_k(\mu) - \TQ_k(\nu)| \le 32D\beta^{-1} \cdot \Wass(\mu, \nu) \]
\end{lem}
\begin{proof}
    Denote $A = \Sigma \mu, A' = \Sigma \nu$. The Hoffman-Wielandt theorem and Proposition \ref{covariance lipschitz} imply:
    \begin{align*}
        |\Tail_k A - \Tail_k A'| =& \bigg |\sum_{i > k} \lambda_k A - \lambda_k A' \bigg| \le \sum_{i > k} |\lambda_k A - \lambda_k A'|
        \le \| \vec\lambda A - \vec\lambda A'\|_1 \\
        \le & \sqrt{D} \cdot \|\vec\lambda A - \vec\lambda A' \|_2
        \le \sqrt{D} \cdot \| A - A' \|_{\on F}
        \le 8D \cdot \Wass(\mu, \nu)
    \end{align*}
    where we also used the fact that $D^{-1/2} \cdot \|x\|_1 \le \|x\|_2$ generally for any $x \in \RR^D$ and $\|B\|_{\on F} \le \sqrt{D} \cdot \|B\|$ generally for any $B \in \RR^{D \times D}$. Define the following notations:
    \begin{align*}
        & \alpha = \Tail_k(\Sigma \mu), \quad \beta = \Tail_0(\Sigma \mu) \\
        & t_1 + \alpha = \Tail_k(\Sigma \nu), \quad t_2 + \beta = \Tail_0(\Sigma \nu), \quad t = 8D \cdot \Wass(\mu, \nu)
    \end{align*}
    From the above we know that $|t_1|, |t_2| \le t$ and by assumption $t \le \beta/2$. By simple calculation the claim follows:
    \begin{align*}
        \bigg| \frac{\alpha + t_1}{\beta + t_2} - \frac{\alpha}{\beta}\bigg|  = \bigg| \frac{t_1 \beta - t_2 \alpha}{\beta(\beta + t_2)} \bigg| \le \frac{t(\beta + \alpha)}{\beta(\beta-t)} \le \frac{2\beta t}{\beta^2/2} = 4\beta^{-1} t
    \end{align*}
\end{proof}

\paragraph{Stability}

In this section we show that when spectral gap is bounded from below and the estimated dimension are constant, then the singularity score obeys a Lipschitz continuity relation. We first note the following variant of the Davis-Kahan theorem \cite{davis_kahan, davis_kahan_variant}:

\begin{thm}[Davis-Kahan-Wang-Samworth]
	Let $A, B \in \RR^{D \times D}$ be real symmetric matrices. Let $1 \le d_1 \le d_2 \le D$ and assume that $\min ( \specgap_{d_1 - 1}A, \specgap_{d_2} A) > 0$. Let $\pi_A$ be the span of the eigenspaces corresponding to $\lambda_{d_1}, \lambda_{d_1 + 1}, \ldots \lambda_{d_2}$, and let $\theta_1 \le \ldots \le \theta_d$ be the principal angles between $(\pi_A, \pi_B)$. Then we have:
	\[ \sqrt{\sin^2 \theta_{d_1} + \cdots + \sin^2 \theta_{d_2}} \le \frac2{\min(\specgap_{d_1 - 1} A, \specgap_{d_2} A)} \cdot \min\bigg( \|A-B\|_{\on F}, \> \sqrt{d} \|A-B\| \bigg) \]
	In particular, for $(d_1, d_2) = (1, d)$, we have:
	\[ \sqrt{\sin^2 \theta_{1} + \cdots + \sin^2 \theta_{d}} \le \frac{2 }{\specgap_d A} \cdot \min\bigg( \|A-B\|_{\on F}, \> \sqrt{d} \|A-B\| \bigg) \]
\end{thm}

We will only be using the case of $(d_1, d_2) = (1, d)$ above.

We then prove the following Lipschitz continuity relation for the singularity score:

\begin{prop}\label{sing lipschitz}
    Let $\mu, \nu \in \mc P$ be measures whose supports are contained in the ball of radius 1 centered at the origin of $\RR^D$. Assume that for some $\eta \in (0,1)$ and $k>0$, we have $\hat d_\eta(\mu) = \hat d_\eta(\nu)$, and define $s = \max(\specgap_k \Sigma \mu , \specgap_k \Sigma \nu)$. Then the following hold:
    \begin{align*}
        \mathfrak{S} ( \tse_\eta \mu , \tse_\eta \nu) \le & 8s^{-1} \cdot \Wass(\mu, \nu) \\
        |\sigma_\eta(\mu) - \sigma_\eta(\nu)| \le & (1 + 8s^{-1}) \Wass(\mu, \nu)
    \end{align*}
    where $\mathfrak{S}(\pi_1, \pi_2) = \sqrt{\sin^2\theta_1 + \cdots + \sin^2 \theta_d}$, with $(\theta_1, \ldots \theta_d)$ being the principal angles between $(\pi_1, \pi_2)$. 
\end{prop}
\begin{proof}
    Let's abbreviate $\sigma = \sigma_\eta, \tse = \tse_\eta$. By assumption $\dim \tse \mu = \dim \tse \nu = k$ and we now apply the Davis-Kahan theorem and Proposition \ref{covariance lipschitz}:
    \begin{align*}
        \mathfrak{S} ( \tse \mu , \tse \nu) \le \frac1s \|\Sigma \mu - \Sigma \nu\| \le \frac8s \Wass(\mu, \nu)
    \end{align*}
    For the singularity score, we get:
    \begin{align*}
        |\sigma(\mu) - \sigma(\nu)| = |\MMD( \mu_\perp, \unif_k) - \MMD( \nu_\perp, \unif_k)|
        \le \MMD(\mu_\perp, \nu_\perp)
        \le \Wass(\mu_\perp, \nu_\perp)
    \end{align*}
    where $\mu_\perp = \Pi(\mu, \tse \mu), \nu_\perp = \Pi(\nu, \tse \nu)$. Furthermore,
    \begin{align*}
        \Wass ( \mu^\perp, \nu^\perp) =& \Wass \bigg( \Pi(\mu, \tse \mu), \Pi(\nu, \tse \nu) \bigg) \\
        \le & \Wass\bigg( \Pi(\mu, \tse \mu), \Pi(\nu, \tse \mu) \bigg) + \Wass \bigg( \Pi(\nu, \tse \mu), \Pi(\nu, \tse \nu) \bigg) \\
        \le & \Wass(\mu, \nu) + \mathfrak{S} (\tse \mu, \tse \nu) \\
        \le & \left( 1 + \frac 8s \right) \Wass(\mu, \nu)
    \end{align*}
    where in the second to last inequality we applied Lemmas \ref{mproj same} and \ref{mproj tilt}. 
\end{proof}

In the following Proposition, spectral gap, tail quotient, dimension estimate, and the singularity score are simultaneously controlled, for a measure $\nu$ sufficiently close to a given measure $\mu$. In our application of the Proposition, $\nu$ will be an empirical measure, from which the empirical singularity score will be calculated.
\begin{prop}\label{sing control}
	Let $\mu, \nu \in \mc P$ be measures supported on the unit ball $\Ball(0, 1) \subset \RR^D$. Let $a \in [0,1]$ and $k\ge 0$ be an integer. Suppose that $\Wass(\mu, \nu)$ is sufficiently small; explicitly, assume that:
	\[ \Wass(\mu, \nu) \le \frac{\min(4as, 4\beta, a\beta G)}{64D}\]
where $\beta = \|\vec \lambda \Sigma \mu\|_1$, $G_k = \TQ_{k-1}(\mu) - \TQ_k(\mu)$, and $s = \specgap_k(\mu)$. Then the following hold:
 %\min\bigg( \frac{\beta(\mu)}{16D}, \> \frac{a \cdot G_k(\mu) \cdot \beta }{64 D}, \> \frac{a \cdot \specgap_k(\mu) }{16D} \bigg) \]
    
    (1) There is a bound for spectral gaps:
    \[ \specgap_k(\nu) \ge (1-a) \specgap_k(\mu) \]

    (2) There are inclusions of intervals of tail quotients:
    \[ J_{k, 0}(\nu) \supseteq J_{k, a}(\mu), \quad J_{k, 0}(\mu) \supseteq J_{k, a}(\mu) \]
    where $J_{k, a}(\mu) = [\TQ_k(\mu) + \frac a2 G_k, \> \TQ_{k-1}(\mu) - \frac a2 G_k]$. 
    
    (3) For a choice of $\eta \in J_{k, a}(\mu)$, the following hold:
	\begin{align*}
		& \hat d_\eta(\mu) = \hat d_\eta(\nu) = k \\
		& \mathfrak{S}(\pi_1, \pi_2) \le 8s^{-1} \cdot \Wass(\mu, \nu) \\
		& |\sigma_\eta \mu - \sigma_\eta \nu| \le (1 + 8s^{-1}) \cdot \Wass(\mu, \nu)
	\end{align*}
    where $\mathfrak{S}(\pi_1, \pi_2) = \sqrt{\sin^2\theta_1 + \cdots + \sin^2 \theta_d}$, with $(\theta_1, \ldots \theta_d)$ being the principal angles between $(\pi_1, \pi_2)$. 
\end{prop}
\begin{proof}
    (1) By Lemma \ref{specgap lipschitz} and the assumption on $\Wass(\mu, \nu)$, we have:
    \[ |\specgap_k \mu - \specgap_k \nu| \le 16D \cdot \Wass(\mu, \nu) \le a \cdot s \]
    (2) By Lemma \ref{tq lipschitz} and the assumption on $\Wass(\mu, \nu)$, for each $j$ we have:
    \begin{align*}
        & |\TQ_j(\mu) - \TQ_j(\nu)| \le \frac{32D}{\beta} \cdot \Wass(\mu, \nu) \le \frac a2 \cdot G_k(\mu)
    \end{align*}
    This implies the first inclusion, and the second inclusion holds trivially.

    (3) By (2), $\eta \in J_{k, a}(\mu)$ implies both $\eta \in J_{k,0}(\nu)$ and $\eta \in J_{k, 0}(\mu)$, so that $\hat d_\eta(\mu) = \hat d_\eta(\nu) = k$ by the definition of $\hat d_\eta$. The rest of the claims follow from Proposition \ref{sing lipschitz}.
\end{proof}

\paragraph{Limit behaviour}

\begin{prop}\label{sing limit smooth}
    Let $M \subset \RR^D$ be a $d$-dimensional submanifold. Suppose $\eta \in (0, (d+2)^{-1})$. Then for any $x \in M$, $\sigma_\eta(\mu_{x, 0}) = 0$.
\end{prop}
\begin{proof}
    By Proposition \ref{wass lipschitz 1}, we have $\mu_{x, 0} = \mc H^d|_{T_x^\circ M}$, where $T_x^\circ M = T_x M \cap \Ball(0, 1)$. Also noting that the spectral gap of $\mu_{x,0}$ is $(d+2)^{-1}$, we obtain the claim.
\end{proof}

\begin{prop}\label{sing limit singular}
    Let $M = M_1 \cup M_2 \subset \RR^D$ be a union of two $d$-dimensional submanifolds such that for any $x \in M_1 \cap M_2$, we have $\dim(T_x M_1 \cap T_x M_2) = d_0$, and all principal angles of $(T_x M_1, T_x M_2)$ are bounded above a fixed constant $\phi > 0$. Suppose $\eta \in (0, (d+2)^{-1} \cdot \sin^2(\phi/2))$. Then the function $x \mapsto \sigma_\eta(\mu_{x, 0})$ is continuous on $M_1 \cap M_2$, and takes positive values. In particular, we have $\inf_{x \in M_1 \cap M_2} \sigma_\eta(\mu_{x, 0}) > 0$. 
\end{prop}
\begin{proof}
    Due to the eigenvalue computation in Proposition \ref{covariance of two disks}, the condition $\eta < d^{-1} \cdot \sin^2(\phi/2)$ implies
    \[ x \in M_1 \cap M_2 \implies \hat d_\eta(\mu_{x, 0}) = 2d-d_0 \]
    Also, we have a lower bound on the $(2d-d_0)$-th spectral gap:
    \[ \specgap_{2d-d_0}(\Sigma \mu_{x, 0}) \ge \frac{\sin^2(\phi/2)}{d+2} \]
    Therefore we can apply Proposition \ref{sing lipschitz}, and see that the function $x \mapsto \sigma_\eta(\mu_{x, 0})$ is (Lipschitz) continuous on $M_1 \cap M_2$.

    The projected measure $(\mu_{x, 0})_\perp$ is the (pushforward along) projection of $\mu_{x, 0}$ to the $(2d-d_0)$-dimensional space spanned by $T_x M_1 + T_x M_2$. This measure, supported along the union of two $d$-dimensional disks, is clearly not equal to the $(2d-d_0)$-dimensional uniform measure $\unif_{2d-d_0}$. Then the universality of kernel MMD implies that:
    \[ \sigma_\eta(\mu_{x,0}) = \MMD\bigg( (\mu_{x, 0})_\perp, \> \unif_{2d-d_0} \bigg) > 0 \]
    Therefore the function $x \mapsto \sigma_\eta(\mu_{x, 0})$ is continuous and positive on a compact set $M_1 \cap M_2$, so that its infimum is also positive.
\end{proof}

%%%%%%%%%%%%%%%%%%%%%%%%%%%%%%%%%%%%%%%%%%%%%%%%%%%%%%%%%
%%%%%%%%%%%%%%%%%%%%%%%%%%%%%%%%%%%%%%%%%%%%%%%%%%%%%%%%%
\subsection{Proof of the main theorem}\label{sect: proof_main_outline}

In this section we prove Theorem \ref{thm:main}, the main mathematical result of this article.

\vspace{5mm}
\textbf{Definitions.}

For the logical clarity of the proof, we will first define some constants, and postpone the explanation for their choice to later parts of the proof. 

$\tau, \psi, \zeta, s_0$ are defined as:
\[ \tau = \min(\tau_1, \tau_2), \quad \psi = \frac{\sin^2(\phi/2)}{d+2}, \quad \zeta = \frac{\psi \cdot d}{128(d+2)}, \quad s_0 = 1 + \frac 8{\psi} \]
$\eta_-, \eta_+$ are defined as:
\[ \eta_- = \frac 14 \psi, \quad \eta_+ = \frac 34 \psi \]
$\xi, \xi_0$ are defined as:
\[ 3\xi = \inf_{x \in M_1 \cap M_2} \sigma_{\psi/2} (\mu_{x, 0}), \quad \xi_0 = \min\bigg( \zeta, \frac{\xi}{s_0}\bigg) \]
$r_0, c_A, c_B$ are defined as:
\begin{align*}
    & r_0 = \min \bigg( c_5, c_7, c_9, \frac{\xi_0}{2c_{8}}, \frac{\xi_0}{8c_6}, \frac{\xi_0}{4c_{10}}\bigg) \cdot \tau \\
    & c_A = \min \bigg( c_5,  \frac{\xi_0}{8c_6} \bigg) \\
    & c_B = \max \bigg( h(M_1, M_2)^{-1}, h(M_2, M_1)^{-1} \bigg) \\
    \text{where } & h(M_1, M_2) = \inf_{x \in M_1} \frac{\d(x, M_2)}{\d(x, M_1 \cap M_2)}
\end{align*}
where the constants $c_5, \ldots c_{10}$, which depend only on $d$, are defined in Propositions \ref{wass lipschitz 1}, \ref{wass lipschitz 2}, and \ref{wass lipschitz 3}.

Finally, we \textit{fix a choice} of $\eta, r$ as any number in the range:
\[ \eta \in [\eta_-, \eta_+], \quad r \in (0, r_0] \]
We remark that if $\nu \in \mc P$ and $\hat d_\eta(\nu) = \hat d_{\eta'}(\nu)$ for some threshold values $\eta, \eta'$, then we have $\sigma_\eta(\nu) = \sigma_{\eta'}(\nu)$. In particular, due to Propositions \ref{covariance of two disks}, \ref{sing limit singular}, $x \in M_1 \cap M_2$ implies $\hat d_{\psi/2}(\mu_{x,0}) = \hat d_{\eta}(\mu_{x, 0})$, and thus:
\[ 3\xi = \inf_{x \in M_1 \cap M_2} \sigma_{\psi/2}(\mu_{x, 0}) = \inf_{x \in M_1 \cap M_2} \sigma_\eta(\mu_{x, 0}) \]

\vspace{5mm}
\textbf{Outline.} 

We will first describe the non-random situation in detail and then describe the randomness using the Wasserstein concentration inequality. Let $\mathbf x = (x_1, \ldots x_n) \subset M$. Define the (non-random) singularity scores:
\[ \sigma_i = \sigma_i(\mathbf x, r, \eta) = \sigma_\eta(\hat \mu_i) \]
where $\hat \mu_i$ is defined using $(\mathbf x, r)$ as described in the Introduction.

Our strategy of proof involves the following successive approximations:
\begin{align*}
    & \text{Singular part: } \d(x_i, M_1 \cap M_2) \le c_A r \implies \sigma(\hat \mu_i) \approx \sigma(\mu_{x_i, r}) \approx \sigma(\mu_{y_i, r}) \approx \sigma(\mu_{y_i, 0}) \ge 3\xi \\
    & \text{Smooth part: } \d(x_i, M_1 \cap M_2) \ge c_B r \implies \sigma(\hat \mu_i) \approx \sigma(\mu_{x_i, r}) \approx \sigma(\mu_{x_i, 0}) = 0
\end{align*}
where $y_i$ is the projection from $x_i$ to $M_1 \cap M_2$. By the choice of parameters made before, the approximations will each amount to at most $\xi$ of error, so that in the smooth case we have $\sigma(\hat \mu_i) \le \xi$ and in the singular case we have $\sigma(\hat \mu_i) \ge 2\xi$. We now describe the proof precisely. 

\vspace{3mm}
\textit{Limit behaviour.} We apply Propositions \ref{sing limit singular} and \ref{sing limit smooth}, and see that:
\begin{align*}
    x \in (M_1 \cup M_2) \backslash (M_1 \cap M_2) \implies & \sigma (\mu_{x, 0}) = 0 \\
    x \in M_1 \cap M_2 \implies & \sigma (\mu_{x, 0}) \ge 3\xi > 0
\end{align*}

\textit{Singular part.} When $\d(x_i, M_1 \cap M_2) \le c_A r$, the following holds:
\begin{align}
    \sigma_i \ge & \sigma(\mu_{y,0}) - |\sigma(\hat\mu_i) - \sigma(\mu_{y_i, 0})| = 3\xi - |\sigma(\hat\mu_i) - \sigma(\mu_{y_i, 0})| \label{eq_sing}
\end{align}
where $y_i \in M_1 \cap M_2$ is a point satisfying $\d(x_i, y_i) = \d(x_i, M_1 \cap M_2)$\footnote{Compactness of $M_1 \cap M_2$ and continuity of the distance function implies that such a $y$ exists.}. 

\textit{Smooth part.} When $\d(x_i, M_1 \cap M_2) \ge c_B r$, the following holds:
\begin{align}
    \sigma_i \le & \sigma(\mu_{x_i,0}) + |\sigma(\hat\mu_i) - \sigma(\mu_{x_i, 0})| = 0 + |\sigma(\hat\mu_i) - \sigma(\mu_{x_i, 0})| \label{eq_sm}
\end{align}
Even though Equations \eqref{eq_sing} and \eqref{eq_sm} didn't use anything specific about the distance $\d(x_i, M_1 \cap M_2)$, this will be used while controlling the error terms.

\vspace{5mm}
\textbf{From singularity score to Wasserstein distance.}

Differences of singularity scores are controlled using Proposition \ref{sing control}, which is a Lipschitz continuity relation with respect to the Wasserstein distance. Our definition of $\zeta$ is obtained by setting $a=1/2$, $\beta = d/(d+2)$, $G = s = \psi$ in the condition in Proposition \ref{sing control}. This then implies that for all $x \in M$ and $\nu \in \mc P$,
\[ \Wass(\mu_{x, 0}, \nu) \le \zeta \implies |\sigma (\mu_{x, 0}) - \sigma(\nu) | \le s_0 \cdot \Wass(\mu_{x, 0}, \nu) \]
where we recall our definition $s_0 = 1+8\psi^{-1}$. The definition of $\xi_0$ allows us to make a more straightforward inference:
\begin{align}
    \Wass(\mu_{x, 0}, \nu) \le  \xi_0 \implies |\sigma (\mu_{x, 0}) - \sigma (\nu) | \le \xi \label{eq wass sing}
\end{align}
Therefore we can control error terms in Equation \eqref{eq_sing}, \eqref{eq_sm} using the Wasserstein distance.

\vspace{5mm}
\textbf{Wasserstein distance control.}

\textit{Singular part.} Suppose that $x$ satisfies $\d(x, M_1 \cap M_2) \le c_A r$. Let $y \in M_1 \cap M_2$ satisfy $\d(x, y) = \d(x, M_1 \cap M_2)$. We denote $\rho = r/\tau$ and also set $s = \|x - y \| / r$. Then Propositions \ref{wass lipschitz 1} and \ref{wass lipschitz 2} imply that:
\begin{align*}
    \rho, \> s \le c_5 & \implies \Wass(\mu_{x,r}, \mu_{y,r}) \le c_6 (\rho + s ) \\
    \rho \le c_9 & \implies \Wass(\mu_{y, r}, \mu_{y, 0}) \le c_{10} \rho
\end{align*}
Our definitions of $r_0, c_A$ allows us to apply the bounds above, and we obtain:
\begin{align}
    \Wass(\mu_{x, r}, \mu_{y, 0}) \le \Wass(\mu_{x, r}, \mu_{y, r}) + \Wass(\mu_{y, r}, \mu_{y, 0}) \le \frac{\xi_0}2 \label{eq wass control 1}
\end{align}

\textit{Smooth part.} Suppose that $x$ satisfies $\d(x, M_1 \cap M_2) \ge c_B r$. Here our choice of $c_B$ allows us to apply Proposition \ref{separation}, so that $\Ball(x, r)$ intersects \textit{either only one} of $M_1$ or $M_2$. Thus we only need to work with one manifold at a time here. Thus Proposition \ref{wass lipschitz 3} implies:
\begin{align*}
    \rho \le c_7 \implies \Wass(\mu_{x, r}, \mu_{x, 0}) \le c_{8} \rho
\end{align*}
and yet again by our definition of $r_0$, this bound implies:
\begin{align}
    \Wass(\mu_{x, r}, \mu_{x, 0}) \le \frac{\xi_0}2 \label{eq wass control 2}
\end{align}

\vspace{5mm}
\textbf{Empirical estimation.}

Almost all of the puzzle pieces have been fit together to complete the proof. It now remains to control the probability of empirical esimtation. 

We reintroduce randomness, and let $\mathbf X_n = (X_1, \ldots X_n)$ be an iid sample drawn uniformly from $M_1 \cup M_2$. The choice of all other parameters remain the same as before. We plug in the error level of $t = r \xi_0/2$ to Proposition \ref{main_wass_conc}\footnote{Instead of $t = \xi_0 / 2$, we plug in $t = r \xi_0 / 2$ because we are controlling the Wasserstein distance between measures that have been rescaled by the factor of $r^{-1}$.}, and obtain the following. Whenever $n \ge \max(N, 2/u_-)$, we have:
\begin{align}
    \Prob\bigg( \max_i \Wass(\hat \mu_i, \mu_{X_i, r}) \le \frac{\xi_0}2 \bigg) \ge 1 - \delta_m
\end{align}
where $\lim_{m \rightarrow \infty} \delta_m = 0$ exponentially fast, given explicitly as:
\[ \delta = c \cdot n^{N+1} \gamma^n \]
where
\begin{align*}
    & c = \bigg(\frac{u_+}{1 - u_+}\bigg)^N, \quad N = \bigg\lceil\bigg( \frac{408}{\xi_0} \bigg)^D\bigg\rceil, \quad \gamma = 1 - u_-(1 - e^{-\xi_0^2/32}) \\
    & \quad u_- = \inf_{x \in \on{supp}\mu} \mu(\Ball(x, r)), \quad u_+ = \sup_{x \in \on{supp}\mu} \mu(\Ball(x, r))
\end{align*}
Therefore there exists some $n_0 > 0$ such that, for the $\delta > 0$ given in our theorem, $n \ge n_0$ implies $\delta_n \le \delta$. Note that this $n_0$ depends on $\delta, \mu, r, \xi_0$, which have already been fixed in the beginning of the proof. 

\vspace{5mm}
\textbf{Combining the bound.}

We now complete the proof. When $n \ge n_0$, the following holds for every $i$, with probability at least $1-\delta$:
\[ \Wass(\hat\mu_i, \mu_{X_i, r}) \le \frac{\xi_0}2 \]
Equations \eqref{eq wass control 1} and \eqref{eq wass control 2} apply verbatim for the random setting:
\begin{align*}
    \d(X_i, M_1 \cap M_2) \le c_A r \implies & \Wass(\mu_{X_i, r}, \mu_{Y_i, 0}) \le \frac{\xi_0}2 \\
    \d(X_i, M_1 \cap M_2) \ge c_B r \implies & \Wass(\mu_{X_i, r}, \mu_{X_i, 0}) \le \frac{\xi_0}2
\end{align*}
where $Y_i \in M_1 \cap M_2$ is a point satisfying $\d(X_i, Y_i) = \d(X_i, M_1 \cap M_2)$. Therefore by the triangle inequality, the above two equations imply that:
\begin{align*}
    \d(X_i, M_1 \cap M_2) \le c_A r \implies & \Wass(\hat \mu_i, \mu_{Y_i, 0}) \le \xi_0 \\
    \d(X_i, M_1 \cap M_2) \ge c_B r \implies & \Wass(\hat \mu_i, \mu_{X_i, 0}) \le \xi_0
\end{align*}
This precisely fits the condition in Equation \eqref{eq wass sing}, from which we obtain that:
\begin{align*}
    \d(X_i, M_1 \cap M_2) \le c_A r \implies & |\sigma(\hat \mu_i) - \sigma(\mu_{Y_i, 0})| \le \xi \\
    \d(X_i, M_1 \cap M_2) \ge c_B r \implies & |\sigma(\hat \mu_i)- \sigma(\mu_{X_i, 0})| \le \xi
\end{align*}
Plugging them into Equations \eqref{eq_sing} and \eqref{eq_sm}, we obtain the conclusion of the theorem:
\begin{align*}
    \d(X_i, M_1 \cap M_2) \le c_A r \implies & \sigma(\hat \mu_i) \ge 2\xi \\
    \d(X_i, M_1 \cap M_2) \ge c_B r \implies & \sigma(\hat \mu_i) \le \xi
\end{align*}

\end{document}